\RequirePackage{fix-cm}
\documentclass[twocolumn]{svjour3}          
\smartqed  

\usepackage{amsmath}
\usepackage{gensymb}
\usepackage{diagbox}
\usepackage{graphics}
\usepackage{epsfig}
\usepackage{threeparttable}
\usepackage{xspace}
\usepackage{siunitx}
\usepackage{graphicx}
\usepackage{subfig}
\usepackage{placeins}
\usepackage{blindtext}
\usepackage{amssymb}
\usepackage{threeparttable}
\usepackage{color}
\usepackage[normalem]{ulem}
\usepackage{multirow}
\usepackage{float}
\usepackage{amsfonts}
\usepackage{bm}
\usepackage{bbm}
\usepackage{array}
\usepackage{tabulary}	
\usepackage{microtype}
\usepackage{booktabs}
\usepackage{xspace}
\usepackage[table]{xcolor}
\usepackage{colortbl}
\usepackage{diagbox}
\usepackage{rotating}
\usepackage{booktabs}
\usepackage{overpic}
\usepackage{enumitem}
\usepackage{colortbl}
\usepackage{soul}
\usepackage{url}
\usepackage{makecell, verbatim, sidecap}
\usepackage{algorithm}
\usepackage{algorithmic}
\usepackage{cite}

\definecolor{citecolor}{HTML}{0071bc}
\usepackage[pagebackref=true,breaklinks=true,colorlinks,citecolor=citecolor,bookmarks=false]{hyperref}

\definecolor{scorered}{HTML}{e4485a}
\definecolor{scoreblue}{HTML}{4a7ee8}
\definecolor{scoregreen}{HTML}{80ba0e}
\definecolor{scoreyellow}{HTML}{d8ac0d}
\definecolor{scorepurple}{HTML}{846bc8}

\definecolor{purple0}{HTML}{e9e9f3}
\definecolor{purple}{HTML}{dcdaed}
\definecolor{purple1}{HTML}{bab6da}
\definecolor{blue0}{HTML}{b6d9f0}
\definecolor{blue1}{HTML}{80b1d1}
\definecolor{blue2}{HTML}{2a8ed1}
\definecolor{blue3}{HTML}{0071bc}

\definecolor{mygray00}{gray}{.3}
\definecolor{mygray0}{gray}{.6}
\definecolor{mygray}{gray}{.85}
\definecolor{mygray1}{gray}{.9}
\definecolor{mygray2}{gray}{.95}

\newcommand{\tabincell}[2]{\begin{tabular}{@{}#1@{}}#2\end{tabular}}

\makeatletter
\newcommand{\thickhline}{%
    \noalign {\ifnum 0=`}\fi \hrule height 1pt
    \futurelet \reserved@a \@xhline
}
\makeatother

\makeatletter
\DeclareRobustCommand\onedot{\futurelet\@let@token\@onedot}
\def\@onedot{\ifx\@let@token.\else.\null\fi\xspace}

\makeatother

\newcommand{\app}{\raise.17ex\hbox{$\scriptstyle\sim$}}

\journalname{IJCV}
\begin{document}
\title{Deep Learning Technique for Human Parsing: A Survey and Outlook}

\author{Lu Yang, Wenhe Jia, Shan Li, Qing Song$\dagger$}

\institute{Lu Yang, Wenhe Jia, Shan Li, Qing Song are with the Beijing University of Posts and Telecommunications,  Beijing, 100876, China (e-mail: soeaver@bupt.edu.cn; jiawh@bupt.edu.cn; ls1995@bupt.edu.cn; priv@bupt.edu.cn) \\
             $\dagger$ Corresponding author: Qing Song
}

\date{Received: 14 June 2023 / Accepted:  9 February 2024}

\maketitle

\begin{abstract}
Human parsing aims to partition humans in image or video into multiple pixel-level semantic parts. In the last decade, it has gained significantly increased interest in the computer vision community and has been utilized in a broad range of practical applications, from security monitoring, to social media, to visual special effects, just to name a few. Although deep learning-based human parsing solutions have made remarkable achievements, many important concepts, existing challenges, and potential research directions are still confusing. In this survey, we comprehensively review three core sub-tasks: single human parsing, multiple human parsing, and video human parsing, by introducing their respective task settings, background concepts, relevant problems and applications, representative literature, and datasets. We also present quantitative performance comparisons of the reviewed methods on benchmark datasets. Additionally, to promote sustainable development of the community, we put forward a transformer-based human parsing framework, providing a high-performance baseline for follow-up research through universal, concise, and extensible solutions. Finally, we point out a set of under-investigated open issues in this field and suggest new directions for future study. We also provide a regularly updated project page, to continuously track recent developments in this fast-advancing field: \url{https://github.com/soeaver/awesome-human-parsing}.
\end{abstract}

\section{Introduction}
\label{sec:introduction}

Human parsing \cite{yamaguchi2012parsing, liang2018look, wang2019learning, wang2020hierarchical, li2022deep}, considered as the fundamental task of human-centric visual understanding \cite{lin2020human}, aims to classify the human parts and clothing accessories in images or videos at pixel-level. Numerous studies have been conducted on human parsing due to its crucial role in widespread application areas, \emph{e.g}., security monitoring, autonomous driving, social media, electronic commerce, visual special effects, artistic creation, giving birth to various excellent human parsing solutions and applications.

\begin{figure}
	\begin{center}
		\includegraphics[width=0.99\linewidth]{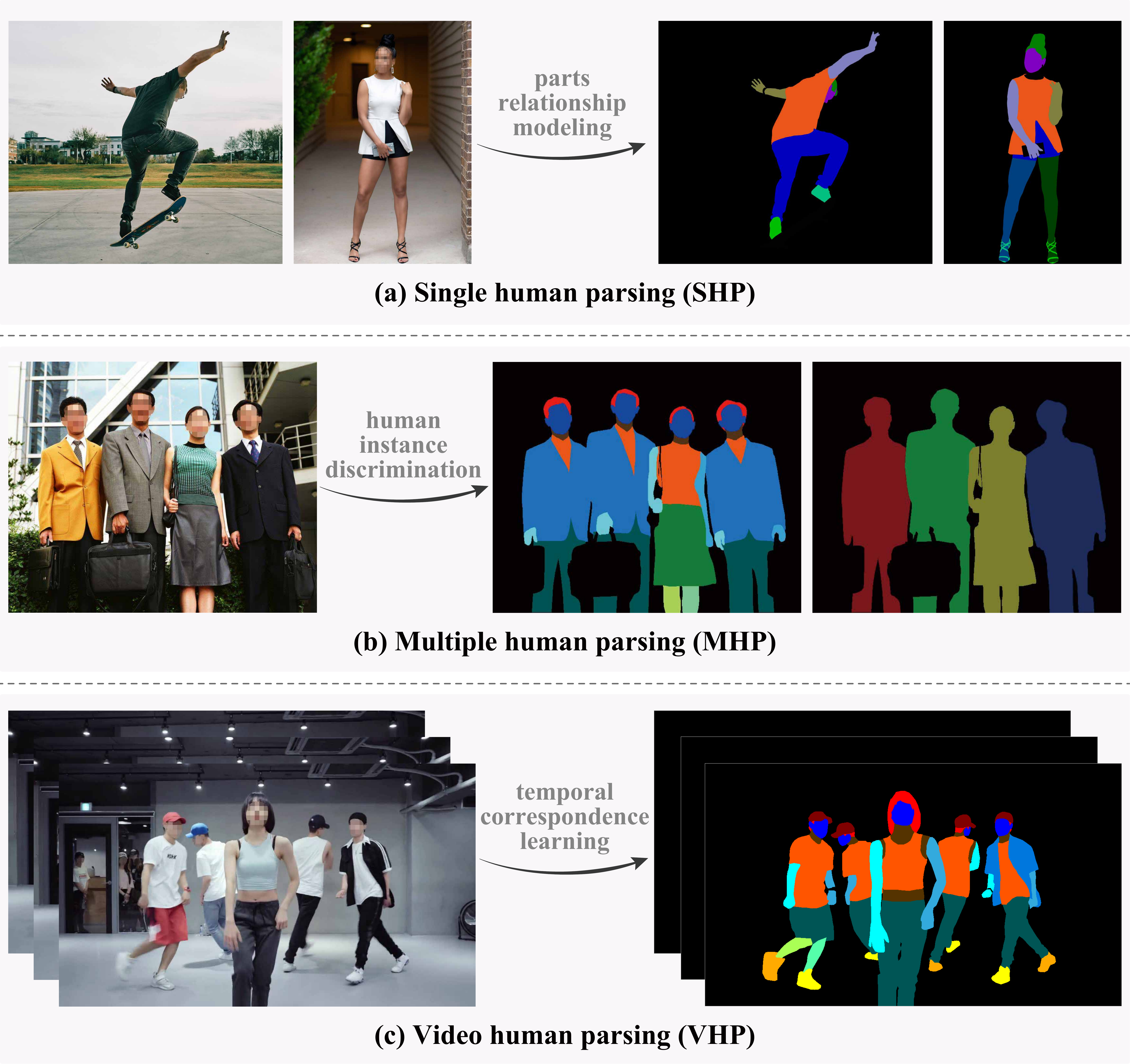}
	\end{center}
	\vspace{-5pt}
	\captionsetup{font=small}
	\caption{\small{\textbf{Human parsing tasks reviewed in this survey:} (a) single human parsing (SHP) \cite{cvpr2021l2id}; (b) multiple human parsing (MHP) \cite{gong2018instance}; (c) video human parsing (VHP) \cite{zhou2018adaptive}.}}
	\label{fig:taxonomy}
	\vspace{-10pt}
\end{figure}

As early as the beginning of this century, some studies tried to identify the level of upper body clothing \cite{borras2003high}, the grammatical representations of clothing \cite{chen2006composite} and the deformation of body contour \cite{guan2010a} under very limited circumstances. These early studies facilitated the research on pixel-level human parts and clothing recognition, \emph{i.e}., human parsing task. Immediately afterward, some traditional machine learning and computer vision techniques were utilized to solve human parsing problems, \emph{e.g}., structured model \cite{yang2011articulated, dong2013deformable, yamaguchi2012parsing}, clustering algorithm \cite{caron2018deep}, grammar model \cite{zhu2008max, dong2014towards}, conditional random field \cite{kae2013augmenting, ladicky2013human, yamaguchi2013paper}, template matching \cite{bo2011shape, liang2015deep} and super-pixel \cite{fulkerson2009class, tighe2010super, liu2013fashion}. Afterward, the prosperity of deep learning and convolutional neural network \cite{krizhevsky2012imagenet, girshick2014rich, jia2014caffe, lecun2015deep, szegedy2015going, shelhamer2016fully, he2016deep} has further promoted the vigorous development of human parsing. Attention mechanism \cite{chen2016attenttion, liang2016object, yang2018attention, cheng2019spgnet}, scale-aware features \cite{liang2015human, xiq2016zoom, zhang2020pcnet, yang2021quality}, tree structure \cite{wang2019learning, ji2020learning}, graph structure \cite{gong2019graphonomy, wang2020hierarchical, zhang2022human}, edge-aware learning \cite{ruan2019devil, zhang2020correlating, liu2020hybrid}, pose-aware learning \cite{liang2018look, nie2018mutual, zhao2022from} and other technologies \cite{liu2018cross, luo2018macro, li2020self, li2020correction} greatly improved the performance of human parsing. However, some existing challenges and under-investigated issues make human parsing still a task worthy of further exploration.


\begin{figure*}
	\begin{center}
		\includegraphics[width=0.95\linewidth]{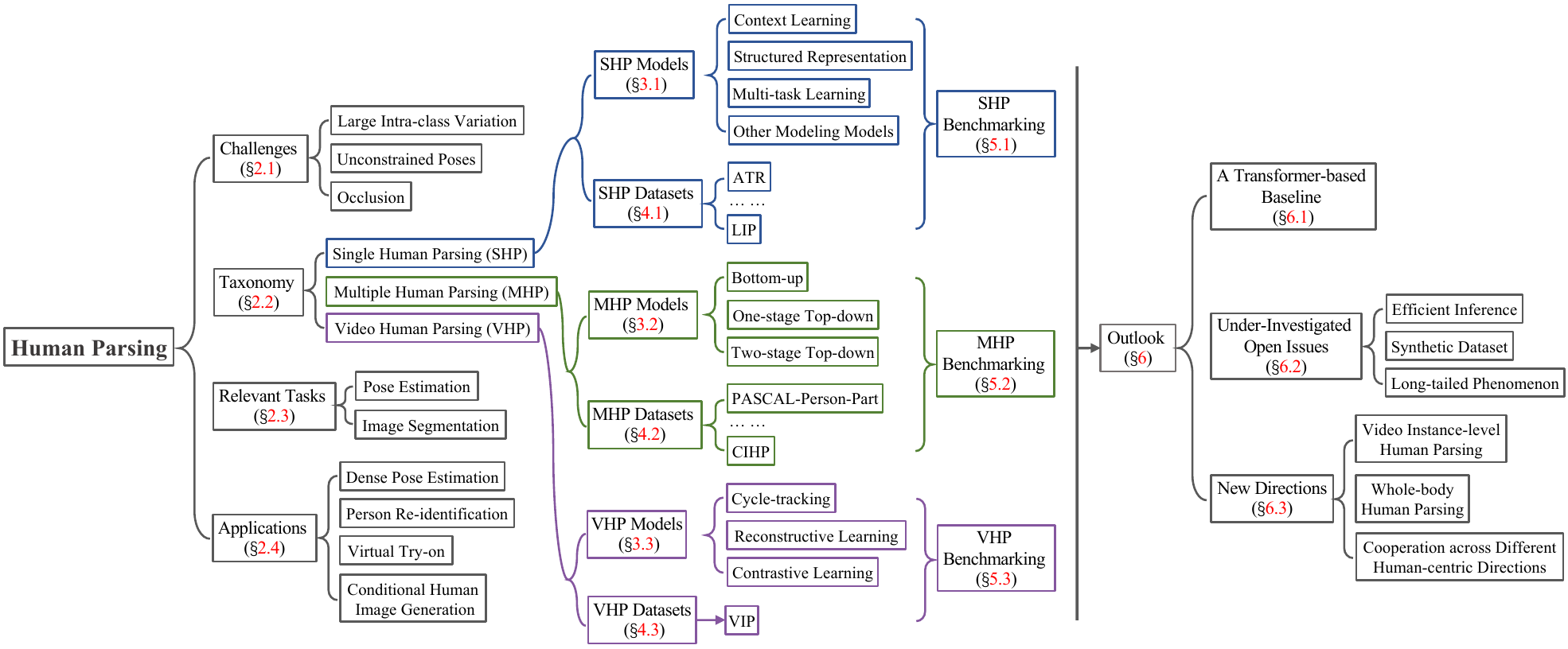}
	\end{center}
	\vspace{-5pt}
	\captionsetup{font=small}
	\caption{\small{\textbf{Outline of this survey.}}}
	\label{fig:outline}
	\vspace{-10pt}
\end{figure*}

With the rapid development of human parsing, several literature reviews have been produced. However, existing surveys are not precise and in-depth: some surveys only provide a superficial introduction of human parsing from a macro fashion/social media perspective \cite{mameli2021deep, cheng2021fashion}, or only review a sub-task of human parsing from a micro face parsing perspective \cite{khan2020face}. In addition, due to the fuzziness of taxonomy and the diversity of methods, comprehensive and in-depth investigation is highly needed and helpful. In response, we provide the first review that systematically introduces background concepts, recent advances, and an outlook on human parsing.

\vspace{-6pt}
\subsection{Scope}
This survey reviews human parsing from a comprehensive perspective, including not only single human parsing (Figure~\ref{fig:taxonomy} (a)) but also multiple human parsing (Figure~\ref{fig:taxonomy} (b)) and video human parsing (Figure~\ref{fig:taxonomy} (c)). 
At the technical level, this survey focuses on the deep learning-based human parsing methods and datasets in recent ten years. To provide the necessary background, it also introduces some relevant literature from non-deep learning and other fields. At the practical level, the advantages and disadvantages of various methods are compared, and detailed performance comparisons are given. In addition to summarizing and analyzing the existing work, we also give an outlook for the future opportunities of human parsing and put forward a new transformer-based baseline to promote sustainable development of the community. A curated list of human parsing methods and datasets and the proposed transformer-based baseline can be found at \url{https://github.com/soeaver/awesome-human-parsing}.

\vspace{-6pt}
\subsection{Organization}
Figure~\ref{fig:outline} shows the outline of this survey. \S\ref{sec:pre} gives some brief background on problem formulation and challenges (\S\ref{sec:problem}), human parsing taxonomy (\S\ref{sec:taxonomy}), relevant tasks (\S\ref{sec:relevant}), and applications of human parsing (\S\ref{sec:parsing-based}). \S\ref{sec:dl-base-hp} provides a detailed review of representative deep learning-based human parsing studies. Frequently used datasets and performance comparisons are reviewed in \S\ref{sec:hp-data} and \S\ref{sec:pc}. An outlook for the future opportunities of human parsing is presented in \S\ref{sec:outlook}, including a new transformer-based baseline (\S\ref{sec:new-baseline}), under-investigated open issues (\S\ref{sec:open-issues}), new directions (\S\ref{sec:new-direc}), and human parsing in foundation models era (\S\ref{sec:hp-in-fme}) for future study. Conclusions will be drawn in \S\ref{sec:concl}.

\section{Preliminaries}
\label{sec:pre}

\subsection{Problem Formulation and Challenges}
\label{sec:problem}
Formally, we use \bm{$x$} to represent input human-centric data, \bm{$y$} to represent pixel-level supervision target, \bm{$\mathcal{X}$} and \bm{$\mathcal{Y}$} to denote the space of input data and supervision target. Human parsing is to map data \bm{$x$} to target \bm{$y$}: $\bm{\mathcal{X}}\mapsto\bm{\mathcal{Y}}$. The problem formulation is consistent with image segmentation \cite{minaee2021image}, but \bm{$\mathcal{X}$} is limited to the human-centric space. Therefore, in many literatures, human parsing is regarded as fine-grained image segmentation.



The central problem of human parsing is how to model human structures. As we all know, the human body presents a highly structured hierarchy, and all parts interact naturally. Most parsers hope to construct this interaction explicitly or implicitly. However, the following challenges make the problem more complicated:

\noindent$\bullet$~\textbf{Large Intra-class Variation.} In human parsing, objects with large visual appearance gaps may share the same semantic categories. For example, ``upper clothes" is an abstract concept without strict visual constraints. Many kinds of objects of color, texture, and shape belong to this category, leading to significant intra-class variations. Further challenges may be added by illumination changes, different viewpoints, noise corruption, low-image resolution, and filtering distortion. Large intra-class variations will increase the difficulty of classifier learning decision boundaries, resulting in semantic inconsistency in prediction.

\noindent$\bullet$~\textbf{Unconstrained Poses.} In the earlier human parsing benchmarks \cite{yamaguchi2012parsing, liu2013fashion, dong2013deformable, liang2015human}, the data is usually collected from fashion media. From them people often stand or have a limited number of simple pose. However, in the wild, human pose is unconstrained, showing great diversity. Therefore, more and more studies begin to pay attention to real-world human parsing. Unconstrained poses will increase the state space of target geometrically, which brings great challenges to the human semantic representations. Moreover, the left-right discrimination problem in human parsing is widespread (\emph{e.g}., left-arm \texttt{vs} right-arm, left-leg \texttt{vs} right-leg), and it is also severely affected by unconstrained poses \cite{liu2018cross, ruan2019devil, liu2019braidnet}.

\noindent$\bullet$~\textbf{Occlusion.} Occlusion mainly presents two modes: (1) occlusion between humans and objects; (2) occlusion between humans. The former will destroy the continuity of human parts or clothing, resulting in incomplete apparent information of the targets, forming local semantic loss, and easily causing ambiguity \cite{liang2015human, zhang2020pcnet}. The latter is a more severe challenge. In addition to continuity destruction, it often causes foreground confusion. In human parsing, only the occluded target human is regarded as the foreground, while the others are regarded as the background. However, they have similar appearance, making it difficult to determine which part belongs to the foreground \cite{yang2022part}.


\noindent\textbf{Remark.} In addition to the above challenges, some scenario-based challenges also hinder the progress of human parsing, such as the trade-off between inference efficiency and accuracy in crowded scenes, motion blur, and camera position changes in movement scenes.

\vspace{-6pt}
\subsection{Human Parsing Taxonomy}
\label{sec:taxonomy}
According to the characteristics (number of humans, data modal) of the input space \bm{$\mathcal{X}$}, human parsing can be categorized into three sub-tasks (see Figure~\ref{fig:taxonomy}): single human parsing, multiple human parsing, and video human parsing. 

\noindent$\bullet$~\textbf{Single Human Parsing (SHP).} SHP is the cornerstone of human parsing, which assumes that there is only one foreground human instance in the image. Therefore, \bm{$y$} just contains corresponding semantic category supervision at the pixel-level. Simple and straightforward task definitions make most related research focus on how to model robust and generalized human parts relationship. In addition to being the cornerstone of human parsing, SHP is also often used as an auxiliary supervision for some tasks, \emph{e.g}., person re-identification, human mesh reconstruction,  virtual try-on.

\noindent$\bullet$~\textbf{Multiple Human Parsing (MHP).} Multiple human parsing, also known as instance-level human parsing, aims to parse multiple human instances in a single pass. Besides category information, \bm{$y$} also provides instance supervision in pixel-level, \emph{i.e}., the person identity of each pixel. The core problems of MHP are how to discriminate different human instances and how to learn each human feature in crowded scenes comprehensively. In addition, inference efficiency is also an important concern of MHP. Ideally, inference should be real-time and independent of human instance numbers. Except as an independent task, MHP sometimes is jointed with other human visual understanding tasks in a multi-task learning manner, such as pose estimation \cite{zhou2021differentiable, liu2021multi}, dense pose \cite{yang2019parsing} or panoptic segmentation \cite{de2021part}.

\noindent$\bullet$~\textbf{Video Human Parsing (VHP).} VHP needs to parse every human in the video data, which can be regarded as a complex visual task integrating video segmentation and image-level human parsing. The current VHP studies mainly adopt the unsupervised video object segmentation settings \cite{wang2021survey}, \emph{i.e}., \bm{$y$} is unknown in the training stage, and the ground-truth of the first frame is given in the inference stage. The temporal correspondence will only be approximated according to \bm{$x$}. Relative to SHP and MHP, VHP faces more challenges that are inevitable in video segmentation settings, \emph{e.g}., motion blur and camera position changes. Benefitting by the gradual popularity of video data, VHP has a wide range of application potential, and the typical cases are intelligent monitoring and video editing.


\noindent\textbf{Remark.} Over recent years, some potential research directions have also received attention, including weakly-supervised human parsing \cite{fang2018weakly, li2020self, zhao2022from}, one-shot human parsing \cite{he2021progressive, he2023end} and interactive human parsing \cite{gao2022clicking, gao2023dynamic}.

\vspace{-6pt}
\subsection{Relevant Tasks}
\label{sec:relevant}
Among the research in computer vision, there are some tasks with strong relevance to human parsing, which are briefly described in the following.

\noindent$\bullet$~\textbf{Pose Estimation.} The purpose of pose estimation \cite{wei2016convolutional, xiao2018simple, zheng2023deep} is to locate human parts and build body representations (such as skeletons) from input data. 
Human parsing and pose estimation share the same input space \bm{$\mathcal{X}$}, but there are some differences in the supervision targets. The most crucial difference is that human parsing is a dense prediction task, which needs to predict the category of each pixel. Meanwhile, pose estimation is a sparse prediction task, only focusing on the location of a limited number of keypoints. These two tasks are also often presented in multi-task learning, or one of them is used as a guiding condition for the other. For example, human parsing as a guide can help pose estimation to reduce the impact of clothing on human appearance \cite{ladicky2013human}.

\noindent$\bullet$~\textbf{Image Segmentation.} Image segmentation \cite{shelhamer2016fully, zhao2017pyramid, minaee2021image} is a fundamental topic in image processing and computer vision. It mainly includes semantic segmentation and instance segmentation.
As a basic visual task, there are many research directions can be regarded as branches, and human parsing is one of them. In the pre-deep learning era, image segmentation focuses on the continuity of color, texture, and edge, while human parsing pays more attention to the body topology modeling. In the deep learning era, the methods in two fields show more similarities. However, more and more human parsing literature choose to model the parts relationship as the goal, which is significantly different from the general goal of image segmentation. Therefore, human parsing and image segmentation are closely related but independent problems.

\noindent\textbf{Remark.} Ordinarily, most human-centric dense prediction task show positively relevance with human parsing, \emph{e.g}., human matting \cite{chen2018semantic, liu2020boosting}, human mesh reconstruction \cite{guler2019holopose, zheng2019deephuman} and face/hand parsing \cite{liang2014parsing, lin2019face}.

\begin{figure*}
	\begin{center}
		\includegraphics[width=0.98\linewidth]{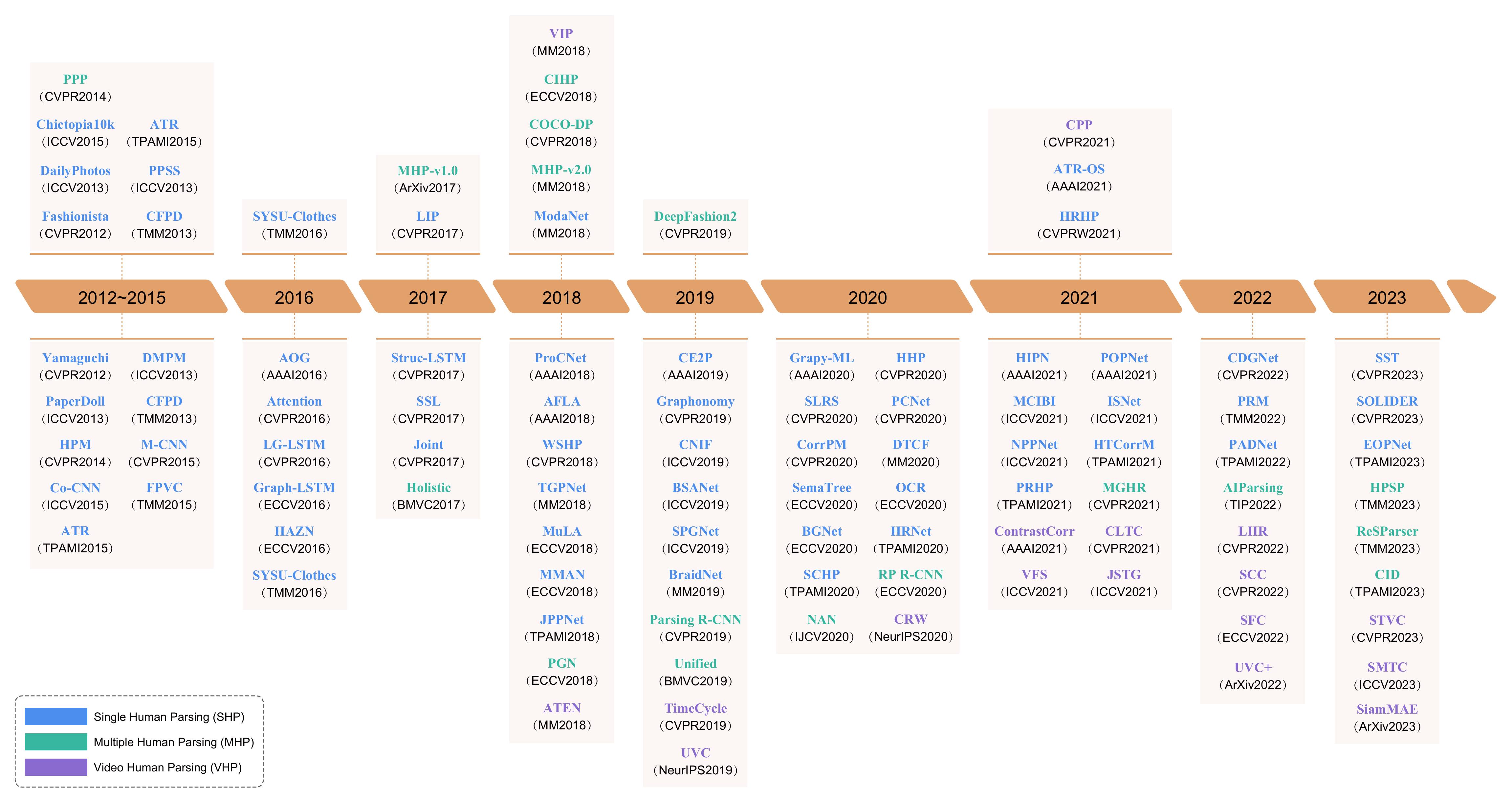}
	\end{center}
	\vspace{-5pt}
	\captionsetup{font=small}
	\caption{\small{\textbf{Timeline of representative human parsing works from 2012 to 2023.} The upper part represents the datasets of human parsing (\S\ref{sec:hp-data}), and the lower part represents the models of human parsing (\S\ref{sec:dl-base-hp}).}}
	\label{fig:timeline}
	\vspace{-10pt}
\end{figure*}

\vspace{-6pt}
\subsection{Applications of Human Parsing}
\label{sec:parsing-based}
As a crucial task in computer vision, there are a large number of applications based on human parsing. We will introduce some common ones below.

\noindent$\bullet$~\textbf{Dense Pose Estimation.} The goal of dense pose estimation is to map all human pixels in an RGB image to the 3D surface of the human body \cite{guler2018densepose}. Human parsing is an important pre-condition that can constrain the mapping of dense points. At present, the mainstream dense pose estimation methods explicitly integrate human parsing supervision, such as DensePose R-CNN \cite{guler2018densepose}, Parsing R-CNN \cite{yang2019parsing}, 
and SimPose \cite{zhu2020simpose}. Therefore, the performance of human parsing will directly affect dense pose estimation results.

\noindent$\bullet$~\textbf{Person Re-identification.} Person re-identification seeks to predict whether two images from different cameras belong to the same person. The apparent characteristics of human body is an important factor affecting the accuracy. Human parsing can provide pixel-level semantic information, helping re-identification models perceive the position and composition of human parts/clothing. Various studies have introduced human parsing explicitly or implicitly into re-identification methods, which improves the model performance in multiple aspects, \emph{e.g}., local visual cues \cite{kalayeh2018human, yang2019towards}, spatial alignment \cite{sun2019learning, huang2020improve, li2021person}, background-bias elimination \cite{tian2018elimiating}, domain adaptation \cite{chen2019instance}, clothes changing \cite{yu2020cocas, qian2020long}.

\noindent$\bullet$~\textbf{Virtual Try-on.} Virtual try-on is a burgeoning and interesting application in the vision and graphic communities \cite{han2018viton, wang2018toward, yu2019vtnfp, wu2019m2e, dong2019towards, liu2021toward, xie2021was, zhao2021m3d}. Most of the research follows the three processes: human parsing, appearance generation, and refinement. Therefore, human parsing is a necessary step to obtain clothing masks, appearance constraints and pose maintenance. Recently, some work began to study the parser-free virtual try-on \cite{issenhuth2020do, chang2022pfvton, lin2022rmgn}. Through teacher-student learning, parsing-based pre-training, and other technologies, the virtual try-on can be realized without the human parsing map during inference. However, most works still introduced the parsing results during training, and the generation quality retains gap from parser-based methods.

\noindent$\bullet$~\textbf{Conditional Human Image Generation.} Image generation/synthesis as a field has seen a lot of progress in recent years \cite{goodfellow2014generative, karras2019style, niemeyer2021giraffe, nichol2021glide}. Non-existent but fidelity images can be created in large quantities. Among them, human image generation has attracted attention because of its rich downstream applications. Compared with unconditional generation, conditional generation can produce corresponding output as needed, and human parsing map is one of the most widely used pre-conditions. There have been a lot of excellent works on parsing-based conditional human image generation, \emph{e.g}., CPFNet \cite{wu2021image}, InsetGAN \cite{fruhstuck2022insetgan}, ControlNet \cite{zhang2023adding} and Composer \cite{huang2023composer}.

\noindent$\bullet$~\textbf{VR / AR.} Virtual Reality (VR) and Augmented Reality (AR) technologies are currently receiving a great deal of attention \cite{schuemie2001research, chen2023virtual, wu2023virtual}, thanks in large part to the commercial availability of new immersive platforms. Human parsing is a crucial visual technology in VR / AR, which can help the system accurately locate human parts, recognize gestures, recognize clothing and actions. Some work has extensively attempted human parsing technology in fields such as interactive games, clothing shopping, and immersive education.

\noindent\textbf{Remark.} Besides the above cases, in general, most of the human-centric generation applications can be built with the help of human parsing, \emph{e.g}., deepfakes \cite{chen2020simswap, yang2021attacks}, style transfer \cite{liu2019swapgan, huo2021manifold, ma2022dual}, clothing editing \cite{kim2019style, ntavelis2020sesame, tseng2020modeling}.

\section{Deep Learning Based Human Parsing}
\label{sec:dl-base-hp}
The existing human parsing can be categorized into three sub-tasks: single human parsing, multiple human parsing, and video human parsing, focusing on parts relationship modeling, human instance discrimination, and temporal correspondence learning, respectively. According to this taxonomy, we sort out the representative works (lower part of Figure~\ref{fig:timeline}) and review them in detail below.

\vspace{-6pt}
\subsection{Single Human Parsing (SHP) Models}
\label{sec:shp-methods}
SHP considers extracting human features through parts relationship modeling. According to the modeling strategy, SHP models can be divided into three main classes: context learning, structured representation, and multi-task learning. Moreover, considering some special but interesting methods, we will review them as ``other modeling models". Table~\ref{table:shp_taxonomy} summarizes the characteristics for reviewed SHP models.

\begin{table*}[htbp]
	\centering
	\captionsetup{font=small}
	\caption{\small{\textbf{Summary of essential characteristics for reviewed SHP models} (\S\ref{sec:shp-methods}). The training datasets and whether it is open source are also listed. See \S\ref{sec:hp-data} for more detailed descriptions of datasets. These notes also apply to the other tables.}}
	\label{table:shp_taxonomy}
	\vspace{-5pt}
	\begin{threeparttable}
	\resizebox{0.95\textwidth}{!}{
	\setlength\tabcolsep{2.5pt}
	\renewcommand\arraystretch{1.02}
	\begin{tabular}{c|r|c||cc|cc|cc|cc||c||c}
	\hline\thickhline
	\rowcolor{mygray1}
	                                  &                                        &                                    &  \multicolumn{2}{c|}{Context} & \multicolumn{2}{c|}{Structured} & \multicolumn{2}{c|}{Multi-task} &          \multicolumn{2}{c||}{Others}   &   & \\
	\cline{4-5} \cline{6-7} \cline{8-9} \cline{10-11}                                 
	\rowcolor{mygray1}                                  
	\multirow{-2}{*}{Year} & \multirow{-2}{*}{Method} & \multirow{-2}{*}{Pub.} &   Attention   &Scale-aware&      Tree     &     Graph      &      Edge      &      Pose      &  Denoising &  Adversarial  & \multirow{-2}{*}{Datasets}   &  \multirow{-2}{*}{Open Source}  \\
	\hline
	\hline
	\multirow{1}{*}{{2012}} & Yamaguchi \cite{yamaguchi2012parsing} &  CVPR  &    -         &       -             &      -              &           -         &        -            &\checkmark   &       -             &        -            & FS    & - \\            
	\hline
	\multirow{3}{*}{{2013}} & DMPM \cite{dong2013deformable} &  ICCV       &       -             &      -              &\checkmark  &       -             &         -           &       -             &        -            &      -              & FS/DP & - \\
	                                    & PaperDoll \cite{yamaguchi2013paper}    & ICCV&     -               &       -             &      -              &        -            &      -              & \checkmark  &    -                &     -               & FS & - \\
	                                    & CFPD \cite{liu2013fashion}    &     TMM             &       -             &       -             &      -              &         -           &       -             & \checkmark  &     -               &     -               & CFPD  & -\\
	\hline
	\multirow{1}{*}{{2014}} & HPM \cite{dong2014towards} &  CVPR              &      -              &      -              & \checkmark &          -          &          -          & \checkmark &     -               &     -               & FS/DP  & -\\
	\hline
	\multirow{4}{*}{{2015}} & M-CNN \cite{liu2015matching} &  CVPR            &        -            &       -              &        -            & \checkmark &       -             &       -             &  -                  &     -              & ATR & - \\
	                                    & Co-CNN \cite{liang2015human} &  ICCV            &       -             & \checkmark &          -          &           -         &         -           &         -           &         -           &        -            & FS/ATR & - \\
	                                    & FPVC \cite{liu2015fashion} &  TMM                    &      -              &       -             &       -              &         -           &       -             & \checkmark &       -             &    -                & FS/DP & - \\
	                                    & ATR \cite{liang2015deep} &  TPAMI                   &        -            &        -            & \checkmark &         -           &       -             &          -            &      -              &      -              & FS/DP& -  \\  
	\hline
	\multirow{6}{*}{{2016}} & AOG \cite{xia2016pose} &      AAAI                     &       -              &      -              & \checkmark &         -          &   -                 & \checkmark  &    -                &     -               & -    & - \\
	                                    & Attention \cite{chen2016attenttion} &  CVPR      & \checkmark &\checkmark   &         -            &          -          &    -                &    -                &     -               &     -               & PPP   & \checkmark\\     
	                                    & LG-LSTM \cite{liang2016object} &      CVPR      &\checkmark  &         -           &         -            &         -          &    -                &     -                 &    -                &   -                  &FS/ATR/PPP & - \\                        
	                                    & Graph-LSTM \cite{liang2016semantic} & ECCV  &\checkmark &          -          &         -            & \checkmark&       -             &       -               &      -              &      -              & FS/ATR/PPP  & -\\     
	                                    & HAZN \cite{xiq2016zoom} & ECCV                     &     -               & \checkmark &          -           &        -           &      -              &     -                 &    -                &     -               & PPP  & -\\  
	                                    & SYSU-Clothes \cite{liang2016clothes} &  TMM   &     -               &       -             &         -            &\checkmark &       -             &     -                 &     -               &   -                 &  SYSU-Clothes  & -\\  
	\hline
	\multirow{3}{*}{{2017}} & Struc-LSTM \cite{liang2017interpretable} &CVPR&\checkmark&         -           & \checkmark & \checkmark&      -              &         -             &        -            &        -            & ATR/PPP& -  \\ 
	                                    & SSL \cite{gong2017look}  &            CVPR           &     -               &       -             &         -             &    -               &-                    & \checkmark  &    -                &    -                & LIP/PPP  & \checkmark\\ 
	                                    & Joint \cite{xia2017joint}    &            CVPR           &      -              &         -           &          -            &     -              &   -                 & \checkmark  &     -               &      -              & PPP & - \\ 
	\hline
	\multirow{7}{*}{{2018}} & ProCNet \cite{zhu2018progressive} &   AAAI      &      -               &        -           & \checkmark  &         -          &       -              &        -            &      -              &     -               & PPP & - \\
	                                    & AFLA \cite{liu2018cross} &               AAAI           &       -              &         -          &         -             &           -        &    -                 &      -              &    -                &\checkmark & LIP  & -\\
	                                    & WSHP \cite{fang2018weakly} &  CVPR              &        -              &        -            &        -             &        -            &   -                 & \checkmark&      -              &      -               &PPP & - \\ 
	                                    & TGPNet \cite{luo2018trusted} & MM                   &        -              & \checkmark &           -          &           -        &      -              &        -            &      -              &     -               & ATR & \checkmark \\            
	                                    & MuLA \cite{nie2018mutual}   &          ECCV        &       -             &         -           &      -                &     -              &     -               & \checkmark  &   -                 &       -             & LIP/PPP & - \\ 
	                                    & MMAN \cite{luo2018macro}  &          ECCV        &       -             &         -           &       -               &       -            &     -               &                     &      -              & \checkmark& LIP/PPP/PPSS & \checkmark  \\ 
	                                    & JPPNet \cite{liang2018look} &   TPAMI               &       -             &         -           &       -               &         -          &      -              & \checkmark &      -              &        -             &LIP/PPP  & \checkmark \\ 
	\hline
	\multirow{8}{*}{{2019}} &   CE2P \cite{ruan2019devil}  &          AAAI         &        -            & \checkmark &         -            &         -           & \checkmark &       -              &-                    &      -              & LIP  & \checkmark\\ 
	                                    & Graphonomy \cite{gong2019graphonomy} &CVPR&   -             &       -             & \checkmark &\checkmark &       -             &           -           &   -                 &         -           & ATR/PPP & \checkmark  \\ 
	                                    & CNIF \cite{wang2019learning}  &       ICCV        &       -             &        -            & \checkmark  &         -          &      -               &          -           &   -                 &         -           & ATR/LIP/PPP & \checkmark \\                         
	                                    & BSANet \cite{zhao2019multi} &        ICCV          &       -              & \checkmark &       -              &        -           & \checkmark &              -      &        -            &              -       &   PPP & -\\               
	                                    & CCNet \cite{huang2023ccnet} &\tabincell{c}{ICCV\\ TPAMI}          &  \checkmark  & -                   &       -              &        -           & \checkmark &              -      &        -            &              -       &   LIP & \checkmark\\                                      
	                                    & SPGNet \cite{cheng2019spgnet} &     ICCV       & \checkmark  &           -          &       -              &          -         &        -            &              -      &       -             &        -             &   PPP & -\\    
	                                    & BraidNet \cite{liu2019braidnet}     &       MM       &        -             & \checkmark &          -           &             -       &        -            &                -    &         -           &     -             & LIP  & -\\  
	\hline
	\multirow{11}{*}{{2020}} & Grapy-ML \cite{he2020grapyml} &     AAAI      & \checkmark  &        -             & \checkmark &\checkmark &  -                  &          -           &         -           &        -             & ATR/PPP & \checkmark  \\  
	                                    & HHP \cite{wang2020hierarchical}  &     CVPR    &        -            &      -              & \checkmark  &\checkmark &   -                  &       -              &        -            &       -             & ATR/LIP/PPP/PPSS & \checkmark \\
	                                    & SLRS \cite{li2020self}     &            CVPR           &         -            &      -              &        -              &\checkmark& \checkmark &        -              & \checkmark &        -             & ATR/LIP  & - \\ 
	                                    & PCNet \cite{zhang2020pcnet}     &    CVPR      &         -            & \checkmark &          -            &\checkmark &       -             &         -             &            -        &       -              & LIP/PPP   & -\\  
	                                    & CorrPM \cite{zhang2020correlating}  &  CVPR   &       -            &       -             &       -               &      -              & \checkmark & \checkmark  &         -           &      -              & ATR/LIP  & \checkmark \\   
	                                    & DTCF \cite{liu2020hybrid} &           MM              &       -              & \checkmark &         -            &        -           & \checkmark &         -             &          -          &       -              &  LIP/PPP & - \\    
	                                    & SemaTree \cite{ji2020learning} &    ECCV         & \checkmark &       -             & \checkmark  &         -          &        -             &              -        &        -            &         -            &   LIP & \checkmark  \\                   
	                                    & OCR \cite{yuan2020object} &        ECCV           & \checkmark & \checkmark &            -         &        -           &        -             &           -           &        -            &         -            &   LIP  & \checkmark \\          
	                                    & BGNet \cite{zhang2020blended} &       ECCV    &         -            &        -            & \checkmark  &\checkmark&      -               &             -         &        -            &          -           &   LIP/PPP/PPSS  & -\\        
	                                    & HRNet \cite{wang2020deep}  &       TPAMI        &        -             & \checkmark &          -           &       -            &         -            &         -             &          -          &          -           &  LIP  & \checkmark\\        
	                                    & SCHP \cite{li2020correction}    &     TPAMI        &       -              &          -          &        -             &         -           &\checkmark  &         -             & \checkmark &          -           & \tabincell{c}{ATR/LIP/PPP}  & \checkmark \\ 
        \hline
	\multirow{7}{*}{{2021}} & HIPN \cite{liu2021hier} &     AAAI                    &        -              &       -            &         -             &         -          &         -           &          -             & \checkmark &       -              & LIP/PPP & - \\  
	                                    & POPNet \cite{he2021progressive}  &  AAAI       &\checkmark   &        -            &         -             &         -           &        -            &        -             &       -             &       -             & ATR-OS & \checkmark  \\
	                                    & MCIBI \cite{jin2021mining}     &        ICCV         &\checkmark   &        -             &        -             &        -            &         -          &         -             &        -            &       -              & LIP & \checkmark  \\ 
	                                    & ISNet \cite{jin2021isnet}     &          ICCV           &\checkmark   &        -             &         -            &        -            &        -           &         -             &        -            &      -               & LIP & \checkmark  \\ 
	                                    & NPPNet \cite{zeng2021neural}     &    ICCV      &         -            & \checkmark  &         -            &        -            &      -             & \checkmark   &           -         &        -             & LIP/PPP  & \checkmark \\ 
	                                    & HTCorrM \cite{zhang2021on}  &        TPAMI     & \checkmark &         -            &            -         &           -        & \checkmark & \checkmark    &         -           &          -           &   ATR/LIP & - \\   
	                                    & PRHP \cite{wang2021hierarchical}  &     TPAMI&        -            &       -             & \checkmark  &\checkmark &         -            &         -              &      -              &        -            & ATR/LIP/PPP/PPSS & \checkmark \\
	\hline
	\multirow{4}{*}{{2022}} & CDGNet \cite{liu2022cdgnet}    &  CVPR            & \checkmark  &        -            &           -           &        -            & \checkmark &          -         &            -        &     -               & ATR/LIP  & \checkmark \\       
	                                    & HSSN \cite{li2022deep}             &  CVPR            &         -            &  \checkmark &\checkmark   &        -            &       -             &          -         &            -        &     -               & LIP/PPP  & \checkmark \\     
	                                    & PRM \cite{zhang2022human} &  TMM                &          -          &        -            &           -           &\checkmark &         -           &          -            &         -           &             -       & LIP/PPP& -  \\
	                                    & PADNet \cite{zhao2022from}  &        TPAMI     &           -           &        -             &          -           &          -         &     -                 & \checkmark  &   -                 &        -             &   PPP & - \\                                                                     
	\hline
	\multirow{3}{*}{{2023}} & SST \cite{yang2023semantic}  &  CVPR              &         -           & \checkmark   &          -            &      -              &         -           &      -               &      -             &       -             & ATR/PPP  & \checkmark \\  
	                                     & SOLIDER \cite{chen2023beyond}  &  CVPR       &\checkmark   &        -            &          -            &      -              &         -           &      -               &      -             &       -             & LIP & \checkmark \\  
	                                     & EOPNet \cite{he2023end}  &  TPAMI                   &\checkmark   &        -            &          -            &      -              &         -           &      -               &      -             &       -             & ATR-OS  & \checkmark \\ 
	                                                           
	\hline                                    
	\end{tabular}
	}
	\end{threeparttable}
	\vspace{-5pt}
\end{table*}

\vspace{-6pt}
\subsubsection{Context Learning}
Context learning, a mainstream paradigm for single human parsing, seeks to learn the connection between local and global features to model human parts relationship. Recent studies have developed various context learning methods to handle single human parsing, including attention mechanism and scale-aware features. 

\noindent$\bullet$~\textbf{Attention Mechanism.} The first initiative was proposed in \cite{chen2016attenttion} that applies an attention mechanism for parts relationship modeling. Specifically, soft weights, learned by attention mechanism, are used to weight different scale features and merge them. At almost the same time, LG-LSTM \cite{liang2016object}, Graph-LSTM \cite{liang2016semantic} and Struc-LSTM \cite{liang2017interpretable} exploit complex local and global context information through Long Short-Term Memory (LSTM) \cite{hochreiter1997long} and achieve very competitive results. Then, \cite{cheng2019spgnet} proposes a Semantic Prediction Guidance (SPG) module that learns to re-weight the local features through the guidance from pixel-wise semantic prediction. With the rise of graph model, researchers realized that attention mechanism is able to establish the correlation between graph model nodes. For example, \cite{he2020grapyml} introduces Graph Pyramid Mutual Learning (Grapy-ML) to address the cross-dataset human parsing problem, in which the self-attention is used to model the correlations between context nodes. Although attention mechanisms have achieved great results in previous work, global context dependency cannot be fully understood due to the lack of explicit prior supervision. CDGNet \cite{liu2022cdgnet} adopts the human parsing labels accumulated in the horizontal and vertical directions as the supervisions, aiming to learn the position distribution of human parts, and weighting them to the global features through attention mechanism to achieve accurate parts relationship modeling. POPNet \cite{he2021progressive} and EOPNet \cite{he2023end} combine attention mechanism with metric learning to attempt solving the one-shot human parsing issue, providing a new solution for fashion applications without predefined human part categories.

\noindent$\bullet$~\textbf{Scale-aware Features.} The most intuitive context learning method is to directly use scale-aware features (\emph{e.g}. multi-scale features \cite{zhao2017pyramid, chen2017deeplab}, features pyramid networks \cite{lin2017feature, kirillov2019pfpn}), which has been widely verified in semantic segmentation \cite{minaee2021image}. The earliest effort can be tracked back to CoCNN \cite{liang2015human}. It integrates cross layer context, global image-level context, super-pixel context, and cross super-pixel neighborhood context into a unified architecture, which solves the obstacle of low-resolution features in FCN \cite{shelhamer2016fully} for modeling parts relationship. Subsequently, \cite{xiq2016zoom} proposes Hierarchical Auto-Zoom Net (HAZN), which adaptively zooms predicted image regions into their proper scales to refine the parsing. TGPNet \cite{luo2018trusted} considers that the label fragmentation and complex annotation in human parsing datasets is a non-negligible problem to hinder accurate parts relationship modeling, 
trying to alleviate this limitation by supervising multi-scale context information. PCNet \cite{zhang2020pcnet} further studies the adaptive contextual features, and captures the representative global context by mining the associated semantics of human parts through proposed part class module, relational aggregation module, and relational dispersion module. 

\vspace{-6pt}
\subsubsection{Structured Representation}
The purpose of structured representation is to learn the inherent combination or decomposition mode of human parts, so as to model parts relationship. Research efforts in this field are mainly made along two directions: using a tree structure to represent the hierarchical relationship between body and parts, and using a graph structure to represent the connectivity relationship between different parts. These two ideas are complementary to each other, so they have often been adopted simultaneously in some recent work.

\noindent$\bullet$~\textbf{Tree Structure.} DMPM \cite{dong2013deformable} and HPM \cite{dong2014towards} solve the single human parsing issue by using the parselets representation, which construct a group of parsable segments by low-level over-segmentation algorithms, and represent these segments as leaf nodes, then search for the best graph configuration to obtain semantic human parsing results. Similarly, \cite{liang2015deep} formulates human parsing as an Active Template Regression (ATR) problem, where each human part is represented as the linear combination of learned mask templates and morphed to a more precise mask with the active shape parameters. Then the human parsing results are generated from the mask template coefficients and the active shape parameters. In the same line of work, ProCNet \cite{zhu2018progressive} deals with human parsing as a progressive recognition task, modeling structured parts relationship by locating the whole body and then segmenting hierarchical components gradually. CNIF \cite{wang2019learning} further extends the human tree structure and represents human body as a hierarchy of multi-level semantic parts, treating human parsing as a multi-source information fusion process. A more efficient solution is developed in \cite{ji2020learning}, which uses a tree structure to encode human physiological composition, then designs a coarse to fine process in a cascade manner to generate accurate parsing results. 

\noindent$\bullet$~\textbf{Graph Structure.} Graph structure is an excellent relationship modeling method. Some researchers consider introducing it into human parsing networks for part-relation reasoning. A clothing co-parsing system is designed by \cite{liang2016clothes}, which takes the segmented regions as the vertices. It incorporates several contexts of clothing configuration to build a multi-image graphical model. To address the cross-dataset human parsing problem, Graphonomy \cite{gong2019graphonomy} proposes a universal human parsing agent, introducing hierarchical graph transfer learning to encode the underlying label semantic elements and propagate relevant semantic information. BGNet \cite{zhang2020blended} hopes to improve the accuracy of human parsing in similar or cluttered scenes through graph structure. It exploits the human inherent hierarchical structure and the relationship between different human parts employing grammar rules in both cascaded and paralleled manner to correct the segmentation performance of easily confused human parts. A landmark work on this line was proposed by Wang \emph{et al}.\cite{wang2020hierarchical, wang2021hierarchical}. A hierarchical human parser (HHP) is constructed, representing the hierarchical human structure by three kinds of part relations: decomposition, composition, and dependency. Besides, HHP uses the prism of a message-passing, feed-back inference scheme to reason the human structure effectively. Following this idea, \cite{zhang2022human} proposes Part-aware Relation Modeling (PRM) to handle human parsing, generating features with adaptive context for various sizes and shapes of human parts.

\vspace{-6pt}
\subsubsection{Multi-task Learning}
The auxiliary supervisions can help the parser better understand the relationship between parts, such as part edges or human pose. Therefore, multi-task learning has become an essential paradigm for single human parsing.

\noindent$\bullet$~\textbf{Edge-aware Learning.} Edge information is implicit in the human parsing dataset. Thus edge-aware supervision or feature can be introduced into the human parser without additional labeling costs. In particular, edge-aware learning can enhance the model's ability to discriminate adjacent parts and improve the fineness of part boundaries. The typical work is \cite{ruan2019devil}, which proposes a Context Embedding with Edge Perceiving (CE2P) framework, using an edge perceiving module to integrate the characteristic of object contour to refine the part boundaries. Because of its excellent performance and scalability, CE2P has become the baseline for many subsequent works. CorrPM \cite{zhang2020correlating} and HTCorrM \cite{zhang2021on}  are built on CE2P, and further use part edges to help model the parts relationship. They construct a heterogeneous non-local module to mix the edge, pose and semantic features into a hybrid representation, and explore the spatial affinity between the hybrid representation and the parsing feature map at all positions. BSANet \cite{zhao2019multi} considers that edge information is helpful to eliminate the part-level ambiguities and proposes a joint parsing framework with boundary and semantic awareness to address this issue. Specifically, a boundary-aware module is employed to make intermediate-level features focus on part boundaries for accurate localization, which is then fused with high-level features for efficient part recognition. To further enrich the edge-aware features, a dual-task cascaded framework (DTCF) is developed in \cite{liu2020hybrid}, which implicitly integrates parsing and edge features to refine the human parsing results progressively. 

\noindent$\bullet$~\textbf{Pose-aware Learning.} Both human parsing and pose estimation seek to predict dense and structured human representation. There is a high intrinsic relationship between them. Therefore, some studies have tried to use pose-aware learning to assist in parts relationship modeling. As early as 2012, Yamaguchi \emph{et al}. \cite{yamaguchi2012parsing, yamaguchi2013paper} exploited the relationship between clothing and the underlying body pose, exploring techniques to accurately parse person wearing clothing into their constituent garment pieces. Almost immediately, Liu \emph{et al}. \cite{liu2013fashion} combined the human pose estimation module with an MRF-based color/category inference module and a super-pixel category classifier module to parse fashion items in images. Subsequently, Liu \emph{et al}. \cite{liu2015fashion} extends this idea to semi-supervised human parsing, collecting a large number of unlabeled videos, using cross-frame context for human pose co-estimation, and then performs video joint human parsing. SSL \cite{gong2017look} and JPPNet \cite{liang2018look} choose to impose human pose structures into parsing results without resorting to extra supervision, and adopt the multi-task learning manner to explore efficient human parts relationship modeling. A similar work is developed by \cite{nie2018mutual}, which presents a Mutual Learning to Adapt model (MuLA) for joint human parsing and pose estimation. MuLA can fast adjust the parsing and pose models to provide more robust and accurate results by incorporating information from corresponding models. Different from the above work, Zeng \emph{et al}. \cite{zeng2021neural}. focus on how to automatically design a unified model and perform two tasks simultaneously to benefit each other. Inspired by NAS \cite{fang2020densely}, they propose to search for an efficient network architecture (NPPNet), searching the encoder-decoder architectures respectively, and embed NAS units in both multi-scale feature interaction and high-level feature fusion. To get rid of annotating pixel-wise human parts masks, a weakly-supervised human parsing approach is proposed by PADNet \cite{zhao2022from}. They develop an iterative training framework to transform pose knowledge into part priors, so that only pose annotations are required during training, greatly alleviating the annotation burdens.

\begin{table}
	\centering
	\captionsetup{font=small}
	\caption{\small{\textbf{Highlights of parts relationship modeling methods for SHP models} (\S\ref{sec:shp-methods}). Representative Works of each method are also give.}}
	\label{table:shp_highlights}
	\vspace{-5pt}
	\begin{threeparttable}
	\resizebox{0.49\textwidth}{!}{
	\setlength\tabcolsep{2.5pt}
	\renewcommand\arraystretch{1.02}
	\begin{tabular}{c|c|c}
	\hline\thickhline
	\rowcolor{mygray1}
	Method                & \tabincell{c}{Representative\\ Works} & Highlights \\
	\hline
	\hline
         Attention              & \tabincell{c}{\cite{chen2016attenttion, liang2016object, liang2016semantic}\\ \cite{liang2017interpretable, cheng2019spgnet, he2020grapyml}\\ \cite{liu2022cdgnet, huang2023ccnet, chen2023beyond} \\\cite{he2023end}}    & \tabincell{c}{It is helpful to locate interested human parts,\\ suppress useless background information.} \\
         \hline
	Scale-aware         & \tabincell{c}{\cite{liang2015human, xiq2016zoom, luo2018trusted}\\ \cite{zhang2020pcnet, yang2023semantic}}   & \tabincell{c}{Fusion low-level texture and high-level semantic\\ features, help to parse small human parts.} \\
	\hline
	\hline
	Tree                     & \tabincell{c}{\cite{dong2013deformable, dong2014towards, liang2015deep}\\ \cite{zhu2018progressive, wang2019learning, ji2020learning}}    & \tabincell{c}{Simulate the composition and decomposition\\ relationship between human parts and body.} \\
         \hline
	Graph                  & \tabincell{c}{\cite{liang2016clothes, gong2019graphonomy, zhang2020blended}\\ \cite{wang2020hierarchical, wang2021hierarchical, zhang2022human}}   & \tabincell{c}{Modeling the correlation and difference between\\ human parts.} \\
	\hline
	\hline
	Edge                     & \tabincell{c}{\cite{ruan2019devil, zhang2020correlating, zhang2021on}\\ \cite{zhao2019multi, liu2020hybrid}}  & \tabincell{c}{Solve the pixel confusion problem on the boundary\\ of adjacent parts, generating finer boundary.} \\
         \hline
	Pose                     & \tabincell{c}{\cite{yamaguchi2012parsing, yamaguchi2013paper, liu2013fashion}\\ \cite{liu2015fashion, gong2017look, liang2018look}\\ \cite{nie2018mutual, zeng2021neural, zhao2022from}}   & \tabincell{c}{As context clues to improve semantic consistency\\ between parsing results and body structure.} \\
	\hline
	\hline
	Denoising             & \tabincell{c}{\cite{li2020self, li2020correction, liu2021hier}}   & \tabincell{c}{Alleviate the impact of super-pixel or annotation\\ errors, improving the robustness.} \\
         \hline
	Adversarial           & \tabincell{c}{\cite{liu2018cross, luo2018macro}}   & \tabincell{c}{Reduce the domain differences between training\\ data and testing data, improving the generalization.} \\
         \hline
	\end{tabular}
	}
	\end{threeparttable}
	\vspace{-5pt}
\end{table}

\vspace{-6pt}
\subsubsection{Other Modeling Models}
Other works attempt to employ techniques outside of the above taxonomy, such as denoising and adversarial learning, which also make specific contributions to the human parts relationship modeling and deserve a separate look.

\noindent$\bullet$~\textbf{Denoising.} To reduce the labeling cost, there is a large amount of noise in the mainstream SHP datasets \cite{liang2015deep, gong2017look}, so denoising learning for accurate human parts relationship modeling has also received some attention. SCHP \cite{li2020correction} is the most representative work. It starts with using inaccurate parsing labels as the initialization and designs a cyclically learning scheduler to infer more reliable pseudo labels
In the same period, Li \emph{et al}. \cite{li2020self} attempt to combine denoising learning and semi-supervised learning, proposing Self-Learning with Rectification (SLR) strategy for human parsing. SLR generates pseudo labels for unlabeled data to retrain the parsing model and introduces a trainable graph reasoning method to correct typical errors in pseudo labels. Based on SLR, HIPN \cite{liu2021hier} further explores to combine denoising learning with semi-supervised learning, which develops the noise-tolerant hybrid learning, taking advantage of positive and negative learning to better handle noisy pseudo labels.

\noindent$\bullet$~\textbf{Adversarial Learning.} Earlier, inspired by the Generative Adversarial Nets (GAN) \cite{goodfellow2014generative}, a few works use adversarial learning to solve problems in parts relationship modeling. For example, to solve the domain adaptation problem, AFLA \cite{liu2018cross} proposes a cross-domain human parsing network, introducing a discriminative feature adversarial network and a structured label adversarial network to eliminate cross-domain differences in visual appearance and environment conditions. MMAN \cite{luo2018macro} hopes to solve the problem of low-level local and high-level semantic inconsistency in pixel-wise classification loss. It contains two discriminators: Macro D, acting on low-resolution label map and penalizing semantic inconsistency; Micro D, focusing on high-resolution label map and restraining local inconsistency.

\noindent\textbf{Remark.} In fact, many single human parsing models use a variety of parts relationship modeling methods. Therefore, our above taxonomy only introduces the core methods of each model. Table~\ref{table:shp_highlights} summarizes the highlights of each parts relationship modeling method.

\vspace{-6pt}
\subsection{Multiple Human Parsing (MHP) Models}
\label{sec:mhp-methods}
MHP seeks to locate and parse each human in the image plane. The task setting is similar to instance segmentation, so it is also called instance-level human parsing. We divide MHP into three paradigms: bottom-up, one-stage top-down, and two-stage top-down, according to its pipeline of discriminating human instances. The essential characteristics of reviewed MHP models are illustrated in Table~\ref{table:mhp_taxonomy}.

\begin{table}
	\centering
	\captionsetup{font=small}
	\caption{\small{\textbf{Summary of essential characteristics for reviewed MHP models} (\S\ref{sec:mhp-methods}). ``BU" indicates bottom-up; ``1S-TD" indicates one-stage top-down; ``2S-TD" indicates two-stage top-down.}}
	\label{table:mhp_taxonomy}
	\vspace{-5pt}
	\begin{threeparttable}
	\resizebox{0.49\textwidth}{!}{
	\setlength\tabcolsep{2.5pt}
	\renewcommand\arraystretch{1.02}
	\begin{tabular}{c|r|c||c||c||c}
	\hline\thickhline
	\rowcolor{mygray1}
	Year & Method & Pub. & Pipeline &  Datasets   & Open Source\\
	\hline
	\hline
	\multirow{1}{*}{{2017}} & Holistic \cite{li2017holistic} & BMVC & 1S-TD & PPP &  -   \\
	\hline
	\multirow{1}{*}{{2018}} & PGN \cite{gong2018instance} & ECCV & BU & PPP/CIHP &  \checkmark \\
	\hline
	\multirow{4}{*}{{2019}} & CE2P \cite{ruan2019devil} & AAAI & 2S-TD & CIHP/MHP-v2.0 &  \checkmark \\
		                            & Parsing R-CNN \cite{yang2019parsing} & CVPR & 1S-TD & CIHP/MHP-v2.0 &  \checkmark  \\
		                            & BraidNet \cite{liu2019braidnet} & MM & 2S-TD & CIHP &    -  \\
		                            & Unified \cite{qin2019top} & BMVC & 1S-TD & PPP/CIHP &  -    \\
	\hline	                            
	\multirow{5}{*}{{2020}} & RP R-CNN \cite{yang2020renovating} & ECCV & 1S-TD & CIHP/MHP-v2.0 &  \checkmark  \\
		                            & SemaTree \cite{ji2020learning} & ECCV & 2S-TD & CIHP/MHP-v2.0 &  \checkmark \\
			                    & NAN \cite{zhao2020fine} & IJCV & BU & MHP-v1.0/MHP-v2.0 &  \checkmark  \\
		                            & SCHP \cite{li2020correction} & TPAMI & 2S-TD & CIHP/MHP-v2.0/VIP &  \checkmark \\
	\hline
        \multirow{1}{*}{{2021}} 
                                             & MGHR \cite{zhou2021differentiable, zhou2023differentiable} & \tabincell{c}{CVPR\\ TPAMI} & BU & \tabincell{c}{PPP/MHP-v2.0\\/COCO-DP} &  \checkmark   \\
	\hline
	\multirow{1}{*}{{2022}} & AIParsing \cite{zhang2022aiparsing} & TIP & 1S-TD & CIHP/MHP-v2.0/VIP & - \\
         \hline
	\multirow{3}{*}{{2023}} & HPSP \cite{li2023end} & TMM & BU & CIHP/MHP-v2.0 & - \\
	                                     & ReSParser \cite{dai2023resparser} & TMM & 1S-TD & CIHP/MHP-v2.0 & - \\	
	                                     & CID \cite{wang2023contextual} & TPAMI & 1S-TD & CIHP/MHP-v2.0 & - \\	
	                                    	
	\hline
	\end{tabular}
	}
	\end{threeparttable}
	\vspace{-5pt}
\end{table}

\noindent$\bullet$~\textbf{Bottom-up.} Bottom-up paradigm regards multiple human parsing as a fine-grained semantic segmentation task, which predicts the category of each pixel and grouping them into corresponding human instance. In a seminal work \cite{gong2018instance}, Gong \emph{et al}. propose a detection-free Part Grouping Network (PGN) that reformulates multiple human parsing as two twinned sub-tasks (semantic part segmentation and instance-aware edge detection) that can be jointly learned and mutually refined via a unified network. Among them, instance-aware edge detection task can group semantic parts into distinct human instances. Then, NAN \cite{zhao2020fine} proposes a deep Nested Adversarial Network for multiple human parsing. NAN consists of three GAN-like sub-nets, performing semantic saliency prediction, instance-agnostic parsing, and instance-aware clustering, respectively. Recently, Zhou \emph{et al}. \cite{zhou2021differentiable} propose a new bottom-up regime to learn category-level multiple human parsing as well as pose estimation in a joint and end-to-end manner, called Multi-Granularity Human Representation (MGHR) learning. MGHR exploits structural information over different human granularities, transforming the difficult pixel grouping problem into an easier multi human joint assembling task to simplify the difficulty of human instances discrimination. Similar to PGN \cite{gong2018instance}, HPSP \cite{li2023end} is also a detection-free multiple human parser, which decomposes this task into two subtasks via a unified network, namely semantic segmentation and instance segmentation, and obtains instance-level human parsing results through Hadamard product.

\noindent$\bullet$~\textbf{One-stage Top-down.} One-stage top-down is the mainstream paradigm of multiple human parsing. It first locates each human instance in the image plane, then segments each human part in an end-to-end manner. An early attempt is Holistic \cite{li2017holistic}, which consists of a human detection network and a part semantic segmentation network, then passing the results of both networks to an instance CRF \cite{kirfel2014human} to perform multiple human parsing. Inspired by Mask R-CNN \cite{he2017mask}, Qin \emph{et al}. \cite{qin2019top} propose a top-down unified framework that simultaneously performs human detection and single human parsing, identifying instances and parsing human parts in crowded scenes. A milestone one-stage top-down multiple human parsing model is proposed by Yang \emph{et al}., that enhances Mask R-CNN in all aspects, and proposes Parsing R-CNN \cite{yang2019parsing} network, greatly improving the accuracy of multiple human parsing concisely. Subsequently, Yang \emph{et al}. propose an improved version of Parsing R-CNN, called RP R-CNN \cite{yang2020renovating}, which introduces a global semantic enhanced feature pyramid network and a parsing re-scoring network into the high-performance pipeline, achieving better performance. AIParsing \cite{zhang2022aiparsing} introduces the anchor-free detector \cite{tian2020fcos} into the one-stage top-down paradigm for discriminating human instances, avoiding the hyper-parameters  sensitivity caused by anchors. Later, CID abandons detection boxes and decouples persons in an image into multiple instance-aware feature maps, which has better robustness to person detection errors.

\noindent$\bullet$~\textbf{Two-stage Top-down.} One-stage top-down and two-stage top-down paradigms are basically the same in operation flow. The difference between them is whether the detector is trained together with the segmentation sub-network in an end-to-end manner. All the two-stage bottom-up multiple human parsing methods consist of a human detector and a single human parser. The earliest attempt is CE2P \cite{ruan2019devil}, which designs a framework called M-CE2P on CE2P and Mask R-CNN, cropping the detected human instances, then sending them to the single human parser, finally combining the parsing results of all instances into a multiple human parsing prediction. Subsequent works, \emph{e.g}., BraidNet \cite{liu2019braidnet}, SemaTree \cite{ji2020learning}, and SCHP \cite{li2020correction}, basically inherit this pipeline.

\noindent\textbf{Remark.} The advantage of bottom-up and one-stage top-down is efficiency, and the advantage of two-stage top-down is accuracy. But as a non-end-to-end pipeline, the inference speed of two-stage top-down is positively correlated with the number of human instances, which also limits its practical application value. The detailed highlights of three human instances discrimination methods are summarized in Table~\ref{table:mhp_highlights}.

\begin{table}
	\centering
	\captionsetup{font=small}
	\caption{\small{\textbf{Highlights of human instances discrimination methods for MHP models} (\S\ref{sec:mhp-methods}). Representative Works of each method are also give.}}
	\label{table:mhp_highlights}
	\vspace{-5pt}
	\begin{threeparttable}
	\resizebox{0.49\textwidth}{!}{
	\setlength\tabcolsep{2.5pt}
	\renewcommand\arraystretch{1.02}
	\begin{tabular}{c|c|c}
	\hline\thickhline
	\rowcolor{mygray1}
	Method                & \tabincell{c}{Representative\\ Works} & Highlights \\
	\hline
	\hline
         Bottom-up              & \tabincell{c}{\cite{gong2018instance, zhao2020fine, zhou2021differentiable}\\ \cite{li2023end}}    & \tabincell{c}{Good model efficiency, good accuracy on pixel-wise\\ segmentation, and poor accuracy on instances\\ discrimination.} \\
         \hline
	\tabincell{c}{One-stage\\ Top-down}         & \tabincell{c}{\cite{li2017holistic, yang2019parsing, qin2019top}\\ \cite{yang2020renovating, zhang2022aiparsing, dai2023resparser}\\ \cite{wang2023contextual}}   & \tabincell{c}{Better trade-off between model efficiency and accuracy. \\But pixel-wise segmentation, especially the part\\ boundary is not fine enough.} \\
	\hline
	\tabincell{c}{Two-stage\\ Top-down}                     & \tabincell{c}{\cite{ruan2019devil, liu2019braidnet, ji2020learning}\\ \cite{li2020correction}}    & \tabincell{c}{Good accuracy and poor efficiency, the model inference\\ time is proportional to human instances number.} \\
         \hline
	\end{tabular}
	}
	\end{threeparttable}
	\vspace{-5pt}
\end{table}

\vspace{-6pt}
\subsection{Video Human Parsing (VHP) Models}
\label{sec:vhp-methods}
Existing VHP studies mainly focus to propagate the first frame into the entire video by the affinity matrix, which represents the temporal correspondences learnt from raw video data. Considering the unsupervised learning paradigms, we can group them into three classes: cycle-tracking, reconstructive learning, and contrastive learning. We summarize the essential characteristics of reviewed VHP models in Table~\ref{table:vhp_taxonomy}.

\begin{table}
	\centering
	\captionsetup{font=small}
	\caption{\small{\textbf{Summary of essential characteristics for reviewed VHP models} (\S\ref{sec:vhp-methods}). ``Cycle." indicates cycle-tracking; ``Recons." indicates reconstructive learning; ``Contra." indicates contrastive learning. All models are test on the VIP dataset.}}
	\label{table:vhp_taxonomy}
	\vspace{-5pt}
	\begin{threeparttable}
	\resizebox{0.49\textwidth}{!}{
	\setlength\tabcolsep{2.5pt}
	\renewcommand\arraystretch{1.02}
	\begin{tabular}{c|r|c||ccc||c}
	\hline\thickhline
	\rowcolor{mygray1}
	Year & Method & Pub.  & \tabincell{c}{Cycle.} & \tabincell{c}{Recons.}  & \tabincell{c}{Contra.}    & Open Source \\
	\hline
	\hline
	\multirow{1}{*}{{2018}} & ATEN \cite{zhou2018adaptive} & MM &  \checkmark  &  -  &   -  &  \checkmark   \\
         \hline	                         
	\multirow{2}{*}{{2019}} & TimeCycle \cite{wang2019corres} & CVPR &  \checkmark   &  - &  -&  \checkmark   \\
		                            & UVC \cite{li2019joint} & NeurIPS & \checkmark  & \checkmark &- &  \checkmark    \\
	\hline
	\multirow{1}{*}{{2020}} & CRW \cite{jabri2020space} & NeurIPS & \checkmark & -& -&  \checkmark  \\
	\hline
	\multirow{4}{*}{{2021}} & ContrastCorr \cite{wang2021contrasive} & AAAI & \checkmark & \checkmark & & \checkmark    \\
	                                    & CLTC \cite{jeon2021mining} & CVPR & - & - &\checkmark &  -   \\
		                            & VFS \cite{xu2021rethinking} & ICCV &  -&  -& \checkmark &  \checkmark    \\
		                            & JSTG \cite{zhao2021modelling} & ICCV & \checkmark    &  -  & \checkmark & -   \\
	\hline
	\multirow{4}{*}{{2022}} & LIIR \cite{li2022locality} & CVPR &    -    & \checkmark  &    -   &  \checkmark   \\
	                                     & SCC \cite{son2022contrastive} & CVPR &  \checkmark  &   -   & \checkmark & -    \\
		                            & SFC \cite{hu2022semantic} & ECCV &  - & - & \checkmark & \checkmark   \\
		                            & UVC+ \cite{mckee2022transfer} & ArXiv &  \checkmark & \checkmark & \checkmark &  -    \\
	\hline
	\multirow{3}{*}{{2023}} & STVC \cite{li2023spatial} & CVPR & - & \checkmark & - & \checkmark   \\
	                                    & SMTC \cite{qian2023semantics} & ICCV & - & - & \checkmark & \checkmark    \\
	                                    & SiamMAE \cite{gupta2023siamese} & ArXiv & - & \checkmark & - &  -    \\
	\hline
	\end{tabular}
	}
	\end{threeparttable}
\end{table}

\noindent$\bullet$~\textbf{Cycle-tracking.}  Early VHP methods model the unsupervised learning target mainly by the cycle-consistency of video frames, \emph{i.e.}, pixels/patches are expected to fall into the same locations after a cycle of forward-backward tracking. ATEN\cite{zhou2018adaptive} first leverages convolutional gated recurrent units to encode temporal feature-level changes, optical flow of non-key frames is wrapped with the temporal memory to generate their features. TimeCycle \cite{wang2019corres} tracks the reference patch backward-forward in the video. The reference and the tracked patch at the end of the tracking cycle are considered to be consistent both in spatial coordinates and feature representation. Meanwhile, UVC \cite{li2019joint} performs the region-level tracking and pixel-level corresponding with a shared affinity matrix, the tracked patch feature and the region-corresponding sub-affinity matrix are used to reconstruct the reference patch. Roles of the target and reference patches are then switched to regularizing the affinity matrix as orthogonal, which satisfies the cycle-consistency constraint. Its later version, UVC+ \cite{mckee2022transfer} combines features learned by image-based tasks with video-based counterparts to further boost the performance. Lately, CRW \cite{jabri2020space} represents video as a graph, where nodes are patches and edges are affinities between nodes in adjacent frames. A cross-entropy loss guides a graph walk to track the initial node bi-directionally in feature space, which is considered the target node after a bunch of cycle paths. However, the cycle-consistency in \cite{wang2019corres}, \cite{jabri2020space} strictly assumes that the target patch preserves visible in consecutive frames. Once it is occluded or disappears, the correspondences will be incorrectly assigned, thus leaving an optimal transport problem between video frames.

\begin{table}
	\centering
	\captionsetup{font=small}
	\caption{\small{\textbf{Highlights of temporal correspondences learning methods for VHP models} (\S\ref{sec:vhp-methods}). Representative Works of each method are also give.}}
	\label{table:vhp_highlights}
	\vspace{-5pt}
	\begin{threeparttable}
	\resizebox{0.49\textwidth}{!}{
	\setlength\tabcolsep{2.5pt}
	\renewcommand\arraystretch{1.02}
	\begin{tabular}{c|c|c}
	\hline\thickhline
	\rowcolor{mygray1}
	Method                & \tabincell{c}{Representative\\ Works} & Highlights \\
	\hline
	\hline
	\tabincell{c}{Cycle-\\ tracking}         & \tabincell{c}{\cite{wang2019corres, li2019joint}\\ \cite{mckee2022transfer, jabri2020space}}   & \tabincell{c}{Capturing temporal variations, may produce\\ wrong correspondences when occlusion occurs.} \\
	\hline
	\tabincell{c}{Reconstructive\\ Learning}         & \tabincell{c}{\cite{wang2021contrasive, li2022locality}\\ \cite{li2023spatial, gupta2023siamese}}   & \tabincell{c}{Modelling fine-grained temporal correspondence\\ and guiding focus on part details.} \\
	\hline
	\tabincell{c}{Contrastive\\ Learning}                     & \tabincell{c}{\cite{jeon2021mining, xu2021rethinking}\\ \cite{zhao2021modelling, son2022contrastive, qian2023semantics}}    & \tabincell{c}{Search for discriminative features to segment\\ similar or position-transformed human instances.} \\
         \hline
	\end{tabular}
	}
	\end{threeparttable}
	\vspace{-5pt}
\end{table}

\begin{figure}
	\begin{center}
		\includegraphics[width=0.98\linewidth]{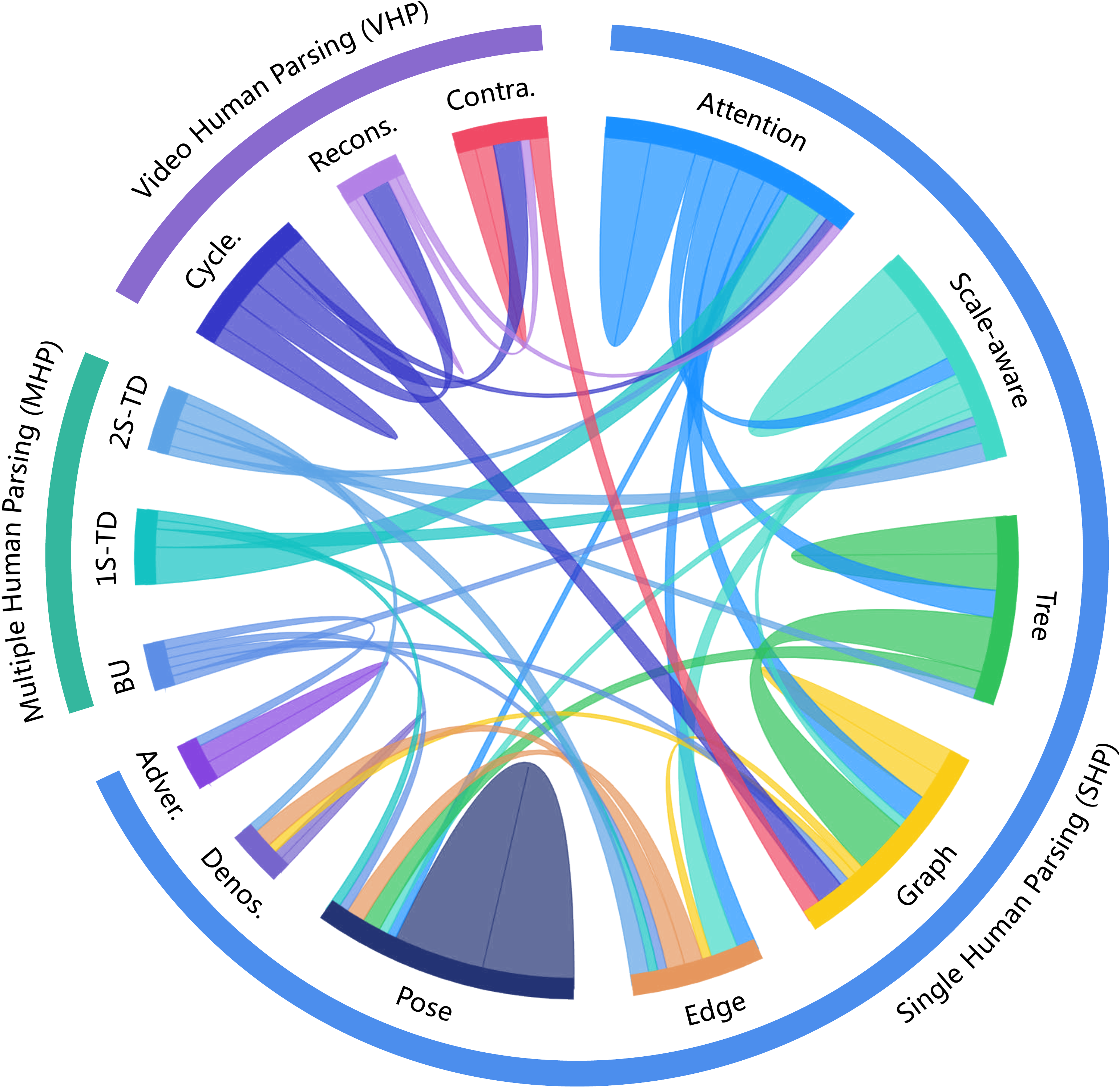}
	\end{center}
	\vspace{-5pt}
	\captionsetup{font=small}
	\caption{\small{\textbf{Correlations of different SHP, MHP and VHP methods} (\S\ref{sec:hp-summary}). We use the connections between the arc edges to summary the correlation between human parsing methods, each connecting line stands for a study that uses both methods. The longer the arc, the more methods of this kind, same for the width of connecting lines. This correlation summary reveals the prevalence of various human parsing methods.}}
	\label{fig:corre-mothods}
	\vspace{-10pt}
\end{figure}

\noindent$\bullet$~\textbf{Reconstructive Learning.} As video contents smoothly shift in time, pixels in a ``query" frame can be considered as copies from a set of pixels in other reference frames \cite{vondrick2018tracking, liu2018switchable}. Following UVC \cite{li2019joint} to establish pixel-level correspondence, several methods \cite{wang2021contrasive, li2022locality} are proposed to learn temporal correspondence completely by reconstructing correlating frames. Subsequently, ContrastCorr \cite{wang2021contrasive} not only learns from intra-video self-supervision, but also steps further to introduce inter-video transformation as negative correspondence. The inter-video distinction enforces the feature extractor to learn discriminations between videos while preserving the fine-grained matching characteristic among intra-video frame pairs. Based on the intra-inter video correlation, LIIR \cite{li2022locality} introduces a locality-aware reconstruction framework, which encodes position information and involves spatial compactness into intra-video correspondence learning, for locality-aware and efficient visual tracking. Most Recently, novel researches \cite{li2023spatial, gupta2023siamese} keep focus on effectively intra-video spatio-temporal reconstruction. STVC \cite{li2023spatial} significantly emphases on maintaining the spatial contexts when manipulating temporal correlating, simultaneously reconstructing video frames and global-local temporal correlations under the pseudo supervision of multi-scale features. SiamMAE \cite{gupta2023siamese} concisely extends MAE \cite{he2022masked} to a siamese architecture, which masks video frames asymmetrically along temporal dimension and reconstructs the highly-masked future frame patches from unchanged current frame.

\begin{figure}
	\begin{center}
		\includegraphics[width=0.98\linewidth]{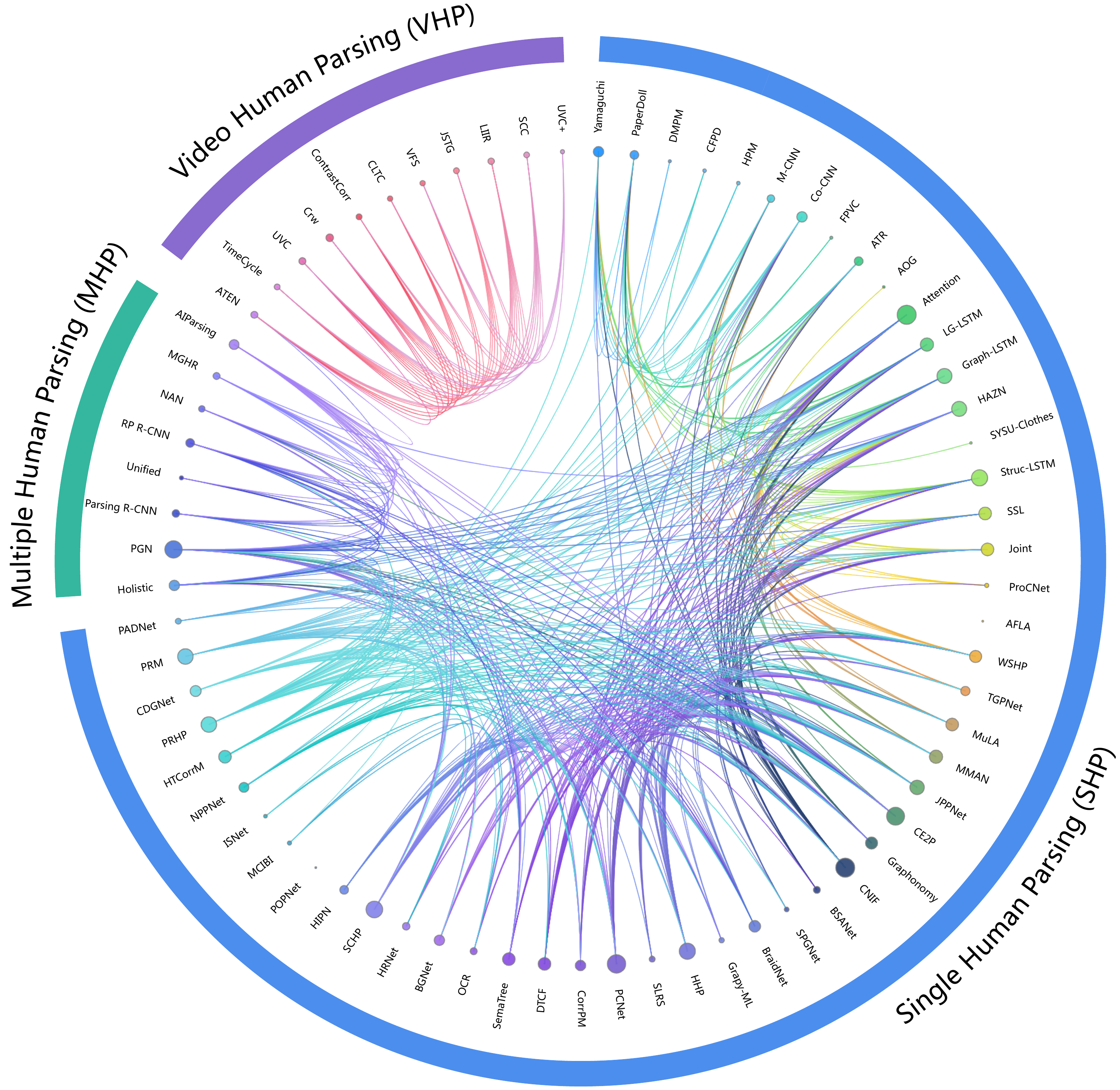}
	\end{center}
	\vspace{-5pt}
	\captionsetup{font=small}
	\caption{\small{\textbf{Correlations of different SHP, MHP and VHP studies} (\S\ref{sec:hp-summary}). We list out all the involved human parsing studies by dots and use connecting lines to represent their citing relations. The citing relation here refers to the citation appears in experimental comparisons, to avoid citations of low correlation in background introduction. As each line represents a citation between two studies, so the larger the dot, the more times cited. These correlations highlight the relatively prominent studies.}}
	\label{fig:corre-studies}
	\vspace{-10pt}
\end{figure}

\noindent$\bullet$~\textbf{Contrastive Learning.} Following the idea of pulling positive pairs close together and pushing negative pairs away from each other, considerable VHP algorithms adopt contrastive learning as a training objective. To solve the optimal transport problem, CLTC \cite{jeon2021mining} proposes to mine positive and semi-hard negative correspondences via consistency estimation and dynamic hardness discrimination, respectively. Subsequently, VFS \cite{xu2021rethinking} learns visual correspondences at frame level, with the guidance of image-level contrastive learning  data augmentation \cite{he2020momentum} and a well-designed temporal sampling strategy. SFC \cite{hu2022semantic} soon reinforces the global semantic correspondence of VFS with a fine-grained contrastive supervision. Encouraging positive temporal neighbors to be consistent, the global and local correspondences are fused together to propagate first frame part labels to consecutive frames. Lately, \cite{zhao2021modelling, son2022contrastive} extend the video graph with space relations of neighbor nodes, which determine the aggregation strength from intra-frame neighbors. The proposed space-time graph draws more attention to the association of center-neighbor pairs, thus explicitly helping learning correspondence between part instances. SCC \cite{son2022contrastive} mixes sequential Bayesian filters to formulate the optimal paths that track nodes from one frame to others, to alleviate the correspondence missing caused by random occlusion. Unlike the previous researches aiming at generic visual correspondence learning, SMTC \cite{qian2023semantics} propose to focus on object-centric spatio-temporal representation on top of the fused semantic features and frame correspondence maps. Specifically, query slot attention is responsible for extracting potential semantic masks and object instances, which are supervised by contrastive objectives to keep temporally consistent.

\noindent\textbf{Remark.} To our investigation scope, the current VHP research essentially follows an unsupervised semi-automatic video object segmentation setup. But considering the potential demand, it is more expectant to fully utilize the annotations and solve the VHP problem through an instance-discriminative manner, \emph{i.e}., a fine-grained video instance segmentation task. The highlights of temporal correspondences learning methods for VHP are shown in Table~\ref{table:vhp_highlights}.


\vspace{-6pt}
\subsection{Summary}
\label{sec:hp-summary}

Through the detailed review, we have subdivided SHP, MHP, and VHP studies into multiple methods and discussed their characteristics. To further investigate the development picture of the human parsing community, we summarize the correlations of the methods in Figure~\ref{fig:corre-mothods} and correlations of the involved studies in Figure~\ref{fig:corre-studies}, respectively.

Figure~\ref{fig:corre-mothods} presents correlations between research methods, \emph{i.e}., two methods are connected if a study uses both as its technical components, making the length of arcs represent the number of studies using them. The connecting line distribution first obviously shows that Graph (Structure), Attention (Mechanism), and Edge(-aware Learning) of SHP are more correlated with multiple other methods, which indicates their compatibility with others and prevalence in the community. It is worth noting that though Tree (Structure) has many correlations with others, a large proportion of them are with Graph method. This phenomenon indicates that Tree method is much less generalizable compared to Graph, Attention, and Edge methods. Regrettably, negligible relations between VHP and other methods show that current VHP studies have not yet gone deep into parts relationship modeling or human instance discrimination. 

The correlations of human parsing studies are presented in form of citing relations as Figure~\ref{fig:corre-studies}, each line represents a citation between two studies. For reliable statistics, we only consider citations that appear in experimental comparisons for all studies. From the citing relations, we can easily observe that Attention \cite{chen2016attenttion}, JPPNet \cite{liang2018look}, CE2P \cite{ruan2019devil}, CNIF \cite{wang2019learning} and PGN \cite{gong2018instance} have the largest dots, \emph{i.e}., they are experimental compared by most other studies, this indicates they are recognized as baseline studies of great prominence by the community. Additionally, since CE2P proposed to handle MHP sub-task by 2S-TD pipeline and make a milestone, lots of SHP studies start to compare their algorithms with MHP studies, this trend breaks down the barriers between the two sub-tasks of human parsing. Lastly, similar to the method correlation, VHP studies form citations strictly along with the proposed order among their own, which once again shows that VHP studies have not focused on human-centric data.

Synthesizing detailed review and correlation analysis, we can draw some conclusions about the historical evolution of human parsing models. First, the research focus has gradually shifted from SHP to MHP and VHP. As more challenging tasks, the latter two also have greater application potential. With the emergence of high-quality annotated datasets and the improvement of computing power, they have received increasing attention. Secondly, the technical diversity is insufficient, and the achievements of representation learning in recent years have not fully benefited the human parsing field. Finally, the number of open source work has increased significantly, but still insufficient. It is hoped that subsequent researchers will open source code and models as much as possible to benefit the follow-up researchers.

\section{Human Parsing Datasets}
\label{sec:hp-data}
In the past decades, a variety of visual datasets have been released for human parsing (upper part of Figure~\ref{fig:timeline}). We summarize the classical and commonly used datasets in Table~\ref{table:hp_datasets}, and give a detailed review from multiple angles.

\begin{table*}[t]
	\centering
	\captionsetup{font=small}
	\caption{\small{\textbf{Statistics of existing human parsing datasets.} See \S\ref{sec:shp-data} - \S\ref{sec:vhp-data} for more detailed descriptions. The 19 datasets are divided into 3 groups according to the human parsing taxonomy. ``Instance" indicates that instance-level human labels are provided; ``Temporal" indicates that video-level labels are provided; ``Super-pixel" indicates that super-pixels are used for labeling.}}
	\label{table:hp_datasets}
	\vspace{-5pt}
	\begin{threeparttable}
	\resizebox{0.99\textwidth}{!}{
	\setlength\tabcolsep{2.5pt}
	\renewcommand\arraystretch{1.02}
	\begin{tabular}{r||c|c|c|c|c|c|c|c|c|c}
	\hline\thickhline
	\rowcolor{mygray1}
	Dataset & Year & Pub. & \#Images & \#Train/Val/Test/ & \#Class & Purpose & Instance & Temporal & Super-pixel  & Other Annotations \\
	\hline
	\hline
        Fashionista \cite{yamaguchi2012parsing} & 2012 & CVPR & 685 & 456/-/299  & 56 &  Clothing & -      &    -   &\checkmark&     Clothing-tag \\
        CFPD \cite{liu2013fashion} & 2013 & TMM & 2,682 & 1,341/-/1,341  & 23 &  Clothing &   -    &    -   & \checkmark &     Color-seg. \\
        DailyPhotos \cite{dong2013deformable} & 2013 & ICCV & 2,500 & 2,500/-/-  & 19 &  Clothing & -      &   -    &\checkmark&     Clothing-tag\\
        PPSS \cite{luo2013pedestrian} & 2013 & ICCV & 3,673 & 1,781/-/1,892  & 6 &  Human &   -    &   -    &   -    & -      \\
        ATR \cite{liang2015deep} & 2015 & TPAMI & 7,700 & 6,000/700/1,000  & 18 &  Human &   -    &   -    &   -    & -      \\
        Chictopia10k \cite{liang2015human} & 2015 & ICCV & 10,000 & 10,000/-/-  & 18 &  Clothing &   -    &  -     &  -     &    Clothing-tag   \\
        SYSU-Clothes \cite{liang2016clothes} & 2016 & TMM & 2,682 & 2,682/-/-  & 57 &  Clothing &   -    &  -     &\checkmark&     Clothing-tag  \\
        LIP \cite{gong2017look} & 2017 & CVPR &  50,462 & 30,462/10,000/10,000 & 20 &  Human &   -    &    -   &\checkmark&   -     \\
        ModaNet \cite{zheng2018modanet} & 2018 & MM & 55,176 & 52,377/2,799/-  & 57 &  Clothing &  -     &   -    & -  &  -       \\
        ATR-OS \cite{he2021progressive} & 2021 & AAAI & 18,000 & -/-/-  & 18 &  Human &     -  &    -   &    -   &  -     \\
        HRHP \cite{cvpr2021l2id} & 2021 & CVPRW & 7,500 & 6,000/500/1,000  & 20 &  Human &  -     &   -    &  -     &   -    \\
        \hline
        PASCAL-Person-Part \cite{chen2014detect} & 2014 & CVPR & 3,533 & 1,716/-/1,817  & 7 &  Human &\checkmark&  -     &-   &     Human-box    \\
        MHP-v1.0 \cite{li2017multiple} & 2017 & ArXiv & 4,980 & 3,000/1,000/980  & 19 &  Human &\checkmark&    -   &-   &     Human-box    \\
        MHP-v2.0 \cite{zhao2018understanding} & 2018 & MM & 25,403 & 15,403/5,000/5,000  & 59 &  Human &\checkmark&    -   & -  &    Human-box    \\
        COCO-DensePose \cite{guler2018densepose} & 2018 & CVPR & 27,659 & 26,151/-/1,508  & 15 &  Human &\checkmark&   -    &  - &    \tabincell{c}{Human-box/\\keypoints/densepoints}     \\
        CIHP \cite{gong2018instance} & 2018 & ECCV & 38,280 & 28,280/5,000/5,000  & 20 &  Human &\checkmark&   -    & -  &    Human-box    \\
        DeepFashion2 \cite{ge2019deepfashion2} & 2019 & CVPR & 491,895 & 390,884/33,669/67,342  & 14 &  Clothing &\checkmark&     -  &  - &    \tabincell{c}{Clothing-box/\\landmark/style}     \\
        \hline
        VIP \cite{zhou2018adaptive} & 2018 & MM & 21,246 & 18,468/-/2,778  & 20 &  Human &\checkmark& \checkmark & -  &    Human-box/identity    \\
        CPP \cite{de2021part} & 2021 & CVPR & 3,475 & 2,975/500/-  & 4 &  \tabincell{c}{Human/Scene} &\checkmark& \checkmark &  - &     \tabincell{c}{Human-box/identity,\\Semantic-/Instance-seg.}    \\
	\hline
	\end{tabular}
	}
	\end{threeparttable}
	\vspace{-5pt}
\end{table*}

\vspace{-6pt}
\subsection{Single Human Parsing (SHP) Datasets}
\label{sec:shp-data}

\noindent$\bullet$~\textbf{Fashionista (FS)} \cite{yamaguchi2012parsing} consists of 685 photographs collected from \texttt{Chictopia.com}, a social networking website for fashion bloggers. There are 456 training images and 299 testing images annotated with 56-class semantic labels, and text tags of garment items and styling are also provided. Fashionista was once the main single human/clothing parsing dataset but was limited by its scale. It is rarely used now.


\noindent$\bullet$~\textbf{Colorful Fashion Parsing Data (CFPD)} \cite{liu2013fashion} is also collected from \texttt{Chictopia.com}, which provides 23-class noisy semantic labels and 13-class color labels. The annotated images are usually grouped into 1,341/1,341 for \texttt{train}/\texttt{test}.

\noindent$\bullet$~\textbf{DailyPhotos (DP)} \cite{dong2013deformable} contains 2,500 high resolution images, which are crawled following the same strategy as the Fashionista dataset and thoroughly annotated with 19 categories.

\noindent$\bullet$~\textbf{PPSS} \cite{luo2013pedestrian} includes 3,673 annotated samples collected from 171 videos of different surveillance scenes and provides pixel-wise annotations for hair, face, upper-/lower-clothes, arm, and leg. It presents diverse real-word challenges, \emph{e.g}. pose variations, illumination changes, and occlusions. There are 1,781 and 1,892 images for training and testing, respectively.

\noindent$\bullet$~\textbf{ATR} \cite{liang2015deep} contains data which combined from three small benchmark datasets: the Fashionista \cite{yamaguchi2012parsing} containing 685 images, the CFPD \cite{liu2013fashion} containing 2,682 images, and the DailyPhotos \cite{dong2013deformable} containing 2,500 images. The labels are merged of Fashionista and CFPD datasets to 18 categories. To enlarge the diversity, another 1,833 challenging images are collected and annotated to construct the Human Parsing in the Wild (HPW) dataset. The final combined dataset contains 7,700 images, which consists of 6,000 images for training, 1,000 for testing, and 700 as the validation set. 

\noindent$\bullet$~\textbf{Chictopia10k} \cite{liang2015human} contains 10,000 real-world human pictures from \texttt{Chictopia.com}, annotating pixel-wise labels following \cite{liang2015deep}. The dataset mainly contains images in the wild (\emph{e.g}., more challenging poses, occlusion, and clothes).

\noindent$\bullet$~\textbf{SYSU-Clothes} \cite{liang2016clothes} consists of 2,098 high resolution fashion photos in high-resolution (about 800$\times$500 on average) from the shopping website. In this dataset, six categories of clothing attributes (\emph{e.g}., clothing category, clothing color, clothing length, clothing shape, collar shape, and sleeve length) and 124 attribute types of all categories are collected. 

\noindent$\bullet$~\textbf{Look into Person (LIP)} \cite{gong2017look} is the most popular single human parsing dataset, which is annotated with pixel-wise annotations with 19 semantic human part labels and one background label. LIP contains 50,462 annotated images and be grouped into 30,462/10,000/10,000 for \texttt{train}/\texttt{val}/\texttt{test}. The images in the LIP dataset are cropped person instances from COCO \cite{lin2014microsoft} training and validation sets.

\noindent$\bullet$~\textbf{ModaNet} \cite{zheng2018modanet} is a large-scale collection of images based on PaperDoll dataset \cite{yamaguchi2013paper}. It provides 55,176 street images and contains 14 clothing categories (including background) with fine polygon annotations. ModaNet generates bounding boxes from the polygon annotations for clothing detection. The dataset is split into 52,377 and 2,799 images for training and evaluation, respectively.

\noindent$\bullet$~\textbf{ATR-OS} \cite{he2021progressive} is a dataset for one-shot human parsing, which is based on ATR \cite{liang2015deep}. ATR-OS divides the samples into support set and query set for training and testing, respectively.

\noindent$\bullet$~\textbf{High-resolution Human Parsing (HRHP)} \cite{cvpr2021l2id} is a high-resolution single human parsing benchmark, which is introduced by Learning from Limited or Imperfect Data (L2ID) workshop on CVPR 2021. The data is collected from high-quality fashion media, and the image resolution is about 4,000$\times$4,000. For high-resolution human parsing, 6,000/500/1,000 images are finely labelled with 20 categories at pixel-level for \texttt{train}/\texttt{val}/\texttt{test}.

\noindent\textbf{Remark.} ATR and LIP are the mainstream benchmarks among these single human parsing datasets. In recent years, the research purpose has changed from ``clothing" to ``human", and the data scale and annotation quality have also been significantly improved.

\vspace{-6pt}
\subsection{Multiple Human Parsing (MHP) Datasets}
\label{sec:mhp-data}

\noindent$\bullet$~\textbf{PASCAL-Person-Part (PPP)} \cite{chen2014detect} is annotated from the PASCAL-VOC-2010 \cite{everingham2010pascal}, which contains 3,533 multi-person images with challenging poses and splits into 1,716 training images and 1,817 test images. Each image is pixel-wise annotated with 7 classes, namely head, torso, upper/lower arms, upper/lower legs, and a background category.

\noindent$\bullet$~\textbf{MHP-v1.0} \cite{li2017multiple} contains 4,980 multi-person images with fine-grained annotations at pixel-level. For each person, it defines 7 body parts, 11 clothing/accessory categories, and one background label. The \texttt{train}/\texttt{val}/\texttt{test} sets contain 3,000/1,000/980 images, respectively.

\noindent$\bullet$~\textbf{MHP-v2.0} \cite{zhao2018understanding} is an extend version of MHP-v1.0 \cite{li2017multiple}, which provides more images and richer categories. MHP-v2.0 contains 25,403 images and has great diversity in image resolution (from 85$\times$100 to 4,511$\times$6,919) and human instance number (from 2 to 26 persons). These images are split into 15,403/5,000/5,000 for \texttt{train}/\texttt{val}/\texttt{test} with 59 categories.

\noindent$\bullet$~\textbf{COCO-DensePose (COCO-DP)} \cite{guler2018densepose} aims at establishing the mapping between all human pixels of an RGB image and the 3D surface of the human body, and has 27,659 images (26,151/1,508 for \texttt{train}/\texttt{test} splits) gathered from COCO \cite{lin2014microsoft}. The dataset provides 15 pixel-wise human parts with dense keypoints annotations.

\noindent$\bullet$~\textbf{Crowd Instance-level Human Parsing (CIHP)} \cite{gong2018instance} is the largest multiple human parsing dataset to date. With 38,280 diverse real-world images, the persons are labelled with pixel-wise annotations on 20 categories.  It consists of 28,280 training and 5,000 validation images with publicly available annotations, as well as 5,000 test images with annotations withheld for benchmarking purposes. All images of the CIHP dataset contain two or more instances with an average of 3.4.

\noindent$\bullet$~\textbf{DeepFashion2} \cite{ge2019deepfashion2} is currently the largest dataset for clothing understanding, which contains 491,895 images (390,884/33,669/67,342 for \texttt{train}/\texttt{val}/\texttt{test}) of 13 clothing categories and a background category. A full spectrum of tasks are defined on them, including clothes detection and recognition, landmark and pose estimation, segmentation, as well as verification and retrieval.

\noindent\textbf{Remark.} So far, several multiple human parsing datasets have high-quality annotation and considerable data scale. In addition to pixel-wise parsing annotations, many datasets provide other rich annotations, such as box, keypoints/landmark and style. PPP, CIHP and MHP-v2.0 are widely studied datasets, and most classical multiple human parsing methods have been verified on them.

\vspace{-6pt}
\subsection{Video Human Parsing (VHP) Datasets}
\label{sec:vhp-data}

\noindent$\bullet$~\textbf{Video Instance-level Parsing (VIP)} \cite{zhou2018adaptive} is the first video human parsing dataset. VIP contains 404 multi-person Full HD sequences, which are collected from \texttt{Youtube} with great diversity. For every 25 consecutive frames in each sequence, one frame is densely annotated with 20 classes and identities. All the sequences are grouped into 354/50 for \texttt{train}/\texttt{test}, containing 18,468/2,778 annotated frames respectively.

\noindent$\bullet$~\textbf{Cityscapes Panoptic Parts (CPP)} \cite{de2021part} aims at part-aware panoptic segmentation, which annotates part-level semantic labels on the popular Cityscapes \cite{cordts2016cityscapes}. CPP inherits the annotation of original Cityscapes (\emph{e.g}. semantic segmentation, instance segmentation, and temporal identity), and the human instance is annotated with only 4 categories, including head, torso, arms and legs. The dataset contains 18/3 urban sequences and 2,975/500 frames for \texttt{train}/\texttt{val}.

\noindent\textbf{Remark.} Since video human parsing has only attracted attention in recent years, there are few publicly available datasets, and its data scale and richness still need to be continuously invested by the community. 

\vspace{-6pt}
\subsection{Summary}
\label{sec:data-summary}
Through Table~\ref{table:hp_datasets}, we can observe that the human parsing datasets show several development trends. Firstly, the scale of datasets continues to increase, from hundreds in the early years \cite{yamaguchi2012parsing} to a tens of thousands now \cite{gong2017look, gong2018instance}. Secondly, the quality of annotation is constantly improving. Some early datasets use super-pixel \cite{yamaguchi2012parsing, liang2016clothes, gong2017look} to reduce the annotation cost, while in recent years, pixel-wise accurate annotation has been adopted. Finally, the annotation dimensions are becoming increasingly diverse, \emph{e.g}., COCO-DensePose \cite{guler2018densepose} provides boxes, keypoints, and UVs annotation in addition to parsing. 

\section{Performance Comparisons}
\label{sec:pc}
To provide a more intuitive comparison, we tabulate the performance of several previously discussed models. It should be noted that the experimental settings of each study are not entirely consistent (\emph{e.g}., backbone, input size, training epochs). 
Therefore, we suggest only taking these comparisons as references, and a more specific analysis needs to study the original articles deeply.

\vspace{-6pt}
\subsection{SHP Performance Benchmarking}
\label{sec:shp-bench}
We select ATR \cite{liang2015deep} and LIP \cite{gong2017look} as the benchmark for single human parsing performance comparison, and compared 14 and 28 models, respectively.

\vspace{-6pt}
\subsubsection{Evaluation Metrics}
The evaluation metrics of single human parsing are basically consistent with semantic segmentation \cite{shelhamer2016fully}, including pixel accuracy, mean pixel accuracy, and mean IoU. In addition, foreground pixel accuracy and F-1 score are also commonly used metrics on the ATR dataset.

\noindent$\bullet$~\textbf{Pixel accuracy (pixAcc)} is the simplest and intuitive metric, which expresses the proportion of pixels with correct prediction in the overall pixel.

\noindent$\bullet$~\textbf{Foreground pixel accuracy (FGAcc)} only calculates the pixel accuracy of foreground human parts.

\noindent$\bullet$~\textbf{Mean pixel accuracy (meanAcc)} is a simple improvement of pixel accuracy, which calculates the proportion of correctly predicted pixels in each category.

\noindent$\bullet$~\textbf{Mean IoU (mIoU)} is short for mean intersection over union, which calculates the ratio of the intersection and union of two sets. The two sets are the ground-truth and predicted results of each category respectively.

\noindent$\bullet$~\textbf{F-1 score (F-1)} is the harmonic average of precision and recall, which is a common evaluation metric.

\begin{table}
	\centering
	\captionsetup{font=small}
	\caption{\small{\textbf{Quantitative SHP results on ATR} \texttt{test} (\S\ref{sec:shp-bench}) in terms of pixel accuracy (pixAcc), foreground pixel accuracy (FGAcc) and F-1 score (F-1). The three best scores are marked in \textcolor{scorered}{\textbf{red}}, \textcolor{scoreblue}{\textbf{blue}}, and \textcolor{scoregreen}{\textbf{green}}, respectively.}}
	\label{table:atrbench}
	\vspace{-5pt}
	\begin{threeparttable}
	\resizebox{0.49\textwidth}{!}{
	\setlength\tabcolsep{2.5pt}
	\renewcommand\arraystretch{1.02}
	\begin{tabular}{c|r|c||c|c|c||c|c|c}
	\hline\thickhline
	\rowcolor{mygray1}
	Year & Method & Pub. & Backbone  & \#Input Size & \#Epoch  & pixAcc & FGAcc & F-1   \\
	\hline
	\hline
	2012 & Yamaguchi \cite{yamaguchi2012parsing} & CVPR  &  - & - & - &  84.38 & 55.59 & 41.80  \\
	\hline
	2013 & Paperdoll \cite{yamaguchi2013paper} & ICCV  &  - & - & - &  88.96 & 62.18 & 44.76  \\
	\hline
	\multirow{3}{*}{{2015}} & M-CNN \cite{liu2015matching} & CVPR  &  - & - & 50 &  89.57 & 73.98 & 62.81  \\
	                                    & Co-CNN \cite{liang2015human} & ICCV  &  - & 150$\times$100  & 90 &  95.23 & 80.90 & 76.95  \\
	                                    & ATR \cite{liang2015deep} & TPAMI  &  - & 227$\times$227 & 120 &  91.11 & 71.04 & 64.38  \\
	\hline                                    
	\multirow{2}{*}{{2016}} & LG-LSTM \cite{liang2016object} & CVPR  &  VGG16 & 321$\times$321  & 60 &  96.18 & 84.79 & 80.97  \\
	                                    & Graph-LSTM \cite{liang2016semantic} & ECCV  &  VGG16 & 321$\times$321  & 60 &  \textcolor{scoreblue}{\textbf{97.60}} &  \textcolor{scoreblue}{\textbf{91.42}} & 83.76  \\
	\hline                                    
	2017 & Struc-LSTM \cite{liang2017interpretable} & CVPR  &  VGG16 & 321$\times$321  & 60 &  \textcolor{scorered}{\textbf{97.71}} & \textcolor{scorered}{\textbf{91.76}} &  \textcolor{scorered}{\textbf{87.88}}  \\
	\hline
	2018 & TGPNet \cite{luo2018trusted} & MM  & VGG16 & 321$\times$321 & 35 &  96.45 & 87.91 & 81.76  \\
	\hline
	2019 & CNIF \cite{wang2019learning} & ICCV  & ResNet101 & 473$\times$473 & 150 &  96.26 & 87.91 & 85.51  \\
	\hline
	\multirow{3}{*}{{2020}} & CorrPM \cite{zhang2020correlating}  & CVPR  & ResNet101 & 384$\times$384 & 150 &  97.12 &  \textcolor{scoregreen}{\textbf{90.40}} & 86.12  \\
	                                    & HHP \cite{wang2020hierarchical}   & CVPR  & ResNet101 & 473$\times$473 & 150 &  96.84 & 89.23 &  \textcolor{scoreblue}{\textbf{87.25}}  \\
	                                    & SCHP \cite{li2020correction}  & TPAMI  & ResNet101 & 473$\times$473 & 150 &  96.25 & 87.97 & 85.55  \\
	\hline                                    
	\multirow{1}{*}{{2022}} & CDGNet \cite{liu2022cdgnet}  & CVPR  & ResNet101 & 512$\times$512 & 250 &  \textcolor{scoregreen}{\textbf{97.39}} & 90.19 & \textcolor{scoregreen}{\textbf{87.16}}  \\

	\hline
	\end{tabular}
	}
	\end{threeparttable}
	\vspace{-5pt}
\end{table}

\begin{table}
	\centering
	\captionsetup{font=small}
	\caption{\small{\textbf{Quantitative SHP results on LIP} \texttt{val} (\S\ref{sec:shp-bench}) in terms of pixel accuracy (pixAcc), mean pixel accuracy (meanAcc) and mean IoU (mIoU). The three best scores are marked in \textcolor{scorered}{\textbf{red}}, \textcolor{scoreblue}{\textbf{blue}}, and \textcolor{scoregreen}{\textbf{green}}, respectively.}}
	\label{table:lipbench}
	\vspace{-5pt}
	\begin{threeparttable}
	\resizebox{0.49\textwidth}{!}{
	\setlength\tabcolsep{2.5pt}
	\renewcommand\arraystretch{1.02}
	\begin{tabular}{c|r|c||c|c|c||c|c|c}
	\hline\thickhline
	\rowcolor{mygray1}
	Year & Method & Pub. & Backbone  & \#Input Size & \#Epoch  & pixAcc & meanAcc & mIoU   \\
	\hline
	\hline
	\multirow{1}{*}{{2017}} & SSL \cite{gong2017look} & CVPR  &  VGG16 & 321$\times$321 & 50 &  - & - & 46.19  \\
	\hline                                    
	\multirow{4}{*}{{2018}} & HSP-PRI \cite{kalayeh2018human} & CVPR  &  InceptionV3 & - & - &  85.07 & 60.54 & 48.16  \\
	                                    & MMAN \cite{luo2018macro} & ECCV  &  ResNet101 & 256$\times$256 & 30 &  85.24 & 57.60 & 46.93  \\
	                                    & MuLA \cite{nie2018mutual} & ECCV  &  Hourglass & 256$\times$256 & 250 &  88.50 & 60.50 & 49.30  \\
	                                    & JPPNet \cite{liang2018look} & TPAMI  &  ResNet101 & 384$\times$384 & 60 &  86.39 & 62.32 & 51.37  \\
	\hline                                             
	\multirow{5}{*}{{2019}} & CE2P \cite{ruan2019devil} & AAAI  &  ResNet101 & 473$\times$473 & 150 &  87.37 & 63.20 & 53.10  \\
		                            & CNIF \cite{wang2019learning} & ICCV  &  ResNet101 & 473$\times$473 & 150 &  88.03 & 68.80 & 57.74  \\
		                            & CCNet \cite{huang2023ccnet} & \tabincell{c}{ICCV\\ TPAMI}   &  ResNet101 & 473$\times$473 & 150 &  88.01 & 63.91 & 55.47  \\
		                            & BraidNet \cite{liu2019braidnet} & MM  &  ResNet101 & 384$\times$384 & 150 &  87.60 & 66.09 & 54.42  \\
	\hline          	                            	
	\multirow{10}{*}{{2020}} & CorrPM \cite{zhang2020correlating} & CVPR  &  ResNet101 & 384$\times$384 & 150 &  - & - & 55.33  \\
		                              & SLRS \cite{li2020self} & CVPR  &  ResNet101 & 384$\times$384 & 150 &  88.33 & 66.53 & 56.34  \\                      
		                              & PCNet \cite{zhang2020pcnet} & CVPR  &  ResNet101 & 473$\times$473 & 120 &  - & - & 57.03  \\
		                              & HHP \cite{wang2020hierarchical} & CVPR  &  ResNet101 & 473$\times$473 & 150 &   \textcolor{scoreblue}{\textbf{89.05}} & \textcolor{scoregreen}{\textbf{70.58}} & 59.25  \\
		                              & DTCF \cite{liu2020hybrid} & MM  &  ResNet101 & 473$\times$473 & 200 &  88.61 & 68.89 & 57.82  \\
		                              & SemaTree \cite{ji2020learning} & ECCV  &  ResNet101 & 384$\times$384 & 200 &  88.05 & 66.42 & 54.73  \\
		                              & OCR \cite{yuan2020object} & ECCV  &  HRNetW48 & 473$\times$473 &  \app100 &  - & - & 56.65  \\
		                              & BGNet \cite{zhang2020blended} & ECCV  &  ResNet101 & 473$\times$473 & 120 &  - & - & 56.82  \\
		                              & HRNet \cite{wang2020deep} & TPAMI  &  HRNetW48 & 473$\times$473 & \app150 &  88.21 & 67.43 & 55.90  \\
		                              & SCHP \cite{li2020correction} & TPAMI  &  ResNet101 & 473$\times$473 & 150 &  - & - & 59.36  \\
	\hline          		                              
	\multirow{5}{*}{{2021}} & HIPN \cite{liu2021hier} & AAAI  &  ResNet101 & 473$\times$473 & 150 &   \textcolor{scorered}{\textbf{89.14}} & \textcolor{scoreblue}{\textbf{71.09}} &  59.61  \\
		                              & MCIBI \cite{jin2021mining} & ICCV  &  ResNet101 & 473$\times$473 & 150 &  - & - & 55.42  \\
		                              & ISNet \cite{jin2021isnet} & ICCV  &  ResNet101 & 473$\times$473 & 160 &  - & - & 56.96  \\
		                              & NPPNet \cite{zeng2021neural} & ICCV  &  NAS & 384$\times$384 & 120 &  - & - & 58.56  \\
		                              & HTCorrM \cite{zhang2021on} & TPAMI  &  HRNetW48 & 384$\times$384 & 180 &  - & - & 56.85  \\
	\hline          		                             
	\multirow{3}{*}{{2022}}   & CDGNet \cite{liu2022cdgnet} & CVPR  &  ResNet101 & 473$\times$473 & 150 &  \textcolor{scoregreen}{\textbf{88.86}} &  \textcolor{scorered}{\textbf{71.49}} &  \textcolor{scoreblue}{\textbf{60.30}}  \\
	                                      & HSSN \cite{li2022deep} & CVPR  &  ResNet101 & 480$\times$480 & \app84 &  - &  - &  \textcolor{scorered}{\textbf{60.37}}  \\
		                              & PRM \cite{zhang2022human} & TMM  &  ResNet101 & 473$\times$473 & 120 &  - & - & 58.86  \\
	\hline     		                             
	\multirow{1}{*}{{2023}}   & SOLIDER \cite{chen2023beyond} & CVPR  &  Swin-S & 473$\times$473 & 150 &  - &  - &  \textcolor{scoregreen}{\textbf{60.21}}  \\

	\hline
	\end{tabular}
	}
	\end{threeparttable}
\end{table}

\vspace{-6pt}
\subsubsection{Results}
Table~\ref{table:atrbench} presents the performance of the reviewed SHP methods on ATR \texttt{test} set. Struc-LSTM \cite{liang2017interpretable} achieves the best performance, scoring 91.71\% pixAcc. and 87.88\% F-1 score, which greatly surpassed other methods. Table~\ref{table:lipbench} shows the method results on the LIP benchmark since 2017. Overall, HIPN \cite{liu2021hier} and HSSN \cite{li2022deep} achieve remarkable results in various metrics, in which HIPN scored 89.14\% pixelAcc and HSSN scored 60.37\% mIoU.

\vspace{-6pt}
\subsection{MHP Performance Benchmarking}
\label{sec:multi-bench}
We select 7 models experimented on PASCAL-Person-Part \cite{chen2014detect}, 12 models experimented on CIHP \cite{gong2018instance} and 11 models experimented on MHP-v2 \cite{zhao2018understanding} to compare the performance of multiple human parsing. 

\begin{table}
	\centering
	\captionsetup{font=small}
	\caption{\small{\textbf{Quantitative MHP results on PASCAL-Person-Part} \texttt{test} (\S\ref{sec:multi-bench}) in terms of mIoU, AP$^\text{r}_\text{vol}$ and AP$^\text{r}_\text{50}$. We only mark the best score in \textcolor{scorered}{\textbf{red}} color.}}
	\label{table:pppbench}
	\vspace{-5pt}
	\begin{threeparttable}
	\resizebox{0.49\textwidth}{!}{
	\setlength\tabcolsep{2.5pt}
	\renewcommand\arraystretch{1.02}
	\begin{tabular}{c|r|c||c|c|c||c|c|c}
	\hline\thickhline
	\rowcolor{mygray1}
	Year & Method & Pub. & Pipeline &  Backbone  & \#Epoch  & mIoU & AP$^\text{r}_\text{vol}$ & AP$^\text{r}_\text{50}$   \\
	\hline
	\hline
	\hline	                            
	\multirow{1}{*}{{2017}} 
		                            & Holistic \cite{li2017holistic} & BMVC & 1S-TD & ResNet101 &  100 & 66.34 & 38.40   &  40.60  \\
	\hline	                            
	\multirow{1}{*}{{2018}} 
		                            & PGN \cite{gong2018instance} & ECCV & BU & ResNet101 &  \app80 & \textcolor{scorered}{\textbf{68.40}} & 39.20   &  39.60  \\
	\hline	                            
	\multirow{2}{*}{{2019}} & Parsing R-CNN \cite{yang2019parsing} & CVPR & 1S-TD & ResNet50 &  75 & 62.70 & 40.40   &  43.70  \\
		                            & Unified \cite{qin2019top} & BMVC & 1S-TD & ResNet101 &  \app600 & - & 43.10  &  48.10  \\
	\hline	                            
	\multirow{2}{*}{{2020}} 
			                      & RP R-CNN \cite{yang2020renovating} & ECCV & 1S-TD & ResNet50 &  75 & 63.30 & 40.90  &  44.10  \\
			                      & NAN \cite{zhao2020fine} & IJCV & BU & - &  80 & - & 52.20   &   \textcolor{scorered}{\textbf{59.70}}   \\
	\hline		                      
	\multirow{1}{*}{{2021}} 
		                            & MGHR \cite{zhou2021differentiable, zhou2023differentiable} & \tabincell{c}{CVPR\\ TPAMI} & BU & ResNet101 &  150 & - & \textcolor{scorered}{\textbf{55.90}}   &  59.00  \\

	\hline
	\end{tabular}
	}
	\end{threeparttable}
	\vspace{-5pt}
\end{table}

\vspace{-6pt}
\subsubsection{Evaluation Metrics}
Generally speaking, multiple human parsing uses mIoU to measure the semantic segmentation performance, and AP$^\text{r}_\text{vol}$/AP$^\text{r}_\text{50}$ or AP$^\text{p}_\text{vol}$/AP$^\text{p}_\text{50}$ to measure the performance of instance discrimination.

\noindent$\bullet$~\textbf{Average precision based on region (AP$^\text{r}_\text{vol}$/AP$^\text{r}_\text{50}$)} \cite{hariharan2014simu} is similar to AP metrics in object detection \cite{lin2014microsoft}. If the IoU between the predicted part and ground-truth part is higher than a certain threshold, the prediction is considered to be correct, and the mean Average Precision is calculated. The defined AP$^\text{r}_\text{vol}$ is the mean of the AP score for overlap thresholds varying from 0.1 to 0.9 in increments of 0.1 and AP$^\text{r}_\text{50}$ is the AP score for threshold equals 0.5.

\noindent$\bullet$~\textbf{Average precision based on part (AP$^\text{p}_\text{vol}$/AP$^\text{p}_\text{50}$)} \cite{li2017multiple, zhao2020fine} is adopted to evaluate the instance-level human parsing performance. AP$^\text{p}$ is very similar to AP$^\text{r}$ in calculation mode, except that it calculates mIoU with the whole human body.

\vspace{-6pt}
\subsubsection{Results}
PASCAL-Person-Part benchmark is the classical benchmark in multiple human parsing. Table~\ref{table:pppbench} gathers the results of 7 models on PASCAL-Person-Part \texttt{test} set. PGN \cite{gong2018instance} is the top one in mIoU metric. In AP$^\text{r}_\text{vol}$/AP$^\text{r}_\text{50}$ metrics, MGHR \cite{zhou2021differentiable, zhou2023differentiable}, and NAN \cite{zhao2020fine} are the best two methods at present. The results on CIHP \texttt{val} set are summarized in Table~\ref{table:cihpbench}. As seen, SCHP \cite{li2020correction} performs the best on all metrics, which yields 67.67\% mIoU, 52.74\% AP$^\text{r}_\text{vol}$, and 58.95\% AP$^\text{r}_\text{50}$. Table~\ref{table:mhpv2bench} summarizes 8 models on MHP-v2 \texttt{val} set. SCHP achieves the best mIoU again. In terms of AP$^\text{p}_\text{vol}$/AP$^\text{p}_\text{50}$, RP R-CNN \cite{yang2020renovating} has won the best results so far.

\begin{table}
	\centering
	\captionsetup{font=small}
	\caption{\small{\textbf{Quantitative MHP results on CIHP} \texttt{val} (\S\ref{sec:multi-bench}) in terms of mIoU, AP$^\text{r}_\text{vol}$ and AP$^\text{r}_\text{50}$. We only mark the best score in \textcolor{scorered}{\textbf{red}} color.}}
	\label{table:cihpbench}
	\vspace{-5pt}
	\begin{threeparttable}
	\resizebox{0.49\textwidth}{!}{
	\setlength\tabcolsep{2.5pt}
	\renewcommand\arraystretch{1.02}
	\begin{tabular}{c|r|c||c|c|c||c|c|c}
	\hline\thickhline
	\rowcolor{mygray1}
	Year & Method & Pub. & Pipeline &  Backbone  & \#Epoch  & mIoU & AP$^\text{r}_\text{vol}$ & AP$^\text{r}_\text{50}$   \\
	\hline
	\hline
	2018 & PGN \cite{gong2018instance} & ECCV & BU & ResNet101 &  \app80 & 55.80 & 33.60   &  35.80  \\
	\hline
	\multirow{4}{*}{{2019}} & CE2P \cite{ruan2019devil} & AAAI & 2S-TD & ResNet101 &  150 & 59.50 & 42.80  &  48.70  \\
		                            & Parsing R-CNN \cite{yang2019parsing} & CVPR & 1S-TD & ResNet50 &  75 & 56.30 & 36.50   &  40.90  \\
		                            & BraidNet \cite{liu2019braidnet} & MM & 2S-TD & ResNet101 &  150 & 60.62 & 43.59  &  48.99  \\
		                            & Unified \cite{qin2019top} & BMVC & 1S-TD & ResNet101 &  \app36 & 53.50 & 37.00  &  41.80  \\
	\hline	                            
	\multirow{3}{*}{{2020}} 
		                            & RP R-CNN \cite{yang2020renovating} & ECCV & 1S-TD & ResNet50 &  150 & 60.20 & 42.30  &  48.20  \\
		                            & SemaTree \cite{ji2020learning} & ECCV & 2S-TD & ResNet101 &  200 & 60.87 & 43.96  &  49.27  \\
		                            & SCHP \cite{li2020correction} & TPAMI & 2S-TD & ResNet101 &  150 &  \textcolor{scorered}{\textbf{67.47}} &  \textcolor{scorered}{\textbf{52.74}}    &  \textcolor{scorered}{\textbf{58.94}}   \\
	\hline
	\multirow{1}{*}{{2022}} 
		                            & AIParsing \cite{zhang2022aiparsing} & TIP & 1S-TD & ResNet101 &  75 & 60.70 & -   &  -  \\
	\hline
	\multirow{3}{*}{{2023}}  & HPSP \cite{li2023end} & TMM & BU & ResNet101 &  150 & 64.30 & -   &  -  \\
	                                     & ReSParser \cite{dai2023resparser} & TMM & 1S-TD & ResNet101 &  75 & 58.90 & -   &  -  \\
	                                     & CID \cite{wang2023contextual} & TPAMI & 1S-TD & HRNetW48 &  140 & 63.90 & -   &  -  \\
	
	\hline
	\end{tabular}
	}
	\end{threeparttable}
\end{table}

\begin{table}
	\centering
	\captionsetup{font=small}
	\caption{\small{\textbf{Quantitative MHP results on MHP-v2} \texttt{val} (\S\ref{sec:multi-bench}) in terms of mIoU, AP$^\text{p}_\text{vol}$ and AP$^\text{p}_\text{50}$. We only mark the best score in \textcolor{scorered}{\textbf{red}} color.}}
	\label{table:mhpv2bench}
	\vspace{-5pt}
	\begin{threeparttable}
	\resizebox{0.49\textwidth}{!}{
	\setlength\tabcolsep{2.5pt}
	\renewcommand\arraystretch{1.02}
	\begin{tabular}{c|r|c||c|c|c||c|c|c}
	\hline\thickhline
	\rowcolor{mygray1}
	Year & Method & Pub. & Pipeline &  Backbone  & \#Epoch  & mIoU & AP$^\text{p}_\text{vol}$ & AP$^\text{p}_\text{50}$   \\
	\hline
	\hline
	\multirow{2}{*}{{2019}} & CE2P \cite{ruan2019devil} & AAAI & 2S-TD & ResNet101 &  150 & 41.11 & 42.70  &  34.47  \\
		                            & Parsing R-CNN \cite{yang2019parsing} & CVPR & 1S-TD & ResNet50 &  75 & 36.20 & 38.50   &  24.50  \\
	\hline	                            
	\multirow{4}{*}{{2020}} & RP R-CNN \cite{yang2020renovating} & ECCV & 1S-TD & ResNet50 &  150 & 38.60 & \textcolor{scorered}{\textbf{46.80}}  &   \textcolor{scorered}{\textbf{45.30}}   \\
		                            & SemaTree \cite{ji2020learning} & ECCV & 2S-TD & ResNet101 &  200 & - & 42.51  &  34.36  \\
			                    & NAN \cite{zhao2020fine} & IJCV & BU & - &  80 & - & 41.78   &   25.14   \\
		                            & SCHP \cite{li2020correction} & TPAMI & 2S-TD & ResNet101 &  150 &  \textcolor{scorered}{\textbf{45.21}} &  45.25    &  35.10   \\
	\hline
        \multirow{1}{*}{{2021}} 
		                            & MGHR \cite{zhou2021differentiable, zhou2023differentiable} & \tabincell{c}{CVPR\\ TPAMI} & BU & ResNet101 &  150 & 41.40 & 44.30   &  39.00  \\
	\hline
	\multirow{1}{*}{{2022}} & AIParsing \cite{zhang2022aiparsing} & TIP & 1S-TD & ResNet101 &  75 & 40.10 & 46.60   &  43.20 \\
	\hline
	\multirow{3}{*}{{2023}} & HPSP \cite{li2023end} & TMM & BU & ResNet101 &  200 & 42.90 & 45.80   &  41.30  \\
	                                    & ReSParser \cite{dai2023resparser} & TMM & 1S-TD & ResNet101 &  75 & 35.40 & 42.70   &  34.30  \\
	                                    & CID \cite{wang2023contextual} & TPAMI & 1S-TD & HRNetW48 &  140 & 39.80 & 44.90   &  37.20  \\
	\hline
	\end{tabular}
	}
	\end{threeparttable}
	\vspace{-5pt}
\end{table}

\vspace{-6pt}
\subsection{VHP Performance Benchmarking}
\label{sec:video-bench}
VIP datasets is widely used to benchmark video human parsing. We selected 14 models since 2018.

\vspace{-6pt}
\subsubsection{Evaluation Metrics}
Similar to multiple human parsing, mIoU and  AP$^\text{r}_\text{vol}$ are also adopted for video human parsing performance evaluation.

\vspace{-6pt}
\subsubsection{Results}
Table~\ref{table:vipbench} gives the results of recent methods on VIP \texttt{val} set. It is clear that LIIR \cite{zhou2021differentiable} and UVC+ \cite{mckee2022transfer} have achieved the best performance in mIoU and AP$^\text{r}_\text{vol}$ metrics respectively.

\vspace{-6pt}
\subsection{Summary}
\label{sec:per-summary}
Through the above performance comparison, we can observe several apparent phenomena. The first and most important is the fairness of the experimental setting. For single human parsing and multiple human parsing, many studies have not given detailed experimental settings, or there are great differences in several essential hyper-parameters, resulting fair comparison impossible. The second is that most methods do not give the parameters number and the inference time, which makes some methods occupy an advantage in comparison by increasing the model capacity, and also brings trouble to some computationally sensitive application scenarios, such as social media and automatic driving.

In addition to the above phenomena, we can also summarize some positive signals. Firstly, in recent years, human parsing research has shown an upward trend, especially from 2020. Secondly, although some studies have achieved high performance on LIP, CIHP and VIP, these benchmarks are still not saturated. Thus the community still needs to continue its efforts. Thirdly, some specific issues and hotspots of human parsing are gradually attracting people's attention, which will further promote the progress of the whole field.

\begin{table}
	\centering
	\captionsetup{font=small}
	\caption{\small{\textbf{Quantitative VHP results on VIP} \texttt{val} (\S\ref{sec:multi-bench}) in terms of mIoU and AP$^\text{r}_\text{vol}$. The three best scores are marked in \textcolor{scorered}{\textbf{red}}, \textcolor{scoreblue}{\textbf{blue}}, and \textcolor{scoregreen}{\textbf{green}}, respectively.}}
	\label{table:vipbench}
	\vspace{-5pt}
	\begin{threeparttable}
	\resizebox{0.42\textwidth}{!}{
	\setlength\tabcolsep{2.5pt}
	\renewcommand\arraystretch{1.02}
	\begin{tabular}{c|r|c||c||c|c}
	\hline\thickhline
	\rowcolor{mygray1}
	Year & Method & Pub.   &  Backbone   & mIoU & AP$^\text{r}_\text{vol}$  \\
	\hline
	\hline
	\multirow{2}{*}{{2019}} & TimeCycle \cite{wang2019corres} & CVPR  & ResNet50  & 28.9 & 15.6    \\
		                            & UVC \cite{li2019joint} & NeurIPS & ResNet18  & 34.1 & 17.7    \\
	\hline
	\multirow{1}{*}{{2020}} & CRW \cite{jabri2020space} & NeurIPS & ResNet18  & 38.6 & -    \\
	\hline
	\multirow{4}{*}{{2021}} & ContrastCorr \cite{wang2021contrasive} & AAAI  & ResNet18  & 37.4 &  \textcolor{scoregreen}{\textbf{21.6}}    \\
	                                    & CLTC \cite{jeon2021mining} & CVPR & ResNet18  & 37.8 & 19.1    \\
		                            & VFS \cite{xu2021rethinking} & ICCV & ResNet18  &  39.9 & -    \\
		                            & JSTG \cite{zhao2021modelling} & ICCV & ResNet18  &  40.2  & -    \\
	\hline
	\multirow{4}{*}{{2022}} & LIIR \cite{li2022locality} & CVPR & ResNet18  & \textcolor{scorered}{\textbf{41.2}}  &  \textcolor{scoreblue}{\textbf{22.1}}    \\
	                                    & SCC \cite{son2022contrastive} & CVPR & ResNet18  & \textcolor{scoregreen}{\textbf{40.8}}  & -    \\
		                           & SFC \cite{hu2022semantic} & ECCV & ResNet18  & 38.4 & -     \\
		                           & UVC+ \cite{mckee2022transfer} & ArXiv & ResNet18  & 38.3 & \textcolor{scorered}{\textbf{22.2}}     \\
	\hline
    \multirow{3}{*}{{2023}} & STVC\cite{li2023spatial} & CVPR & ResNet18  & \textcolor{scoreblue}{\textbf{41.0}}  &  -    \\
				      & SMTC \cite{qian2023semantics} & ICCV & ResNet50  & 38.8 & -    \\
			  	      & SiamMAE \cite{gupta2023siamese} & ArXiv & ViT-S/16  & 37.3 & -     \\
	\hline
	\end{tabular}
	}
	\end{threeparttable}
	\vspace{-5pt}
\end{table}

\section{An Outlook: Future Opportunities of Human Parsing}
\label{sec:outlook}
After ten years of long development, with the whole community's efforts, human parsing has made remarkable achievements, but it has also encountered a bottleneck. In this section, we will discuss the opportunities of human parsing in the next era from multiple perspectives, hoping to promote progress in the field.

\begin{figure*}
	\begin{center}
		\includegraphics[width=0.90\linewidth]{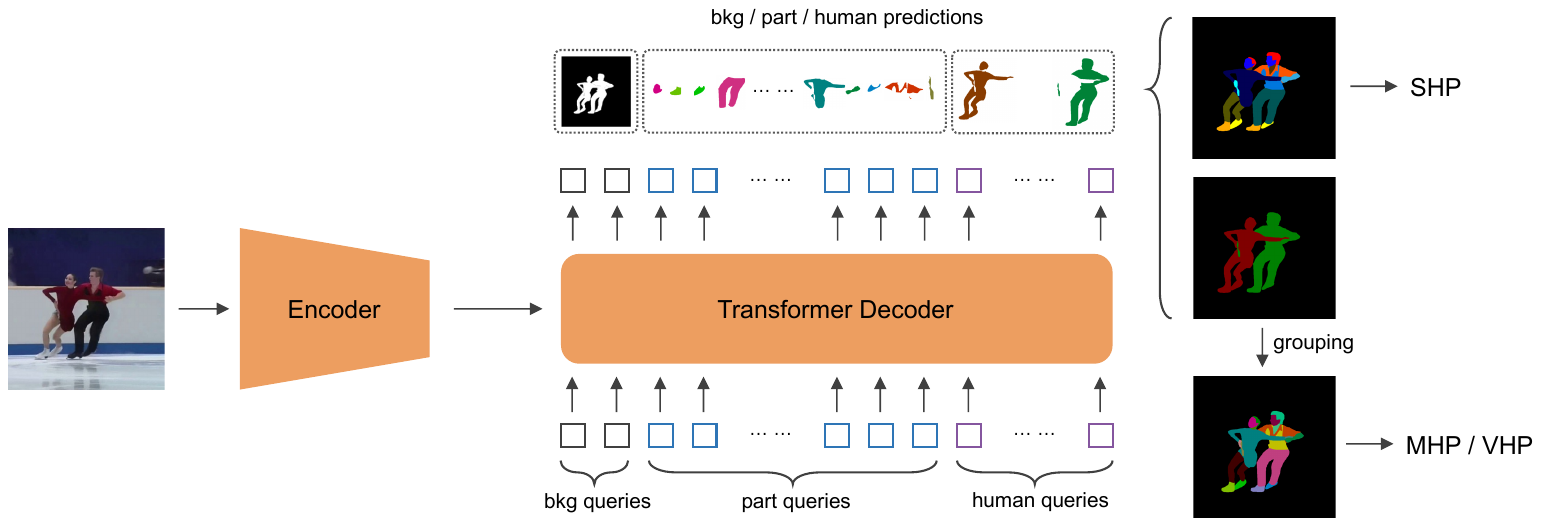}
	\end{center}
	\vspace{-5pt}
	\captionsetup{font=small}
	\caption{\small{\textbf{Architecture of the proposed M2FP} (\S\ref{sec:new-baseline}). Through the explicit construction of background, part and human queries, we can model the relationship between humans and parts, and predict high-quality masks.}}
	\label{fig:m2fp_arch}
	\vspace{-10pt}
\end{figure*}

\vspace{-6pt}
\subsection{A Transformer-based Baseline for Human Parsing}
\label{sec:new-baseline}
Although several mainstream benchmarks of human parsing have not been saturated, the accuracy growth has slowed down. The reason for this, we believe, is that some advances in deep learning have not yet benefited the human parsing task (\emph{e.g}., transformer \cite{vaswani2017attention, dosovitskiy2020an, carion2020end}, unsupervised representation learning \cite{devlin2019bert, he2020momentum, bao2022beit, he2022masked}), and the lack of a concise and easily extensible code base for researchers. Therefore, the community urgently needs a new and strong baseline.

We consider that a new human parsing baseline should have the following four characteristics: a) \textbf{Universality}, which can be applied to all mainstream human parsing tasks, including SHP, MHP, and VIP; b) \textbf{Conciseness}, the baseline method should not be too complex; c) \textbf{Extensibility}, complete code base, easy to modify or expand other modules or methods; d) \textbf{High performance}, state-of-the-arts or at least comparable performance can be achieved on the mainstream benchmarks under the fair experimental setting. Based on the above views, we design a new transformer-based baseline for human parsing. The proposed new baseline is based on the \texttt{Mask2Former} \cite{cheng2022masked} architecture, with a few improvements adapted to human parsing, called \texttt{Mask2Former for Parsing (M2FP)}. M2FP can adapt to almost all human parsing tasks and yield amazing performances.

\vspace{-6pt}
\subsubsection{A Brief Review of Mask2Former}
Mask2Former is a universal image segmentation method, which achieves state-of-the-art on common image segmentation tasks (\emph{i.e}., panoptic, instance, and semantic). The main idea of Mask2Former is to introduce mask classification \cite{cheng2021perpixel}, masked attention and set prediction objective \cite{carion2020end}. The combination of these three advantages can realize end-to-end high-performance universal image segmentation. Mask2Former is verified on several image/video segmentation benchmarks, including COCO panoptic segmentation \cite{kirillov2019panoptic}, COCO instance segmentation \cite{lin2014microsoft}, ADE20K semantic segmentation \cite{zhou2017scene}, YouTubeVIS video instance segmentation \cite{yang2019video, cheng2021m2forvis} and so on \cite{cordts2016cityscapes, neuhold2017the}.

\vspace{-6pt}
\subsubsection{Mask2Former for Parsing}
\noindent$\bullet$~\textbf{Modeling Human as Group Queries.} To solve the three human parsing sub-tasks, we need to simultaneously model the parts relationship and distinguish human instances. DETR series work \cite{carion2020end, zhu2021deformable, cheng2021perpixel, cheng2022masked} regard objects as queries, and transform object detection or instance segmentation task into a direct set prediction problem. A naive idea is to regard human parts as queries, then use mask classification to predict the category and mask of each part. However, this creates two problems that cannot be ignored. Firstly, only modeling parts will make it difficult to learn the global relationship between parts and humans; Secondly, the subordination between part and human instance is unknown, resulting in the inadaptability for MHP task. Thus, we introduce the body hierarchy into the queries and use the powerful sequence encoding ability of transformer to build multiple hierarchical relationships between parts and humans. Specifically, we explicitly divide the queries into three groups: background queries, part queries and human queries. Through the relationship modeling ability of self-attention mechanism, besides the basic part-part relationship, the part-human, human-human, and part/human-background relationships are also modeled. Thanks to the direct modeling of parts and the introduction of multiple hierarchical granularities, M2FP can be applied to all supervised human parsing tasks.

\noindent$\bullet$~\textbf{Architecture and Pipeline.} The architecture of proposed M2FP is illustrated in Figure~\ref{fig:m2fp_arch}. We try to make the smallest modification to the Mask2Former. An encoder is used to extract image or video features, which is composed of a backbone and a pixel decoder \cite{zhu2021deformable}. Then the features are flattened and sent into a transformer decoder. The transformer decoder consists of multiple repeated units, each containing a masked attention module, a self-attention module, and a shared feed-forward network (FFN) in turn. The grouped queries and flattened features conduct sufficient information exchange through the transformer decoder, and finally use bipartite matcher to match between queries and ground-truths uniquely. For SHP, in the inference stage, the background and part masks are combined with their class predictions to compute the final semantic segmentation prediction through matrix multiplication. For MHP, the intersection ratio of semantic segmentation prediction and human masks is calculated to obtain the final instance-level human parsing prediction. M2FP can also be extended to supervised VHP task. Follow \cite{cheng2021m2forvis}, the background, parts, and humans in the video can be regarded as 3D spatial-temporal masks, and using the sequence encoding ability of transformer to make an end-to-end prediction.

\vspace{-6pt}
\subsubsection{Experiments}
\noindent$\bullet$~\textbf{Experimental Setup.} We validate M2FP on several mainstream benchmarks, including LIP, PASCAL-Person-Part, CIHP, and MHP-v2. All models are trained with nearly identical hyper-parameters under 8 NVIDIA V100 GPUs. Specifically, we use AdamW \cite{loshchilov2018decoupled} optimizer with a mini-batch size of 16, an initial learning rate of 0.0004 with poly (LIP) or step (PASCAL-Person-Part, CIHP, and MHP-v2) learning rate schedule, then train each model for 150 epochs. Large scale jittering in the range of [0.1, 2.0] and typical data augmentation techniques, \emph{e.g}., fixed size random crop (512$\times$384 for LIP, 800$\times$800 for PASCAL-Person-Part, CIHP, and MHP-v2), random rotation from [-40\degree, +40\degree], random color jittering and horizontal flip, are also used. For fair comparison, horizontal flipping is adopted during testing, and multi-scale test is used for LIP. The default backbone is ResNet-101 with pre-training on ImageNet-1K \cite{russakovsky2015imagenet}.

\begin{figure}
	\begin{center}
 		\includegraphics[width=0.99\linewidth]{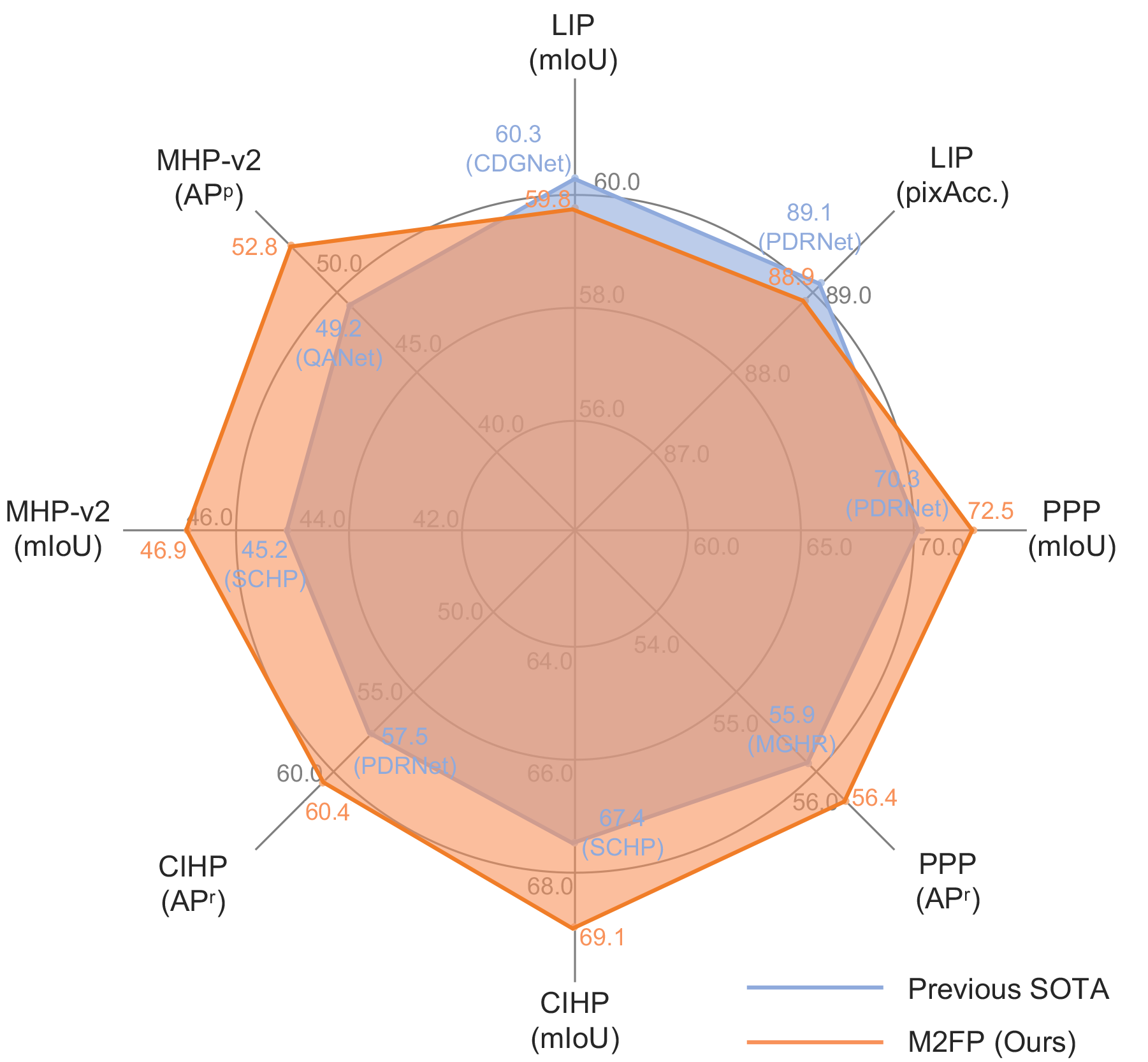}
 	\end{center}
 	\vspace{-5pt}
 	\captionsetup{font=small}
 	\caption{\small{\textbf{Comparison of M2FP with previous human parsing state-of-the-art models}. M2FP achieves state-of-the-art (PPP, CIHP and MHP-v2) or comparable performance (LIP) on all human parsing sub-tasks.}}
 	\label{fig:m2fp_performance}
 	\vspace{-5pt}
  \end{figure}

\begin{table}
	\centering
	\captionsetup{font=small}
	\caption{\small{\textbf{Overview of M2FP results on various human parsing benchmarks}. \underline{~~~~} denotes the previous state-of-the-art results. \textbf{Bold} results denote M2FP achieve new state-of-the-art.}}
	\label{table:m2fp_compare}
	\vspace{-5pt}
	\begin{threeparttable}
	\resizebox{0.49\textwidth}{!}{
	\setlength\tabcolsep{2.5pt}
	\renewcommand\arraystretch{1.02}
	\begin{tabular}{r||c||c||c|c||c|c||c|c}
	\hline\thickhline
	\rowcolor{mygray1}
	                                                                &   \multicolumn{2}{c||}{LIP}                                           & \multicolumn{2}{c||}{PPP}                             & \multicolumn{2}{c||}{CIHP}                              &     \multicolumn{2}{c}{MHP-v2}    \\
	\cline{2-9}                               
	\rowcolor{mygray1}                                  
	 \multirow{-2}{*}{Method}                         &   pixAcc.                             &   mIoU                             &      mIoU     &     AP$^\text{r}_\text{vol}$         &      mIoU               &      AP$^\text{r}_\text{vol}$      &  mIoU &  AP$^\text{p}_\text{vol}$   \\
	\hline
	\hline
        HIPN \cite{liu2021hier}                           &        \underline{89.14}          &    59.61                            &      -                                  &      -                        &       -                       &     -                                           &        -                   &            -                       \\
	HSSN \cite{li2022deep}                           &       -                                     &     \underline{60.37}        &       -                                 &       -                       &        -                     &          -                                      &       -                    &             -                      \\
	PGN \cite{gong2018instance}                 &          -                                  &     -                                    &     \underline{68.40}        &  39.20                   &   55.80                  &       33.60                               &        -                    &               -                    \\
	MGHR \cite{zhou2021differentiable, zhou2023differentiable}       &         -                                   &    -                                     &       -                                 &  \underline{55.90} &    -                         &             -                                   &   41.40                &         44.30                 \\
	SCHP                   \cite{li2020correction} &           -                                 &    -                                     &       -                                 &     -                         & \underline{67.47}  &        \underline{52.74}            &\underline{45.21}&         45.25                 \\
	RP R-CNN \cite{yang2020renovating}    &              -                              &    -                                     &   63.30                            &     40.90                & 60.20                     &        42.30                              &  38.60                &         \underline{46.80} \\
	\hline   
	M2FP (ours)                                            &      88.93                             &      59.86                          &      \textbf{72.54}             &  \textbf{56.46}    &\textbf{69.15}&        \textbf{60.47}                  &\textbf{46.94} &         \textbf{52.82}    \\  
	\hline                               
	\end{tabular}
	}
	\end{threeparttable}
	\vspace{-5pt}
\end{table}

\noindent$\bullet$~\textbf{Main Results.} As shown in Table~\ref{table:m2fp_compare}, and Figure~\ref{fig:m2fp_performance}, M2FP achieves state-of-the-art or comparable performance across a broad range of human parsing benchmarks. For SHP, M2FP only falls behind HIPN \cite{liu2021hier} and CDGNet \cite{liu2022cdgnet}, obtaining 88.93\% pixAcc. and 59.86\% mIoU, showing great potential in the parts relationship modeling. For MHP, M2FP shows amazing performance, greatly surpassing the existing methods on all metrics and even exceeding the state-of-the-art two-stage top-down method, \emph{i.e}., SCHP \cite{li2020correction}. Specifically, M2FP outperforms PGN \cite{gong2018instance} with 4.14 point mIoU and MGHR \cite{zhou2021differentiable, zhou2023differentiable} with 0.56 point AP$^\text{r}_\text{vol}$ on PASCAL-Person-Part. On the more challenging CIHP and MHP-v2, M2FP beats SCHP in terms of mIoU while running in an end-to-end manner. Meanwhile, M2FP is also 7.73 points ahead of SCHP in AP$^\text{r}_\text{vol}$ (CIHP) and 6.02 points ahead of RP R-CNN \cite{yang2020renovating} in AP$^\text{p}_\text{vol}$ (MHP-v2). These results demonstrate that M2FP surpasses almost all human parsing methods in a concise, effective and universal way, and can be regarded as a new baseline in the next era. 

\noindent$\bullet$~\textbf{Ablation Study.} We also show the impact of different types of queries on PPP dataset in Table~\ref{table:m2fp_ablation}. When only retaining the part queries (Table~\ref{table:m2fp_ablation} (a)), M2FP is equivalent to naive Mask2Former, and we can adopt the heuristic greedy algorithm to generate human part segmentation results, yielding 71.35\% mIoU and 54.25\% AP$^\text{r}_\text{vol}$. Adding the background queries (Table~\ref{table:m2fp_ablation} (b)) can eliminate the heuristic greedy algorithm and achieve a slight performance improvement. Continuing to incorporate the human queries (Table~\ref{table:m2fp_ablation} (c)), which is the proposed M2FP, shows a significant performance improvement, particularly with 1.19 points mIoU and 2.21 points AP$^\text{r}_\text{vol}$ improvement compared to (a). This indicates that modeling human as group queries is a concise and effective approach, making the powerful Mask2Former architecture suitable for human parsing tasks.

\begin{table}
	\centering
	\captionsetup{font=small}
	\caption{\small{\textbf{Ablation study on the impact of different types of queries on PPP dataset}. When lacking background queries, the heuristic greedy algorithm is used to generate human parts segmentation results.}}
	\label{table:m2fp_ablation}
	\vspace{-5pt}
	\begin{threeparttable}
	\resizebox{0.49\textwidth}{!}{
	\setlength\tabcolsep{2.5pt}
	\renewcommand\arraystretch{1.02}
	\begin{tabular}{c|c|c|c||c|c}
	\hline\thickhline
	\rowcolor{mygray1}
                                                                      &       part queries       &    background queries      &     human queries   &      mIoU       &     AP$^\text{r}_\text{vol}$     \\      
	\hline
	\hline 
           (a)   &                                          \checkmark       &                                         &                                & 71.35(\textcolor{scorered}{-1.19})            &    54.25(\textcolor{scorered}{-2.21})                                          \\
            (b)   &                                         \checkmark       &  \checkmark                     &                                & 71.81(\textcolor{scorered}{-0.73})            &    54.33(\textcolor{scorered}{-2.13})                                   \\
            (c)   &                                         \checkmark      &      \checkmark                 &      \checkmark         & \textbf{72.54}             &  \textbf{56.46}                                 \\  
	\hline                               
	\end{tabular}
	}
	\end{threeparttable}
	\vspace{-5pt}
\end{table}


\vspace{-6pt}
\subsection{Under-Investigated Open Issues}
\label{sec:open-issues}
Based on the reviewed research, we list several under-investigated open issues that we believe should be pursued.

\noindent$\bullet$~\textbf{Efficient Inference.} In practical applications, human parsing models generally need real-time or even faster inference speed. The current research has not paid enough attention to this issue, especially the multiple human parsing research. Although some literature \cite{zhou2021differentiable, zhang2022aiparsing} has discussed the model efficiency, it can not achieve real-time inference, and there is no human parser designed for this purpose. Therefore, from the perspective of practical application, it is an under-investigated open issue to design an efficient inference human parsing model.

\noindent$\bullet$~\textbf{Synthetic Dataset.} It is a common practice in many fields to use synthetic datasets to train models and transfer them to real scenes. Through CG technology (\emph{e.g}., NVIDIA Omniverse\footnote{\fontsize{7pt}{1em}\url{https://developer.nvidia.com/nvidia-omniverse}}), we can obtain almost unlimited synthetic human data at a very low cost, as well as parsing annotations. Considering the labeling cost of human parsing dataset, this is a very attractive scheme. Wood \emph{et al}. have made a preliminary attempt on the face parsing task and achieved very excellent performance \cite{wood2021fake}, but at present, there is a lack of research on the human parsing field.

\noindent$\bullet$~\textbf{Long-tailed Phenomenon.} The long-tailed distribution is the most common phenomenon in the real world, and also exists in the human parsing field. For example, the Gini coefficient of MHP-v2.0 is as high as 0.747 \cite{yang2022longtailed}, exceeding some artificially created long-tailed datasets, but this problem is currently ignored. Therefore, the existing methods are often brittle once exposed to the real world, where they are unable to adapt and robustly deal with tail categories effectively. This calls for a more general human parsing model, with the ability to adapt to long-tailed distributions in the real world.

\noindent$\bullet$~\textbf{Interpretability.} Interpretability affects people's trust in deep learning systems. Human parsing is an important visual perception technology that expresses the temporal-spatial attribute of human body in the real world, and its interpretability can bring significant impact to many applications, \emph{e.g}., security monitoring, autonomous driving, social media, \emph{etc}. Yu \emph{et al}. employ capsule network to establish an unsupervised face part discovery system \cite{yu2022hp-capsule}, which partly reveal the interpretability of face parsing. However, the research on human parsing interpretability is still vacant. It is a non-negligible issue that must be investigated to build a trustworthy human visual perception system.

\vspace{-6pt}
\subsection{New Directions}
\label{sec:new-direc}
Considering some potential applications, we shed light on several possible research directions.

\noindent$\bullet$~\textbf{Video Instance-level Human Parsing.} The current VHP research basically follows an unsupervised semi-automatic video object segmentation setting, which reduces the labeling cost in a way that greatly loses accuracy. However, most of the practical requirements of video human parsing require extremely high precision. Therefore, making full use of annotations and solving the VHP issue through an instance-discriminative manner, \emph{i.e}., a fine-grained video instance segmentation task, has great research prospects.

\noindent$\bullet$~\textbf{Panoptic Parts Parsing.} Panoptic Parts Parsing, or Part-aware Panoptic Segmentation \cite{de2021part}, is a new issue recently proposed. This task aims to simultaneously understand a scene at two levels of abstraction: scene parsing and part parsing. At present, the research on this issue is still at an early stage. We consider that in-depth research on Panoptic Parts Parsing can reveal how humans perceive the scene at a deeper level.

\noindent$\bullet$~\textbf{Whole-body Human Parsing.} Besides human parsing, face parsing and hand parsing \cite{liang2014parsing, lin2019face} are also important issues. To fully understand the pixel-wise temporal-spatial attributes of human in the wild, it is necessary to parse body, face, and hands simultaneously, which implies a new direction to end-to-end parse the whole body: Whole-body Human Parsing. Natural hierarchical annotation and large-scale variation bring new challenges to existing parsing techniques. Thus the targeted datasets and whole-body parsers are necessary.

\noindent$\bullet$~\textbf{3D Human Parsing.} Due to the popularity of 3D sensors (\emph{e.g}., LIDAR and depth-sensing cameras), 3D human parsing \cite{yu2020humbi, tang2021motion} has gradually become a new focus. Its purpose is to predict each point in the point cloud to partition the human body into semantic parts. Unlike (2D) human parsing, 3D human parsing requires processing of irregular point cloud data, so algorithms designed based on 2D human parsing are not applicable. Therefore, research on 3D human parsing with point clouds is still in its infancy.

\noindent$\bullet$~\textbf{Cooperation across Different Human-centric Directions.} Some human-centric visual tasks (\emph{e.g}., human attribute recognition \cite{yang2020hier}, pose estimation \cite{zheng2023deep}, human mesh reconstruction \cite{guler2019holopose}) face similar challenges to human parsing. Different tasks can play a positive role in promoting each other, although developments of these fields are independent. Moreover, the settings of different human-centric visual tasks are related, while there are no precedents for modeling these tasks in a unified framework. Thus, we call for closer collaboration across different human-centric visual tasks.

\vspace{-6pt}
\subsection{Human Parsing in Foundation Models Era}
\label{sec:hp-in-fme}
In recent years, the most exciting developments in artificial intelligence have undoubtedly been foundation models (Large Language \cite{radford2019language, brown2020language} / Vision \cite{oquab2023dinov2, kirillov2023segment} / Multimodal Models \cite{radford2021learning, rombach2022high, girdhar2023imagebind, betker2023improving}). These methods have greatly influenced classical machine learning and also brought many challenges and opportunities to human parsing.

\noindent$\bullet$~\textbf{Challenges.} The foundation models have emerged with a large number of capabilities that conventional models do not have, resulting in significant challenges for conventional task-driven small models. On the one hand, large vision models (\emph{e.g}., DINO \cite{caron2021emerging, oquab2023dinov2} and SAM \cite{kirillov2023segment}) exhibit impressive zero-shot segmentation capability, which means that supervised learning human parsing may not be a generalized solution. We need to consider new methods to address the human parsing problems of the new era, for dealing with massive amounts of out-of-domain data and categories. On the other hand, contrastive learning aligns the feature space of images and natural language \cite{radford2021learning}, and the powerful multimodal features enhance the network's capability to handle issues such as zero-shot, few-shot, and long-tailed phenomena. However, we have not yet seen human parsing tasks benefit from multimodal representation.

\noindent$\bullet$~\textbf{Opportunities.} Fortunately, challenges always bring opportunities. Large vision models bring a more universal representation, and human parsing should be considered as part of the human-centric visual understanding to seek more unified solutions. Firstly, human-centric pre-training foundation models have become possible \cite{chen2023beyond}, which will directly assist numerous downstream tasks (\emph{e.g}., human parsing, pose estimation, and person re-identification) to improve generalization or reduce the required labels during fine-tuning. Secondly, exploiting human-centric homogeneity to design a universal model has also begun to showcase its advantages \cite{ci2023unihcp, tang2023humanbench} , outputting several predictions including human parsing in an end-to-end manner. This has great significance for exploring the promotion or inhibition relationships between different human-centric visual tasks, and for learning more universal human visual representation. In addition, aligning visual and language embeddings of human parts has also become possible in the era of foundation models. Prompt-based generative models can enrich scope of human parsing applications, \emph{e.g}., combining ControlNet \cite{zhang2023adding} to control human image/video generation, or as a visual prompting method to unleash the visual grounding abilities of large multimodal models \cite{yang2023set} (such as GPT-4V).

\section{Conclusions}
\label{sec:concl}

As far as we know, this is the first survey to comprehensively review deep learning techniques in human parsing, covering three sub-tasks: SHP, MHP, and VHP. We first provided the readers with the necessary knowledge, including task settings, background concepts, relevant problems, and applications. Afterward, we summarized the mainstream deep learning methods based on human parsing taxonomy, and analyzing them according to the theoretical background, technical contributions, and solving strategies. We also reviewed 19 popular human parsing datasets, benchmarking results on the 6 most widely-used ones. To promote sustainable community development, we analyzed the under-investigated open issues, provided insight into new directions, and discussed the challenges and opportunities of human parsing in the foundation models era. We also put forward a new transformer-based human parsing framework, servicing a high-performance baseline for follow-up research through universal, concise, and extensible solutions. In summary, we hope this survey to provide an effective way to understand the current state-of-the-art human parsing models and promote the sustainable development of this research field.

\begin{acknowledgements}
This work was supported by the China National Postdoctoral Program for Innovative Talents (No. BX2021047), China Postdoctoral Science Foundation (No. 2022M710466), and Young Scientists Fund of NSFC (Grant No. 62206025).
\end{acknowledgements}

%
%

\bibliographystyle{spmpsci}      
\bibliography{egbib}   

\begin{thebibliography}{100}
\providecommand{\url}[1]{{#1}}
\providecommand{\urlprefix}{URL }
\expandafter\ifx\csname urlstyle\endcsname\relax
  \providecommand{\doi}[1]{DOI~\discretionary{}{}{}#1}\else
  \providecommand{\doi}{DOI~\discretionary{}{}{}\begingroup
  \urlstyle{rm}\Url}\fi

\bibitem{bao2022beit}
Bao, H., Dong, L., Piao, S., Wei, F.: Beit: Bert pre-training of image
  transformers.
\newblock In: Proceedings of the International Conference on Learning
  Representations (2022)

\bibitem{betker2023improving}
Betker, J., Goh, G., Jing, L., Brooks, T., Wang, J., Li, L., Ouyang, L.,
  Zhuang, J., Lee, J., Guo, Y., Manassra, W., Dhariwal, P., Chu, C., Jiao, Y.:
  Improving image generation with better captions.
\newblock OpenAI blog  (2023)

\bibitem{bo2011shape}
Bo, Y., Fowlkes, C.C.: Shape-based pedestrian parsing.
\newblock In: Proceedings of the IEEE Conference on Computer Vision and Pattern
  Recognition, pp. 2265--2272 (2011)

\bibitem{borras2003high}
Borras, A., Tous, F., Llados, J., Vanrell, M.: High-level clothes description
  based on colour-texture and structural features.
\newblock In: Iberian Conference on Pattern Recognition and Image Analysis, pp.
  108--116 (2003)

\bibitem{brown2020language}
Brown, T., Mann, B., Ryder, N., Subbiah, M., Kaplan, J.D., Dhariwal, P.,
  Neelakantan, A., Shyam, P., Sastry, G., Askell, A., et~al.: Language models
  are few-shot learners.
\newblock In: Advances in Neural Information Processing Systems, vol.~33, pp.
  1877--1901 (2020)

\bibitem{carion2020end}
Carion, N., Massa, F., Synnaeve, G., Usunier, N., Kirillov, A., Zagoruyko, S.:
  End-to-end object detection with transformers.
\newblock In: Proceedings of the European Conference on Computer Vision, pp.
  213--229 (2020)

\bibitem{caron2018deep}
Caron, M., Bojanowski, P., Joulin, A., Douze, M.: Deep clustering for
  unsupervised learning of visual features.
\newblock In: Proceedings of the European Conference on Computer Vision, pp.
  139--156 (2018)

\bibitem{caron2021emerging}
Caron, M., Touvron, H., Misra, I., Jegou, H., Mairal, J., Bojanowski, P.,
  Joulin, A.: Emerging properties in self-supervised vision transformers.
\newblock In: Proceedings of the IEEE/CVF International Conference on Computer
  Vision, pp. 9650--9660 (2021)

\bibitem{chang2022pfvton}
Chang, Y., Peng, T., He, R., Hu, X., Liu, J., Zhang, Z., Jiang, M.: Pf-vton:
  Toward high-quality parser-free virtual try-on network.
\newblock In: International Conference on Multimedia Modeling, pp. 28--40
  (2022)

\bibitem{chen2006composite}
Chen, H., Xu, Z., Liu, Z., Zhu, S.C.: Composite templates for cloth modeling
  and sketching.
\newblock In: Proceedings of the IEEE Conference on Computer Vision and Pattern
  Recognition, pp. 943--950 (2006)

\bibitem{chen2017deeplab}
Chen, L.C., Papandreou, G., Kokkinos, I., Murphy, K., Yuille, A.L.: Deeplab:
  Semantic image segmentation with deep convolutional nets, atrous convolution,
  and fully connected crfs.
\newblock IEEE Transactions on Pattern Analysis and Machine Intelligence
  \textbf{40}(4), 834--848 (2017)

\bibitem{chen2016attenttion}
Chen, L.C., Yang, Y., Wang, J., Xu, W., Yuille, A.L.: Attention to scale:
  Scale-aware semantic image segmentation.
\newblock In: Proceedings of the IEEE Conference on Computer Vision and Pattern
  Recognition, pp. 3640--3649 (2016)

\bibitem{chen2018semantic}
Chen, Q., Ge, T., Xu, Y., Zhang, Z., Yang, X., Gai, K.: Semantic human matting.
\newblock In: Proceedings of the 26th ACM International Conference on
  Multimedia, pp. 618--626 (2018)

\bibitem{chen2020simswap}
Chen, R., Chen, X., Ni, B., Ge, Y.: Simswap: An efficient framework for high
  fidelity face swapping.
\newblock In: Proceedings of the 28th ACM International Conference on
  Multimedia, pp. 2003--2011 (2020)

\bibitem{chen2023virtual}
Chen, S., Wang, J.: Virtual reality human--computer interactive english
  education experience system based on mobile terminal.
\newblock International Journal of Human--Computer Interaction pp. 1--10 (2023)

\bibitem{chen2023beyond}
Chen, W., Xu, X., Jia, J., Luo, H., Wang, Y., Wang, F., Jin, R., Sun, X.:
  Beyond appearance: a semantic controllable self-supervised learning framework
  for human-centric visual tasks.
\newblock In: Proceedings of the IEEE/CVF Conference on Computer Vision and
  Pattern Recognition, pp. 15050--15061 (2023)

\bibitem{chen2014detect}
Chen, X., Mottaghi, R., Liu, X., Fidler, S., Urtasun, R., Yuille, A.: Detect
  what you can: Detecting and representing objects using holistic models and
  body parts.
\newblock In: Proceedings of the IEEE Conference on Computer Vision and Pattern
  Recognition, pp. 1971--1978 (2014)

\bibitem{chen2019instance}
Chen, Y., Zhu, X., Gong, S.: Instance-guided context rendering for cross-domain
  person re-identification.
\newblock In: Proceedings of the IEEE/CVF International Conference on Computer
  Vision, pp. 232--242 (2019)

\bibitem{cheng2019spgnet}
Cheng, B., Chen, L.C., Wei, Y., Zhu, Y., Huang, Z., Xiong, J., Huang, T.S.,
  Hwu, W.M., Shi, H.: Spgnet: Semantic prediction guidance for scene parsing.
\newblock In: Proceedings of the IEEE/CVF International Conference on Computer
  Vision, pp. 5218--5228 (2019)

\bibitem{cheng2021m2forvis}
Cheng, B., Choudhuri, A., Misra, I., Kirillov, A., Girdhar, R., Schwing, A.G.:
  Mask2former for video instance segmentation.
\newblock arXiv preprint arXiv:2112.10764  (2021)

\bibitem{cheng2022masked}
Cheng, B., Misra, I., Schwing, A.G., Kirillov, A., Girdhar, R.:
  Masked-attention mask transformer for universal image segmentation.
\newblock In: Proceedings of the IEEE/CVF Conference on Computer Vision and
  Pattern Recognition (2022)

\bibitem{cheng2021perpixel}
Cheng, B., Schwing, A.G., Kirillov, A.: Per-pixel classification is not all you
  need for semantic segmentation.
\newblock In: Advances in Neural Information Processing Systems, pp.
  17864--17875 (2021)

\bibitem{cheng2021fashion}
Cheng, W., Song, S., Chen, C.Y., Hidayati, S.C., Liu, J.: Fashion meets
  computer vision: A survey.
\newblock ACM Computing Surveys \textbf{54}(4), 1--41 (2021)

\bibitem{ci2023unihcp}
Ci, Y., Wang, Y., Chen, M., Tang, S., Bai, L., Zhu, F., Zhao, R., Yu, F., Qi,
  D., Ouyang, W.: Unihcp: A unified model for human-centric perceptions.
\newblock In: Proceedings of the IEEE/CVF Conference on Computer Vision and
  Pattern Recognition., pp. 17840--17852 (2023)

\bibitem{cordts2016cityscapes}
Cordts, M., Omran, M., Ramos, S., Rehfeld, T., Enzweiler, M., Benenson, R.,
  Franke, U., Roth, S., Schiele, B.: The cityscapes dataset for semantic urban
  scene understanding.
\newblock In: Proceedings of the IEEE Conference on Computer Vision and Pattern
  Recognition, pp. 3213--3223 (2016)

\bibitem{dai2023resparser}
Dai, Y., Chen, X., Wang, X., Pang, M., Gao, L., Shen, H.T.: Resparser: Fully
  convolutional multiple human parsing with representative sets.
\newblock IEEE Transactions on Multimedia  (2023)

\bibitem{devlin2019bert}
Devlin, J., Chang, M.W., Lee, K., Toutanova, K.: Bert: Pre-training of deep
  bidirectional transformers for language understanding.
\newblock In: Proceedings of the Annual Conference of the North American
  Chapter of the Association for Computational Linguistics: Human Language
  Technologies, pp. 4171--4186 (2019)

\bibitem{dong2019towards}
Dong, H., Liang, X., Shen, X., Wang, B., Lai, H., Zhu, J., Hu, Z., Yin, J.:
  Towards multi-pose guided virtual try-on network.
\newblock In: Proceedings of the IEEE/CVF International Conference on Computer
  Vision, pp. 9026--9035 (2019)

\bibitem{dong2014towards}
Dong, J., Chen, Q., Shen, X., Yang, J., Yan, S.: Towards unified human parsing
  and pose estimation.
\newblock In: Proceedings of the IEEE Conference on Computer Vision and Pattern
  Recognition, pp. 843--850 (2014)

\bibitem{dong2013deformable}
Dong, J., Chen, Q., Xia, W., Huang, Z., Yan, S.: A deformable mixture parsing
  model with parselets.
\newblock In: Proceedings of the IEEE International Conference on Computer
  Vision, pp. 3408--3415 (2013)

\bibitem{dosovitskiy2020an}
Dosovitskiy, A., Beyer, L., Kolesnikov, A., Weissenborn, D., Zhai, X.,
  Unterthiner, T., Dehghani, M., Minderer, M., Heigold, G., Gelly, S.,
  Uszkoreit, J., Houlsby, N.: An image is worth 16x16 words: Transformers for
  image recognition at scale.
\newblock In: Proceedings of the International Conference on Learning
  Representations (2020)

\bibitem{everingham2010pascal}
Everingham, M., Van~Gool, L., Williams, C.K., Winn, J., Zisserman, A.: The
  pascal visual object classes (voc) challenge.
\newblock International Journal of Computer Vision \textbf{88}(2), 303--338
  (2010)

\bibitem{fang2018weakly}
Fang, H.S., Lu, G., Fang, X., Xie, J., Tai, Y.W., Lu, C.: Weakly and semi
  supervised human body part parsing via pose-guided knowledge transfer.
\newblock In: Proceedings of the IEEE Conference on Computer Vision and Pattern
  Recognition, pp. 70--78 (2018)

\bibitem{fang2020densely}
Fang, J., Sun, Y., Zhang, Q., Li, Y., Liu, W., Wang, X.: Densely connected
  search space for more flexible neural architecture search.
\newblock In: Proceedings of the IEEE/CVF Conference on Computer Vision and
  Pattern Recognition, pp. 10628--10637 (2020)

\bibitem{fruhstuck2022insetgan}
Fruhstuck, A., Singh, K.K., Shechtman, E., Mitra Niloy, J., Wonka, P., Lu, J.:
  Insetgan for full-body image generation.
\newblock arXiv preprint arXiv:2203.07293  (2022)

\bibitem{fulkerson2009class}
Fulkerson, B., Vedaldi, A., Soatto, S.: Class segmentation and object
  localization with superpixel neighborhoods.
\newblock In: Proceedings of the IEEE International Conference on Computer
  Vision, pp. 670--677 (2009)

\bibitem{gao2023dynamic}
Gao, Y., Lang, C., Liu, F., Cao, Y., Sun, L., Wei, Y.: Dynamic interaction
  dilation for interactive human parsing.
\newblock IEEE Transactions on Multimedia  (2023)

\bibitem{gao2022clicking}
Gao, Y., Liang, L., Lang, C., Feng, S., Li, Y., Wei, Y.: Clicking matters:
  Towards interactive human parsing.
\newblock IEEE Transactions on Multimedia  (2022)

\bibitem{ge2019deepfashion2}
Ge, Y., Zhang, R., Wang, X., Tang, X., Luo, P.: Deepfashion2: A versatile
  benchmark for detection, pose estimation, segmentation and re-identification
  of clothing images.
\newblock In: Proceedings of the IEEE/CVF Conference on Computer Vision and
  Pattern Recognition, pp. 5337--5345 (2019)

\bibitem{de2021part}
de~Geus, D., Meletis, P., Lu, C., Wen, X., Dubbelman, G.: Part-aware panoptic
  segmentation.
\newblock In: Proceedings of the IEEE/CVF Conference on Computer Vision and
  Pattern Recognition, pp. 5485--5494 (2021)

\bibitem{girdhar2023imagebind}
Girdhar, R., El-Nouby, A., Liu, Z., Singh, M., Alwala, K.V., Joulin, A., Misra,
  I.: Imagebind: One embedding space to bind them all.
\newblock In: Proceedings of the IEEE/CVF Conference on Computer Vision and
  Pattern Recognition, pp. 15180--15190 (2023)

\bibitem{girshick2014rich}
Girshick, R., Donahue, J., Darrell, T., Malik, J.: Rich feature hierarchies for
  accurate object detection and semantic segmentation.
\newblock In: Proceedings of the IEEE Conference on Computer Vision and Pattern
  Recognition, pp. 580--587 (2014)

\bibitem{gong2019graphonomy}
Gong, K., Gao, Y., Liang, X., Shen, X., Wang, M., Lin, L.: Graphonomy:
  Universal human parsing via graph transfer learning.
\newblock In: Proceedings of the IEEE/CVF Conference on Computer Vision and
  Pattern Recognition, pp. 7450--7459 (2019)

\bibitem{gong2018instance}
Gong, K., Liang, X., Li, Y., Chen, Y., Yang, M., Lin, L.: Instance-level human
  parsing via part grouping network.
\newblock In: Proceedings of the European Conference on Computer Vision, pp.
  770--785 (2018)

\bibitem{gong2017look}
Gong, K., Liang, X., Zhang, D., Shen, X., Lin, L.: Look into person:
  Self-supervised structure-sensitive learning and a new benchmark for human
  parsing.
\newblock In: Proceedings of the IEEE Conference on Computer Vision and Pattern
  Recognition, pp. 932--940 (2017)

\bibitem{goodfellow2014generative}
Goodfellow, I., Pouget-Abadie, J., Mirza, M., Xu, B., Warde-Farley, D., Ozair,
  S., Courville, A., Bengio, Y.: Generative adversarial nets.
\newblock In: Advances in Neural Information Processing Systems (2014)

\bibitem{guan2010a}
Guan, P., Freifeld, O., Black, M.J.: A 2d human body model dressed in eigen
  clothing.
\newblock In: Proceedings of the European Conference on Computer Vision, pp.
  285--298 (2010)

\bibitem{guler2019holopose}
Guler, R.A., Kokkinos, I.: Holopose: Holistic 3d human reconstruction
  in-the-wild.
\newblock In: Proceedings of the IEEE/CVF Conference on Computer Vision and
  Pattern Recognition, pp. 10884--10894 (2019)

\bibitem{guler2018densepose}
Guler, R.A., Neverova, N., Kokkinos, I.: Densepose: Dense human pose estimation
  in the wild.
\newblock In: Proceedings of the IEEE Conference on Computer Vision and Pattern
  Recognition, pp. 7297--7306 (2018)

\bibitem{gupta2023siamese}
Gupta, A., Wu, J., Deng, J., Fei-Fei, L.: Siamese masked autoencoders.
\newblock arXiv preprint arXiv:2305.14344  (2023)

\bibitem{han2018viton}
Han, X., Wu, Z., Wu, Z., Yu, R., Davis, L.S.: Viton: An image-based virtual
  try-on network.
\newblock In: Proceedings of the IEEE Conference on Computer Vision and Pattern
  Recognition, pp. 7543--7552 (2018)

\bibitem{hariharan2014simu}
Hariharan, B., Arbelaez, P., Girshick, R., Malik, J.: Simultaneous detection
  and segmentation.
\newblock In: Proceedings of the European Conference on Computer Vision, pp.
  297--312 (2014)

\bibitem{he2021progressive}
He, H., Zhang, J., Thuraisingham, B., Tao, D.: Progressive one-shot human
  parsing.
\newblock In: Proceedings of the AAAI Conference on Artificial Intelligence,
  pp. 1522--1530 (2021)

\bibitem{he2020grapyml}
He, H., Zhang, J., Zhang, Q., Tao, D.: Grapy-ml: Graph pyramid mutual learning
  for cross-dataset human parsing.
\newblock In: Proceedings of the AAAI Conference on Artificial Intelligence,
  pp. 10949--10956 (2020)

\bibitem{he2023end}
He, H., Zhang, J., Zhuang, B., Cai, J., Tao, D.: End-to-end one-shot human
  parsing.
\newblock IEEE Transactions on Pattern Analysis and Machine Intelligence
  (2023)

\bibitem{he2022masked}
He, K., Chen, X., Xie, S., Li, Y., Dollar, P., Girshick, R.: Masked
  autoencoders are scalable vision learners.
\newblock In: Proceedings of the IEEE/CVF Conference on Computer Vision and
  Pattern Recognition (2022)

\bibitem{he2020momentum}
He, K., Fan, H., Wu, Y., Xie, S., Girshick, R.: Momentum contrast for
  unsupervised visual representation learning.
\newblock In: Proceedings of the IEEE/CVF Conference on Computer Vision and
  Pattern Recognition, pp. 9729--9738 (2020)

\bibitem{he2017mask}
He, K., Gkioxari, G., Dollar, P., Girshick, R.: Mask r-cnn.
\newblock In: Proceedings of the IEEE International Conference on Computer
  Vision, pp. 2961--2969 (2017)

\bibitem{he2016deep}
He, K., Zhang, X., Ren, S., Sun, J.: Deep residual learning for image
  recognition.
\newblock In: Proceedings of the IEEE Conference on Computer Vision and Pattern
  Recognition, pp. 770--778 (2016)

\bibitem{hochreiter1997long}
Hochreiter, S., Schmidhuber, J.: Long short-term memory.
\newblock Neural Computation \textbf{9}(8), 1735---1780 (1997)

\bibitem{hu2022semantic}
Hu, Y., Wang, R., Zhang, K., Gao, Y.: Semantic-aware fine-grained
  correspondence.
\newblock In: European Conference on Computer Vision, pp. 97--115 (2022)

\bibitem{huang2020improve}
Huang, H., Yang, W., Lin, J., Huang, G., Xu, J., Wang, G., Chen, X., Huang, K.:
  Improve person re-identification with part awareness learning.
\newblock 2 \textbf{29}, 7468--7481 (2020)

\bibitem{huang2023composer}
Huang, L., Chen, D., Liu, Y., Shen, Y., Zhao, D., Zhou, J.: Composer: Creative
  and controllable image synthesis with composable conditions.
\newblock arXiv preprint arXiv:2302.09778  (2023)

\bibitem{huang2023ccnet}
Huang, Z., Wang, X., Wei, Y., Huang, L., Shi, H., Liu, W., Huang, T.S.: Ccnet:
  Criss-cross attention for semantic segmentation.
\newblock IEEE Transactions on Pattern Analysis and Machine Intelligence
  \textbf{45}(06), 6896--6908 (2023)

\bibitem{huo2021manifold}
Huo, J., Jin, S., Li, W., Wu, J., Lai, Y.K., Shi, Y., Gao, Y.: Manifold
  alignment for semantically aligned style transfer.
\newblock In: Proceedings of the IEEE/CVF International Conference on Computer
  Vision, pp. 14861--14869 (2021)

\bibitem{issenhuth2020do}
Issenhuth, T., Mary, J., Calauzenes, C.: Do not mask what you do not need to
  mask: a parser-free virtual try-on.
\newblock In: Proceedings of the European Conference on Computer Vision, pp.
  619--635 (2020)

\bibitem{jabri2020space}
Jabri, A.A., Owens, A., Efros, A.A.: Space-time correspondence as a contrastive
  random walk.
\newblock In: Advances in Neural Information Processing Systems, pp.
  19545--19560 (2020)

\bibitem{jeon2021mining}
Jeon, S., Min, D., Kim, S., Sohn, K.: Mining better samples for contrastive
  learning of temporal correspondence.
\newblock In: Proceedings of the IEEE/CVF Conference on Computer Vision and
  Pattern Recognition, pp. 1034--1044 (2021)

\bibitem{ji2020learning}
Ji, R., Du, D., Zhang, L., Wen, L., Wu, Y., Zhao, C., Huang, F., Lyu, S.:
  Learning semantic neural tree for human parsing.
\newblock In: Proceedings of the European Conference on Computer Vision, pp.
  205--221 (2020)

\bibitem{jia2014caffe}
Jia, Y., Shelhamer, E., Donahue, J., Karayev, S., Long, J., Girshick, R.,
  Guadarrama, S., Darrell, T.: Caffe: Convolutional architecture for fast
  feature embedding.
\newblock In: Proceedings of the 22nd ACM international conference on
  Multimedia, pp. 675--678 (2014)

\bibitem{jin2021mining}
Jin, Z., Gong, T., Yu, D., Chu, Q., Wang, J., Wang, C., Shao, J.: Mining
  contextual information beyond image for semantic segmentation.
\newblock In: Proceedings of the IEEE/CVF International Conference on Computer
  Vision, pp. 7231--7241 (2021)

\bibitem{jin2021isnet}
Jin, Z., Liu, B., Chu, Q., Yu, N.: Isnet: Integrate image-level and
  semantic-level context for semantic segmentation.
\newblock In: Proceedings of the IEEE/CVF International Conference on Computer
  Vision, pp. 7189--7198 (2021)

\bibitem{kae2013augmenting}
Kae, A., Sohn, K., Lee, H., Learned-Miller, E.: Augmenting crfs with boltzmann
  machine shape priors for image labeling.
\newblock In: Proceedings of the IEEE Conference on Computer Vision and Pattern
  Recognition, pp. 2019--2026 (2013)

\bibitem{kalayeh2018human}
Kalayeh, M.M., Basaran, E., Gokmen, M., Kamasak, M.E., Shah, M.: Human semantic
  parsing for person re-identification.
\newblock In: Proceedings of the IEEE Conference on Computer Vision and Pattern
  Recognition, pp. 1062--1071 (2018)

\bibitem{karras2019style}
Karras, T., Laine, S., Aila, T.: A style-based generator architecture for
  generative adversarial networks.
\newblock In: Proceedings of the IEEE/CVF Conference on Computer Vision and
  Pattern Recognition, pp. 4401--4410 (2019)

\bibitem{khan2020face}
Khan, K., Khan, R.U., Ahmad, K., Ali, F., Kwak, K.S.: Face segmentation: A
  journey from classical to deep learning paradigm, approaches, trends, and
  directions.
\newblock IEEE Access \textbf{8}, 58683--58699 (2020)

\bibitem{kirfel2014human}
Kiefel, M., Gehler, P.: Human pose estimation with fields of parts.
\newblock In: Proceedings of the European Conference on Computer Vision, pp.
  331---346 (2014)

\bibitem{kim2019style}
Kim, B.K., Kim, G., Lee, S.Y.: Style-controlled synthesis of clothing segments
  for fashion image manipulation.
\newblock IEEE Transactions on Multimedia \textbf{22}(2), 298--310 (2019)

\bibitem{kirillov2019pfpn}
Kirillov, A., Girshick, R., He, K., Dollar, P.: Panoptic feature pyramid
  networks.
\newblock In: Proceedings of the IEEE/CVF Conference on Computer Vision and
  Pattern Recognition, pp. 6399--6408 (2019)

\bibitem{kirillov2019panoptic}
Kirillov, A., He, K., Girshick, R., Rother, C., Dollar, P.: Panoptic
  segmentation.
\newblock In: Proceedings of the IEEE/CVF Conference on Computer Vision and
  Pattern Recognition, pp. 9404--9413 (2019)

\bibitem{kirillov2023segment}
Kirillov, A., Mintun, E., Ravi, N., Mao, H., Rolland, C., Gustafson, L., Xiao,
  T., Whitehead, S., Berg, A.C., Lo, W.Y., et~al.: Segment anything.
\newblock In: Proceedings of the IEEE/CVF International Conference on Computer
  Vision (2023)

\bibitem{krizhevsky2012imagenet}
Krizhevsky, A., Sutskever, I., Hinton, G.: Imagenet classification with deep
  convolutional neural networks.
\newblock In: Advances in Neural Information Processing Systems (2012)

\bibitem{cvpr2021l2id}
L2ID: Learning from limited or imperfect data (l2id) workshop.
\newblock \url{https://l2id.github.io/challenge_localization.html} (2021)

\bibitem{ladicky2013human}
Ladicky, L., Torr, P.H., Zisserman, A.: Human pose estimation using a joint
  pixel-wise and part-wise formulation.
\newblock In: Proceedings of the IEEE Conference on Computer Vision and Pattern
  Recognition, pp. 3578--3585 (2013)

\bibitem{lecun2015deep}
LeCun, Y., Bengio, Y., Hinton, G.: Deep learning.
\newblock Nature \textbf{521}(7553), 436--444 (2015)

\bibitem{li2017multiple}
Li, J., Zhao, J., Wei, Y., Lang, C., Li, Y., Sim, T., Yan, S., Feng, J.:
  Multiple-human parsing in the wild.
\newblock arXiv preprint arXiv:1705.07206  (2017)

\bibitem{li2022deep}
Li, L., Zhou, T., Wang, W., Li, J., Yang, Y.: Deep hierarchical semantic
  segmentation.
\newblock In: Proceedings of the IEEE/CVF Conference on Computer Vision and
  Pattern Recognition, pp. 1246--1257 (2022)

\bibitem{li2022locality}
Li, L., Zhou, T., Wang, W., Yang, L., Li, J., Yang, Y.: Locality-aware
  inter-and intra-video reconstruction for self-supervised correspondence
  learning.
\newblock In: Proceedings of the IEEE/CVF Conference on Computer Vision and
  Pattern Recognition (2022)

\bibitem{li2020correction}
Li, P., Xu, Y., Wei, Y., Yang, Y.: Self-correction for human parsing.
\newblock IEEE Transactions on Pattern Analysis and Machine Intelligence
  (2020)

\bibitem{li2017holistic}
Li, Q., Arnab, A., Torr, P.H.: Holistic, instance-level human parsing.
\newblock In: British Machine Vision Conference (2017)

\bibitem{li2023spatial}
Li, R., Liu, D.: Spatial-then-temporal self-supervised learning for video
  correspondence.
\newblock In: Proceedings of the IEEE/CVF Conference on Computer Vision and
  Pattern Recognition, pp. 2279--2288 (2023)

\bibitem{li2020self}
Li, T., Liang, Z., Zhao, S., Gong, J., Shen, J.: Self-learning with
  rectification strategy for human parsing.
\newblock In: Proceedings of the IEEE/CVF Conference on Computer Vision and
  Pattern Recognition, pp. 9263--9272 (2020)

\bibitem{li2019joint}
Li, X., Liu, S., Mello, S.D., Wang, X., Kautz, J., Yang, M.H.: Joint-task
  self-supervised learning for temporal correspondence.
\newblock In: Advances in Neural Information Processing Systems, pp. 318--328
  (2019)

\bibitem{li2023end}
Li, Z., Cao, L., Wang, H., Xu, L.: End-to-end instance-level human parsing by
  segmenting persons.
\newblock IEEE Transactions on Multimedia  (2023)

\bibitem{li2021person}
Li, Z., Lv, J., Chen, Y., Yuan, J.: Person re-identification with part
  prediction alignment.
\newblock Computer Vision and Image Understanding \textbf{205} (2021)

\bibitem{liang2014parsing}
Liang, H., Yuan, J., Thalmann, D.: Parsing the hand in depth images.
\newblock IEEE Transactions on Multimedia \textbf{16}(5), 1241--1253 (2014)

\bibitem{liang2018look}
Liang, X., Gong, K., Shen, X., Lin, L.: Look into person: Joint body parsing
  pose estimation network and a new benchmark.
\newblock IEEE Transactions on Pattern Analysis and Machine Intelligence
  \textbf{41}(4), 871--885 (2018)

\bibitem{liang2017interpretable}
Liang, X., Lin, L., Shen, X., Feng, J., Yan, S., Xing, E.P.: Interpretable
  structure-evolving lstm.
\newblock In: Proceedings of the IEEE Conference on Computer Vision and Pattern
  Recognition, pp. 1010--1019 (2017)

\bibitem{liang2016clothes}
Liang, X., Lin, L., Yang, W., Luo, P., Huang, J., Yan, S.: Clothes co-parsing
  via joint image segmentation and labeling with application to clothing
  retrieval.
\newblock IEEE Transactions on Multimedia \textbf{18}(6), 1175--1186 (2016)

\bibitem{liang2015deep}
Liang, X., Liu, S., Shen, X., Yang, J., Liu, L., Dong, J., Lin, L., Yan, S.:
  Deep human parsing with active template regression.
\newblock IEEE Transactions on Pattern Analysis and Machine Intelligence
  \textbf{37}(12), 2402--2414 (2015)

\bibitem{liang2016semantic}
Liang, X., Shen, X., Feng, J., Lin, L., Yan, S.: Semantic object parsing with
  graph lstm.
\newblock In: Proceedings of the European Conference on Computer Vision, pp.
  125--143 (2016)

\bibitem{liang2016object}
Liang, X., Shen, X., Xiang, D., Feng, J., Lin, L., Yan, S.: Semantic object
  parsing with local-global long short-term memory.
\newblock In: Proceedings of the IEEE Conference on Computer Vision and Pattern
  Recognition, pp. 3185--3193 (2016)

\bibitem{liang2015human}
Liang, X., Xu, C., Shen, X., Yang, J., Liu, S., Tang, J., Lin, L., Yan, S.:
  Human parsing with contextualized convolutional neural network.
\newblock In: Proceedings of the IEEE International Conference on Computer
  Vision, pp. 1386--1394 (2015)

\bibitem{lin2022rmgn}
Lin, C., Li, Z., Zhou, S., Hu, S., Zhang, J., Luo, L., Zhang, J., Huang, L.,
  He, Y.: Rmgn: A regional mask guided network for parser-free virtual try-on.
\newblock arXiv preprint arXiv:2204.11258  (2022)

\bibitem{lin2019face}
Lin, J., Yang, H., Chen, D., Zeng, M., Wen, F., Yuan, L.: Face parsing with roi
  tanh-warping.
\newblock In: Proceedings of the IEEE/CVF Conference on Computer Vision and
  Pattern Recognition, pp. 5654--5663 (2019)

\bibitem{lin2020human}
Lin, L., Zhang, D., Zuo, W.: Human centric visual analysis with deep learning.
\newblock Singapore: Springer (2020)

\bibitem{lin2017feature}
Lin, T.Y., Dollar, P., Girshick, R., He, K., Hariharan, B., Belongie, S.:
  Feature pyramid networks for object detection.
\newblock In: Proceedings of the IEEE Conference on Computer Vision and Pattern
  Recognition, pp. 2117--2125 (2017)

\bibitem{lin2014microsoft}
Lin, T.Y., Maire, M., Belongie, S., Hays, J., Perona, P., Ramanan, D.,
  Doll{\'a}r, P., Zitnick, C.L.: Microsoft coco: Common objects in context.
\newblock In: Proceedings of the European Conference on Computer Vision, pp.
  740--755 (2014)

\bibitem{liu2021toward}
Liu, G., Song, D., Tong, R., Tang, M.: Toward realistic virtual try-on through
  landmark-guided shape matching.
\newblock In: Proceedings of the AAAI Conference on Artificial Intelligence,
  pp. 2118--2126 (2021)

\bibitem{liu2020boosting}
Liu, J., Yao, Y., Hou, W., Cui, M., Xie, X., Zhang, C., Hua, X.S.: Boosting
  semantic human matting with coarse annotations.
\newblock In: Proceedings of the IEEE/CVF Conference on Computer Vision and
  Pattern Recognition, pp. 8563--8572 (2020)

\bibitem{liu2022cdgnet}
Liu, K., Choi, O., Wang, J., Hwang, W.: Cdgnet: Class distribution guided
  network for human parsing.
\newblock In: Proceedings of the IEEE/CVF Conference on Computer Vision and
  Pattern Recognition, pp. 4473--4482 (2021)

\bibitem{liu2013fashion}
Liu, S., Feng, J., Domokos, C., Xu, H., Huang, J., Hu, Z., Yan, S.: Fashion
  parsing with weak color-category labels.
\newblock IEEE Transactions on Multimedia \textbf{16}(1), 253--265 (2013)

\bibitem{liu2015fashion}
Liu, S., Liang, X., Liu, L., Lu, K., Lin, L., Cao, X., Yan, S.: Fashion parsing
  with video context.
\newblock IEEE Transactions on Multimedia \textbf{17}(8), 1347--1358 (2015)

\bibitem{liu2015matching}
Liu, S., Liang, X., Liu, L., Shen, X., Yang, J., Xu, C., Lin, L.: Matching-cnn
  meets knn: Quasi-parametric human parsing.
\newblock In: Proceedings of the IEEE Conference on Computer Vision and Pattern
  Recognition, pp. 1419--1427 (2015)

\bibitem{liu2018cross}
Liu, S., Sun, Y., Zhu, D., Ren, G., Chen, Y., Feng, J., Han, J.: Cross-domain
  human parsing via adversarial feature and label adaptation.
\newblock In: Proceedings of the AAAI Conference On Artificial Intelligence,
  pp. 7146--7153 (2018)

\bibitem{liu2018switchable}
Liu, S., Zhong, G., Mello, S.D., Gu, J., Jampani, V., Yang, M.H., Kautz, J.:
  Switchable temporal propagation network.
\newblock In: Proceedings of the European Conference on Computer Vision, pp.
  87--102 (2018)

\bibitem{liu2019braidnet}
Liu, X., Zhang, M., Liu, W., Song, J., Mei, T.: Braidnet: Braiding semantics
  and details for accurate human parsing.
\newblock In: Proceedings of the 27th ACM International Conference on
  Multimedia, pp. 338--346 (2019)

\bibitem{liu2019swapgan}
Liu, Y., Chen, W., Liu, L., Lew, M.S.: Swapgan: A multistage generative
  approach for person-to-person fashion style transfer.
\newblock IEEE Transactions on Multimedia \textbf{21}(9), 2209--2222 (2019)

\bibitem{liu2021hier}
Liu, Y., Zhang, S., Yang, J., Yuen, P.: Hierarchical information passing based
  noise-tolerant hybrid learning for semi-supervised human parsing.
\newblock In: Proceedings of the AAAI Conference on Artificial Intelligence,
  pp. 2207--2215 (2021)

\bibitem{liu2020hybrid}
Liu, Y., Zhao, L., Zhang, S., Yang, J.: Hybrid resolution network using edge
  guided region mutual information loss for human parsing.
\newblock In: Proceedings of the 28th ACM International Conference on
  Multimedia, pp. 1670--1678 (2020)

\bibitem{liu2021multi}
Liu, Z., Zhu, X., Yang, L., Yan, X., Tang, M., Lei, Z., Zhu, G., Feng, X.,
  Wang, Y., Wang, J.: Multi-initialization optimization network for accurate 3d
  human pose and shape estimation.
\newblock In: Proceedings of the 29th ACM International Conference on
  Multimedia, pp. 1976--1984 (2021)

\bibitem{loshchilov2018decoupled}
Loshchilov, I., Hutter, F.: Decoupled weight decay regularization.
\newblock In: Proceedings of the International Conference on Learning
  Representations (2018)

\bibitem{luo2013pedestrian}
Luo, P., Wang, X., Tang, X.: Pedestrian parsing via deep decompositional
  network.
\newblock In: Proceedings of the IEEE International Conference on Computer
  Vision, pp. 2648--2655 (2013)

\bibitem{luo2018trusted}
Luo, X., Su, Z., Guo, J.: Trusted guidance pyramid network for human parsing.
\newblock In: Proceedings of the 26th ACM International Conference on
  Multimedia, pp. 654--662 (2018)

\bibitem{luo2018macro}
Luo, Y., Zheng, Z., Zheng, L., Guan, T., Yu, J., Yang, Y.: Macro-micro
  adversarial network for human parsing.
\newblock In: Proceedings of the European Conference on Computer Vision, pp.
  418--434 (2018)

\bibitem{ma2022dual}
Ma, Z., Lin, T., Li, X., Li, F., He, D., Ding, E., Wang, N., Gao, X.:
  Dual-affinity style embedding network for semantic-aligned image style
  transfer.
\newblock IEEE Transactions on Neural Networks and Learning Systems  (2022)

\bibitem{mameli2021deep}
Mameli, M., Paolanti, M., Pietrini, R., Pazzaglia, G., Frontoni, E.,
  Zingaretti, P.: Deep learning approaches for fashion knowledge extraction
  from social media: a review.
\newblock IEEE Access  (2021)

\bibitem{mckee2022transfer}
Mckee, D., Zhan, Z., Shuai, B., Modolo, D., Tighe, J., Lazebnik, S.: Transfer
  of representations to video label propagation: implementation factors matter.
\newblock arXiv preprint arXiv:2203.05553.  (2022)

\bibitem{minaee2021image}
Minaee, S., Boykov, Y., Porikli, F., Plaza, A., Kehtarnavaz, N., Terzopoulos,
  D.: Image segmentation using deep learning: A survey.
\newblock IEEE Transactions on Pattern Analysis and Machine Intelligence
  (2021)

\bibitem{neuhold2017the}
Neuhold, G., Ollmann, T., Bulo, S.R., Kontschieder, P.: The mapillary vistas
  dataset for semantic understanding of street scenes.
\newblock In: Proceedings of the IEEE International Conference on Computer
  Vision, pp. 4990--4999 (2017)

\bibitem{nichol2021glide}
Nichol, A., Dhariwal, P., Ramesh, A., Shyam, P., Mishkin, P., McGrew, B.,
  Sutskever, I., Chen, M.: Glide: Towards photorealistic image generation and
  editing with text-guided diffusion models.
\newblock arXiv preprint arXiv:2112.10741  (2021)

\bibitem{nie2018mutual}
Nie, X., Feng, J., Yan, S.: Mutual learning to adapt for joint human parsing
  and pose estimation.
\newblock In: Proceedings of the European Conference on Computer Vision, pp.
  502--517 (2018)

\bibitem{niemeyer2021giraffe}
Niemeyer, M., Geiger, A.: Giraffe: Representing scenes as compositional
  generative neural feature fields.
\newblock In: Proceedings of the IEEE/CVF Conference on Computer Vision and
  Pattern Recognition, pp. 11453--11464 (2021)

\bibitem{ntavelis2020sesame}
Ntavelis, E., Romero, A., Kastanis, I., Gool, L.V., Timofte, R.: Sesame:
  Semantic editing of scenes by adding, manipulating or erasing objects.
\newblock In: Proceedings of the European Conference on Computer Vision, pp.
  394--411 (2020)

\bibitem{oquab2023dinov2}
Oquab, M., Darcet, T., Moutakanni, T., Vo, H., Szafraniec, M., Khalidov, V.,
  Fernandez, P., Haziza, D., Massa, F., El-Nouby, A., et~al.: Dinov2: Learning
  robust visual features without supervision.
\newblock arXiv preprint arXiv:2304.07193  (2023)

\bibitem{qian2023semantics}
Qian, R., Ding, S., Liu, X., Lin, D.: Semantics meets temporal correspondence:
  Self-supervised object-centric learning in videos.
\newblock In: Proceedings of the IEEE/CVF International Conference on Computer
  Vision, pp. 16675--16687 (2023)

\bibitem{qian2020long}
Qian, X., Wang, W., Zhang, L., Zhu, F., Fu, Y., Tao, X., Jiang, Y.G., Xue, X.:
  Long-term cloth-changing person re-identification.
\newblock In: Proceedings of the Asian Conference on Computer Vision, pp.
  71--88 (2020)

\bibitem{qin2019top}
Qin, H., Hong, W., Hung, W.C., Tsai, Y.H., Yang, M.H.: A top-down unified
  framework for instance-level human parsing.
\newblock In: British Machine Vision Conference (2019)

\bibitem{radford2021learning}
Radford, A., Kim, J.W., Hallacy, C., Ramesh, A., Goh, G., Agarwal, S., Sastry,
  G., Askell, A., Mishkin, P., Clark, J., Krueger, G., Sutskever, I.: Learning
  transferable visual models from natural language supervision.
\newblock In: International Conference on Machine Learning, pp. 8748--8763
  (2021)

\bibitem{radford2019language}
Radford, A., Wu, J., Child, R., Luan, D., Amodei, D., Sutskever, I., et~al.:
  Language models are unsupervised multitask learners.
\newblock OpenAI blog \textbf{1}(8), 9 (2021)

\bibitem{rombach2022high}
Rombach, R., Blattmann, A., Lorenz, D., Esser, P., Ommer, B.: High-resolution
  image synthesis with latent diffusion models.
\newblock In: IEEE/CVF Conference on Computer Vision and Pattern Recognition,
  pp. 10684--10695 (2022)

\bibitem{ruan2019devil}
Ruan, T., Liu, T., Huang, Z., Wei, Y., Wei, S., Zhao, Y.: Devil in the details:
  Towards accurate single and multiple human parsing.
\newblock In: Proceedings of the AAAI Conference on Artificial Intelligence,
  pp. 4814--4821 (2019)

\bibitem{russakovsky2015imagenet}
Russakovsky, O., Deng, J., Su, H., Krause, J., Satheesh, S., Ma, S., Huang, Z.,
  Karpathy, A., Khosla, A., Bernstein, M., Berg, A.C., Li, F.F.: Imagenet large
  scale visual recognition challenge.
\newblock International Journal of Computer Vision \textbf{115}(3), 211--252
  (2015)

\bibitem{schuemie2001research}
Schuemie, M.J., Straaten, P.v.d., Krijn, M., Mast, C.A.v.d.: Research on
  presence in virtual reality: A survey.
\newblock Cyberpsychology behavior \textbf{4}(2), 183--201 (2001)

\bibitem{shelhamer2016fully}
Shelhamer, E., Long, J., Darrell, T.: Fully convolutional networks for semantic
  segmentation.
\newblock IEEE Transactions on Pattern Analysis and Machine Intelligence
  \textbf{39}(4), 640--651 (2016)

\bibitem{son2022contrastive}
Son, J.: Contrastive learning for space-time correspondence via self-cycle
  consistency.
\newblock In: Proceedings of the IEEE/CVF Conference on Computer Vision and
  Pattern Recognition, pp. 14679--14688 (2022)

\bibitem{sun2019learning}
Sun, Y., Zheng, L., Li, Y., Yang, Y., Tian, Q., Wang, S.: Learning part-based
  convolutional features for person re-identification.
\newblock IEEE Transactions on Pattern Analysis and Machine Intelligence
  \textbf{43}(3), 902--917 (2019)

\bibitem{szegedy2015going}
Szegedy, C., Liu, W., Jia, Y., Sermanet, P., Reed, S., Anguelov, D., Erhan, D.,
  Vanhoucke, V., Rabinovich, A.: Going deeper with convolutions.
\newblock In: Proceedings of the IEEE Conference on Computer Vision and Pattern
  Recognition, pp. 1--9 (2015)

\bibitem{tang2021motion}
Tang, B., Jin, C., Zhang, D., Zheng, Q.: Motion human parsing: A new benchmark
  for 3d human parsing.
\newblock In: IEEE International Conference on Big Data, pp. 3203--3208 (2021)

\bibitem{tang2023humanbench}
Tang, S., Chen, C., Xie, Q., Chen, M., Wang, Y., Ci, Y., Bai, L., Zhu, F.,
  Yang, H., Yi, L., Zhao, R., Ouyang, W.: Humanbench: Towards general
  human-centric perception with projector assisted pretraining.
\newblock In: Proceedings of the IEEE/CVF Conference on Computer Vision and
  Pattern Recognition, pp. 21970--21982 (2023)

\bibitem{tian2018elimiating}
Tian, M., Yi, S., Li, H., Li, S., Zhang, X., Shi, J., Yan, J., Wang, X.:
  Eliminating background-bias for robust person re-identification.
\newblock In: Proceedings of the IEEE Conference on Computer Vision and Pattern
  Recognition, pp. 5794--5803 (2018)

\bibitem{tian2020fcos}
Tian, Z., Shen, C., Chen, H., He, T.: Fcos: A simple and strong anchor-free
  object detector.
\newblock IEEE Transactions on Pattern Analysis and Machine Intelligence
  \textbf{44}(4), 1922--1933 (2020)

\bibitem{tighe2010super}
Tighe, J., Lazebnik, S.: Superparsing: scalable nonparametric image parsing
  with superpixels.
\newblock In: Proceedings of the European Conference on Computer Vision, pp.
  352--365 (2010)

\bibitem{tseng2020modeling}
Tseng, H.Y., Fisher, M., Lu, J., Li, Y., Kim, V., Yang, M.H.: Modeling artistic
  workflows for image generation and editing.
\newblock In: Proceedings of the European Conference on Computer Vision, pp.
  158--174 (2020)

\bibitem{vaswani2017attention}
Vaswani, A., Shazeer, N., Parmar, N., Uszkoreit, J., Jones, L., Gomez, A.N.,
  Kaiser, L., Polosukhin, I.: Attention is all you need.
\newblock In: Advances in Neural Information Processing Systems, pp. 6000--6010
  (2017)

\bibitem{vondrick2018tracking}
Vondrick, C., Shrivastava, A., Fathi, A., Guadarrama, S., Murphy, K.: Tracking
  emerges by colorizing videos.
\newblock In: Proceedings of the European Conference on Computer Vision, pp.
  391--408 (2018)

\bibitem{wang2018toward}
Wang, B., Zheng, H., Liang, X., Chen, Y., Lin, L., Yang, M.: Toward
  characteristic-preserving image-based virtual try-on network.
\newblock In: Proceedings of the European Conference on Computer Vision, pp.
  589--604 (2018)

\bibitem{wang2023contextual}
Wang, D., Zhang, S.: Contextual instance decoupling for instance-level human
  analysis.
\newblock IEEE Transactions on Pattern Analysis and Machine Intelligence
  (2023)

\bibitem{wang2020deep}
Wang, J., Sun, K., Cheng, T., Jiang, B., Deng, C., Zhao, Y., Liu, D., Mu, Y.,
  Tan, M., Wang, X., Liu, W., Xiao, B.: Deep high-resolution representation
  learning for visual recognition.
\newblock IEEE Transactions on Pattern Analysis and Machine Intelligence
  \textbf{43}(10), 3349--3364 (2020)

\bibitem{wang2021contrasive}
Wang, N., Zhou, W., Li, H.: Contrastive transformation for self-supervised
  correspondence learning.
\newblock In: Proceedings of the AAAI Conference on Artificial Intelligence,
  pp. 10174--10182 (2021)

\bibitem{wang2019learning}
Wang, W., Zhang, Z., Qi, S., Shen, J., Pang, Y., Shao, L.: Learning
  compositional neural information fusion for human parsing.
\newblock In: Proceedings of the IEEE/CVF International Conference on Computer
  Vision, pp. 5703--5713 (2019)

\bibitem{wang2021survey}
Wang, W., Zhou, T., Porikli, F., Crandall, D., Gool, L.V.: A survey on deep
  learning technique for video segmentation.
\newblock arXiv preprint arXiv:2107.01153  (2021)

\bibitem{wang2021hierarchical}
Wang, W., Zhou, T., Qi, S., Shen, J., Zhu, S.C.: Hierarchical human semantic
  parsing with comprehensive part-relation modeling.
\newblock IEEE Transactions on Pattern Analysis and Machine Intelligence
  (2021)

\bibitem{wang2020hierarchical}
Wang, W., Zhu, H., Dai, J., Pang, Y., Shen, J., Shao, L.: Hierarchical human
  parsing with typed part-relation reasoning.
\newblock In: Proceedings of the IEEE/CVF Conference on Computer Vision and
  Pattern Recognition, pp. 8929--8939 (2020)

\bibitem{wang2019corres}
Wang, X., Jabri, A., Efros, A.A.: Learning correspondence from the
  cycle-consistency of time.
\newblock In: Proceedings of the IEEE/CVF Conference on Computer Vision and
  Pattern Recognition, pp. 2566--2576 (2019)

\bibitem{wei2016convolutional}
Wei, S.E., Ramakrishna, V., Kanade, T., Sheikh, Y.: Convolutional pose
  machines.
\newblock In: Proceedings of the IEEE conference on Computer Vision and Pattern
  Recognition, pp. 4724--4732 (2016)

\bibitem{wood2021fake}
Wood, E., Baltrusaitis, T., Hewitt, C., Dziadzio, S., Johnson, M., Estellers,
  V., Cashman, T.J., Shotton, J.: Fake it till you make it: Face analysis in
  the wild using synthetic data alone.
\newblock In: Proceedings of the IEEE/CVF Conference on Computer Vision and
  Pattern Recognition, pp. 3681--3691 (2021)

\bibitem{wu2021image}
Wu, B., Xie, Z., Liang, X., Xiao, Y., Dong, H., Lin, L.: Image comes dancing
  with collaborative parsing-flow video synthesis.
\newblock IEEE Transactions on Image Processing \textbf{30}, 9259--9269 (2021)

\bibitem{wu2023virtual}
Wu, D., Yang, Z., Zhang, P., Wang, R., Yang, B.: Virtual-reality interpromotion
  technology for metaverse: A survey.
\newblock IEEE Internet of Things Journal  (2023)

\bibitem{wu2019m2e}
Wu, Z., Lin, G., Tao, Q., Cai, J.: M2e-try on net: Fashion from model to
  everyone.
\newblock In: Proceedings of the 27th ACM International Conference on
  Multimedia, pp. 293--301 (2019)

\bibitem{xiq2016zoom}
Xia, F., Wang, P., Chen, L.C., Yuille, A.L.: Zoom better to see clearer: Human
  and object parsing with hierarchical auto-zoom net.
\newblock In: Proceedings of the European Conference on Computer Vision, pp.
  648--663 (2016)

\bibitem{xia2017joint}
Xia, F., Wang, P., Chen, X., Yuille, A.L.: Joint multi-person pose estimation
  and semantic part segmentation.
\newblock In: Proceedings of the IEEE Conference on Computer Vision and Pattern
  Recognition, pp. 6769--6778 (2017)

\bibitem{xia2016pose}
Xia, F., Zhu, J., Wang, P., Yuille, A.L.: Pose-guided human parsing by an
  and/or graph using pose-context features.
\newblock Proceedings of the AAAI Conference on Artificial Intelligence pp.
  3632--3640 (2016)

\bibitem{xiao2018simple}
Xiao, B., Hu, H., Wei, Y.: Simple baselines for human pose estimation and
  tracking.
\newblock In: European Conference on Computer Vision, pp. 466--481 (2018)

\bibitem{xie2021was}
Xie, Z., Zhang, X., Zhao, F., Dong, H., Kampffmeyer, M., Yan, H., Liang, X.:
  Was-vton: Warping architecture search for virtual try-on network.
\newblock In: Proceedings of the 29th ACM International Conference on
  Multimedia, pp. 3350--3359 (2021)

\bibitem{xu2021rethinking}
Xu, J., Wang, X.: Rethinking self-supervised correspondence learning: A video
  frame-level similarity perspective.
\newblock In: Proceedings of the IEEE/CVF International Conference on Computer
  Vision, pp. 10075--10085 (2021)

\bibitem{yamaguchi2013paper}
Yamaguchi, K., Hadi~Kiapour, M., Berg, T.L.: Paper doll parsing: Retrieving
  similar styles to parse clothing items.
\newblock In: Proceedings of the IEEE International Conference on Computer
  Vision, pp. 3519--3526 (2013)

\bibitem{yamaguchi2012parsing}
Yamaguchi, K., Kiapour, M.H., Ortiz, L.E., Berg, T.L.: Parsing clothing in
  fashion photographs.
\newblock In: Proceedings of the IEEE Conference on Computer Vision and Pattern
  Recognition, pp. 3570--3577 (2012)

\bibitem{yang2023semantic}
Yang, J., Wang, C., Li, Z., Wang, J., Zhang, R.: Semantic human parsing via
  scalable semantic transfer over multiple label domains.
\newblock In: Proceedings of the IEEE/CVF Conference on Computer Vision and
  Pattern Recognition, pp. 19424--19433 (2023)

\bibitem{yang2023set}
Yang, J., Zhang, H., Li, F., Zou, X., Li, C., Gao, J.: Set-of-mark prompting
  unleashes extraordinary visual grounding in gpt-4v.
\newblock arXiv preprint arXiv:2310.11441  (2023)

\bibitem{yang2019video}
Yang, L., Fan, Y., Xu, N.: Video instance segmentation.
\newblock In: Proceedings of the IEEE/CVF International Conference on Computer
  Vision, pp. 5188--5197 (2019)

\bibitem{yang2022longtailed}
Yang, L., Jiang, H., Song, Q., Guo, J.: A survey on long-tailed visual
  recognition.
\newblock International Journal of Computer Vision  (2022)

\bibitem{yang2022part}
Yang, L., Liu, Z., Zhou, T., Song, Q.: Part decomposition and refinement
  network for human parsing.
\newblock IEEE/CAA Journal of Automatica Sinica  (2022)

\bibitem{yang2020hier}
Yang, L., Song, Q., Wang, Z., Hu, M., Liu, C.: Hier r-cnn: Instance-level human
  parts detection and a new benchmark.
\newblock IEEE Transactions on Image Processing \textbf{30}, 39--54 (2020)

\bibitem{yang2020renovating}
Yang, L., Song, Q., Wang, Z., Hu, M., Liu, C., Xin, X., Jia, W., Xu, S.:
  Renovating parsing r-cnn for accurate multiple human parsing.
\newblock In: Proceedings of the European Conference on Computer Vision, pp.
  421--437 (2020)

\bibitem{yang2019parsing}
Yang, L., Song, Q., Wang, Z., Jiang, M.: Parsing r-cnn for instance-level human
  analysis.
\newblock In: Proceedings of the IEEE/CVF Conference on Computer Vision and
  Pattern Recognition, pp. 364--373 (2019)

\bibitem{yang2021quality}
Yang, L., Song, Q., Wang, Z., Liu, Z., Xu, S., Li, Z.: Quality-aware network
  for human parsing.
\newblock IEEE Transactions on Multimedia  (2022)

\bibitem{yang2021attacks}
Yang, L., Song, Q., Wu, Y.: Attacks on state-of-the-art face recognition using
  attentional adversarial attack generative network.
\newblock Multimedia Tools and Applications \textbf{80}(1), 855--875 (2021)

\bibitem{yang2018attention}
Yang, L., Song, Q., Wu, Y., Hu, M.: Attention inspiring receptive-fields
  network for learning invariant representations.
\newblock IEEE Transactions on Neural Networks and Learning Systems
  \textbf{30}(6), 1744--1755 (2018)

\bibitem{yang2019towards}
Yang, W., Huang, H., Zhang, Z., Chen, X., Huang, K., Zhang, S.: Towards rich
  feature discovery with class activation maps augmentation for person
  re-identification.
\newblock In: Proceedings of the IEEE/CVF Conference on Computer Vision and
  Pattern Recognition, pp. 1389--1398 (2019)

\bibitem{yang2011articulated}
Yang, Y., Ramanan, D.: Articulated pose estimation with flexible
  mixtures-of-parts.
\newblock In: Proceedings of the IEEE Conference on Computer Vision and Pattern
  Recognition, pp. 1385--1392 (2011)

\bibitem{yu2022hp-capsule}
Yu, C., Zhu, X., Zhang, X., Wang, Z., Zhang, Z., Lei, Z.: Hp-capsule:
  Unsupervised face part discovery by hierarchical parsing capsule network.
\newblock In: Proceedings of the IEEE/CVF Conference on Computer Vision and
  Pattern Recognition, pp. 4032--4041 (2022)

\bibitem{yu2019vtnfp}
Yu, R., Wang, X., Xie, X.: Vtnfp: An image-based virtual try-on network with
  body and clothing feature preservation.
\newblock In: Proceedings of the IEEE/CVF International Conference on Computer
  Vision, pp. 10511--10520 (2019)

\bibitem{yu2020cocas}
Yu, S., Li, S., Chen, D., Zhao, R., Yan, J., Qiao, Y.: Cocas: A large-scale
  clothes changing person dataset for re-identification.
\newblock In: Proceedings of the IEEE/CVF Conference on Computer Vision and
  Pattern Recognition, pp. 3400--3409 (2020)

\bibitem{yu2020humbi}
Yu, Z., Yoon, J.S., Li, I.K., Venkatesh, P., Park, J., Yu, J., Park, H.S.:
  Humbi: A large multiview dataset of human body expressions.
\newblock In: Proceedings of the IEEE/CVF Conference on Computer Vision and
  Pattern Recognition, pp. 2990--3000 (2020)

\bibitem{yuan2020object}
Yuan, Y., Chen, X., Wang, J.: Object-contextual representations for semantic
  segmentation.
\newblock In: Proceedings of the European Conference on Computer Vision, pp.
  173--190 (2020)

\bibitem{zeng2021neural}
Zeng, D., Huang, Y., Bao, Q., Zhang, J., Su, C., Liu, W.: Neural architecture
  search for joint human parsing and pose estimation.
\newblock In: Proceedings of the IEEE/CVF International Conference on Computer
  Vision, pp. 11385--11394 (2021)

\bibitem{zhang2023adding}
Zhang, L., Rao, A., Agrawala, M.: Adding conditional control to text-to-image
  diffusion models.
\newblock In: Proceedings of the IEEE/CVF International Conference on Computer
  Vision, pp. 3836--3847 (2023)

\bibitem{zhang2022aiparsing}
Zhang, S., Cao, X., Qi, G.J., Song, Z., Zhou, J.: Aiparsing: Anchor-free
  instance-level human parsing.
\newblock IEEE Transactions on Image Processing  (2022)

\bibitem{zhang2022human}
Zhang, X., Chen, Y., Tang, M., Wang, J., Zhu, X., Lei, Z.: Human parsing with
  part-aware relation modeling.
\newblock IEEE Transactions on Multimedia  (2022)

\bibitem{zhang2020blended}
Zhang, X., Chen, Y., Zhu, B., Wang, J., Tang, M.: Blended grammar network for
  human parsing.
\newblock In: Proceedings of the European Conference on Computer Vision, pp.
  189--205 (2020)

\bibitem{zhang2020pcnet}
Zhang, X., Chen, Y., Zhu, B., Wang, J., Tang, M.: Part-aware context network
  for human parsing.
\newblock In: Proceedings of the IEEE/CVF Conference on Computer Vision and
  Pattern Recognition, pp. 8971--8980 (2020)

\bibitem{zhang2020correlating}
Zhang, Z., Su, C., Zheng, L., Xie, X.: Correlating edge, pose with parsing.
\newblock In: Proceedings of the IEEE/CVF Conference on Computer Vision and
  Pattern Recognition, pp. 8900--8909 (2020)

\bibitem{zhang2021on}
Zhang, Z., Su, C., Zheng, L., Xie, X., Li, Y.: On the correlation among edge,
  pose and parsing.
\newblock IEEE Transactions on Pattern Analysis and Machine Intelligence
  (2021)

\bibitem{zhao2021m3d}
Zhao, F., Xie, Z., Kampffmeyer, M., Dong, H., Han, S., Zheng, T., Zhang, T.,
  Liang, X.: M3d-vton: A monocular-to-3d virtual try-on network.
\newblock In: Proceedings of the IEEE/CVF International Conference on Computer
  Vision, pp. 13239--13249 (2021)

\bibitem{zhao2017pyramid}
Zhao, H., Shi, J., Qi, X., Wang, X., Jia, J.: Pyramid scene parsing network.
\newblock In: Proceedings of the IEEE Conference on Computer Vision and Pattern
  Recognition, pp. 2881--2890 (2017)

\bibitem{zhao2018understanding}
Zhao, J., Li, J., Cheng, Y., Sim, T., Yan, S., Feng, J.: Understanding humans
  in crowded scenes: Deep nested adversarial learning and a new benchmark for
  multi-human parsing.
\newblock In: Proceedings of the 26th ACM International Conference on
  Multimedia, pp. 792--800 (2018)

\bibitem{zhao2020fine}
Zhao, J., Li, J., Liu, H., Yan, S., Feng, J.: Fine-grained multi-human parsing.
\newblock International Journal of Computer Vision \textbf{128}(8), 2185--2203
  (2020)

\bibitem{zhao2019multi}
Zhao, Y., Li, J., Zhang, Y., Tian, Y.: Multi-class part parsing with joint
  boundary-semantic awareness.
\newblock In: Proceedings of the IEEE/CVF International Conference on Computer
  Vision, pp. 9177--9186 (2019)

\bibitem{zhao2022from}
Zhao, Y., Li, J., Zhang, Y., Tian, Y.: From pose to part: Weakly-supervised
  pose evolution for human part segmentation.
\newblock IEEE Transactions on Pattern Analysis and Machine Intelligence
  (2022)

\bibitem{zhao2021modelling}
Zhao, Z., Jin, Y., Heng, P.A.: Modelling neighbor relation in joint space-time
  graph for video correspondence learning.
\newblock In: Proceedings of the IEEE/CVF International Conference on Computer
  Vision, pp. 9960--9969 (2021)

\bibitem{zheng2023deep}
Zheng, C., Wu, W., Yang, T., Zhu, S., Chen, C., Liu, R., Shen, J., Kehtarnavaz,
  N., Shah, M.: Deep learning-based human pose estimation: A survey.
\newblock ACM Computing Surveys \textbf{56}(1), 1--37 (2023)

\bibitem{zheng2018modanet}
Zheng, S., Yang, F., Kiapour, M.H., Piramuthu, R.: Modanet: A large-scale
  street fashion dataset with polygon annotations.
\newblock In: Proceedings of the 26th ACM International Conference on
  Multimedia, pp. 1670--1678 (2018)

\bibitem{zheng2019deephuman}
Zheng, Z., Yu, T., Wei, Y., Dai, Q., Liu, Y.: Deephuman: 3d human
  reconstruction from a single image.
\newblock In: Proceedings of the IEEE/CVF International Conference on Computer
  Vision, pp. 7739--7749 (2019)

\bibitem{zhou2017scene}
Zhou, B., Zhao, H., Puig, X., Fidler, S., Barriuso, A., Torralba, A.: Scene
  parsing through ade20k dataset.
\newblock In: Proceedings of the IEEE Conference on Computer Vision and Pattern
  Recognition, pp. 633--641 (2017)

\bibitem{zhou2018adaptive}
Zhou, Q., Liang, X., Gong, K., Lin, L.: Adaptive temporal encoding network for
  video instance-level human parsing.
\newblock In: Proceedings of the 26th ACM International Conference on
  Multimedia, pp. 1527--1535 (2018)

\bibitem{zhou2021differentiable}
Zhou, T., Wang, W., Liu, S., Yang, Y., Gool, L.V.: Differentiable
  multi-granularity human representation learning for instance-aware human
  semantic parsing.
\newblock In: Proceedings of the IEEE/CVF Conference on Computer Vision and
  Pattern Recognition, pp. 1622--1631 (2021)

\bibitem{zhou2023differentiable}
Zhou, T., Yang, Y., Wang, W.: Differentiable multi-granularity human parsing.
\newblock IEEE Transactions on Pattern Analysis and Machine Intelligence
  (2023)

\bibitem{zhu2018progressive}
Zhu, B., Chen, Y., Tang, M., Wang, J.: Progressive cognitive human parsing.
\newblock In: Proceedings of the AAAI Conference on Artificial Intelligence,
  pp. 7607--7614 (2018)

\bibitem{zhu2008max}
Zhu, L., Chen, Y., Lu, Y., Lin, C., Yuille, A.: Max margin and/or graph
  learning for parsing the human body.
\newblock In: Proceedings of the IEEE Conference on Computer Vision and Pattern
  Recognition, pp. 1--8 (2008)

\bibitem{zhu2020simpose}
Zhu, T., Karlsson, P., Bregler, C.: Simpose: Effectively learning densepose and
  surface normals of people from simulated data.
\newblock In: Proceedings of the European Conference on Computer Vision, pp.
  225--242 (2020)

\bibitem{zhu2021deformable}
Zhu, X., Su, W., Lu, L., Li, B., Wang, X., Dai, J.: Deformable detr: Deformable
  transformers for end-to-end object detection.
\newblock In: Proceedings of the International Conference on Learning
  Representations (2021)

\end{thebibliography}


\end{document}